\pdfoutput=1
\documentclass[11pt]{article}
\usepackage[final]{acl}
\usepackage{changepage}

\usepackage{times}
\usepackage{latexsym}
\usepackage[T1]{fontenc}
\usepackage[utf8]{inputenc}

\usepackage{microtype}
\usepackage{inconsolata}
\usepackage{graphicx}
\usepackage{tcolorbox}
\usepackage{amsmath}
\usepackage{subcaption}
\usepackage{url} 
\usepackage{booktabs}
\usepackage{amssymb}
\usepackage{todonotes}
\usepackage{multirow}
\usepackage{colortbl}
\usepackage{xcolor}

\definecolor{lightgreen}{RGB}{200,255,200}
\definecolor{mediumgreen}{RGB}{120,200,120}

\usepackage{CJKutf8}

\usepackage[russian,english]{babel}
\usepackage[T2A,T1]{fontenc}

\title{Language Arithmetics: Towards Systematic Language Neuron Identification and Manipulation}
\author{
  Daniil Gurgurov\textsuperscript{\normalfont1,2} \enspace Katharina Trinley\textsuperscript{\normalfont1} \enspace Yusser Al Ghussin\textsuperscript{\normalfont1,2} \\ \textbf{Tanja Bäumel}\textsuperscript{\normalfont1,2,3} \enspace \textbf{Josef van Genabith}\textsuperscript{\normalfont1,2} \enspace \textbf{Simon Ostermann}\textsuperscript{\normalfont1,2,3} \\
  \\
  \textsuperscript{1}Saarland University \\ \textsuperscript{2}German Research Center for AI (DFKI) 
  \\ \textsuperscript{3}Centre for European Research in Trusted AI (CERTAIN)\\
  \texttt{\small daniil.gurgurov@dfki.de}
}

\begin{document}
\maketitle

\begin{abstract}
Large language models (LLMs) exhibit strong multilingual abilities, yet the neural mechanisms behind language-specific processing remain unclear. We analyze language-specific neurons in Llama-3.1-8B, Mistral-Nemo-12B, and Aya-Expanse-8B \& 32B across \textbf{21 typologically diverse languages}, identifying neurons that control language behavior. Using the Language Activation Probability Entropy (LAPE) method, we show that these neurons cluster in deeper layers, with non-Latin scripts showing greater specialization. Related languages share overlapping neurons, reflecting internal representations of linguistic proximity.

Through \textbf{language arithmetics}, i.e. systematic activation addition and multiplication, we steer models to deactivate unwanted languages and activate desired ones, outperforming established replacement approaches. These interventions effectively guide behavior across five multilingual tasks: language forcing, translation, QA, comprehension, and NLI. Manipulation is more successful for high-resource languages, while typological similarity improves effectiveness. We also demonstrate that neuron steering enhances downstream performance and reveal internal \textbf{"fallback"} mechanisms for language selection when neurons are progressively deactivated. Our code is made publicly available at \url{https://github.com/d-gurgurov/Language-Neurons-Manipulation}.
\end{abstract}

\section{Introduction}

The emergence of large language models (LLMs) with impressive multilingual capabilities has raised fundamental questions about how these systems internally represent and process different languages \cite{wendler2024llamas, zhao2024largelanguagemodelshandle}. While models like Llama-3 \cite{grattafiori2024llama} and Gemma-3 \cite{team2024gemma} perform well across dozens of languages despite limited multilingual training data, the neural mechanisms underlying this competence are not fully understood. Understanding these mechanisms is crucial not only for advancing theoretical insights into multilingual representation learning, but also for building more controllable and interpretable language technologies \cite{amodei2016concreteproblemsaisafety, Gabriel_2020, singh2024rethinkinginterpretabilityeralarge}.

\begin{figure}[t]
  \centering
    \includegraphics[width=1.0\linewidth]{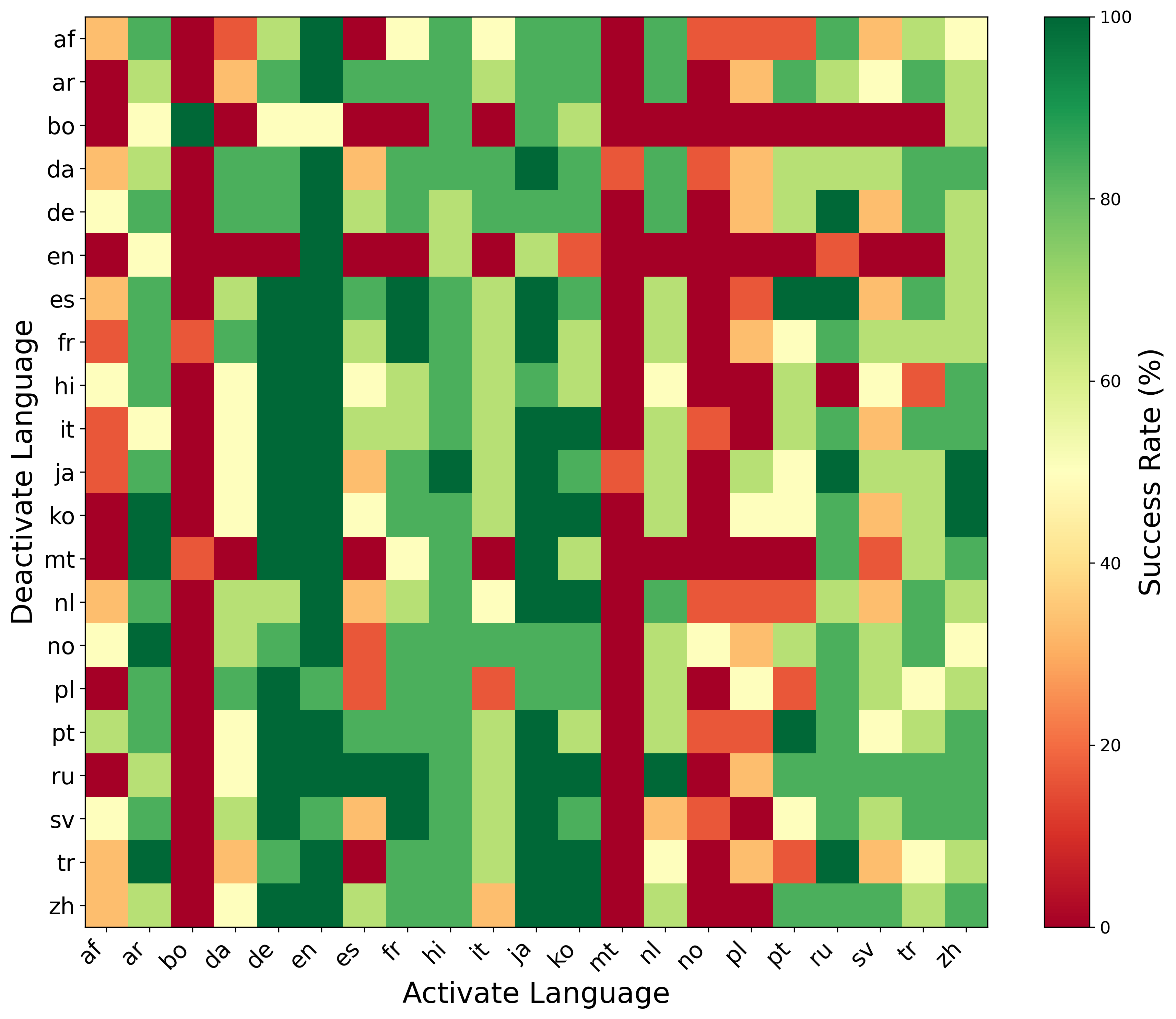}
  \caption{Success rates of language forcing when deactivating neurons for the input language and activating those of a target language for Llama-3.1-8B. The input question is presented in the language corresponding to the deactivated neurons. Top 5\% of neurons are considered.}  
  \label{fig:manip}
\end{figure}

Recent studies have begun to explore how multilingual LLMs process language internally, showing that specific neurons may specialize in particular languages \citep{tang2024language, kojima2024multilingual, zhao2024largelanguagemodelshandle, mondal-etal-2025-language}. However, these works often focus on narrow language sets or specific architectures, leaving open questions about how language-specific processing scales across typologically diverse languages and how these insights can support model control and enhancement.

In this work, \textbf{we present the first large-scale investigation of language-specific neuron identification and manipulation} in the Llama-3.1-8B \cite{grattafiori2024llama}, Mistral-Nemo-12B \cite{mistralnemo2407}, and Aya-Expanse-8B \& 32B \cite{dang2024ayaexpansecombiningresearch} models across 21 typologically diverse languages. We examine the distribution of language-sensitive neurons, their cross-linguistic overlap, and the evolution of output language distributions across layers. Using the Language Activation Probability Entropy (LAPE) method \citep{tang2024language}, we identify neurons with strong language preferences and report several key findings. Unlike previously tested models such as Llama-2 \cite{touvron2023llama}, BLOOM \cite{le2023bloom}, and Phi-2 \cite{javaheripi2023phi}, the models we study exhibit concentrated language-specific activity in their deeper layers. Analyzing output language distributions with logit lens \cite{Nostalgebraist2020} across layers reveals that language generation predominantly occurs in the final layers, aligning with the concentration of language-sensitive neurons. Non-Latin script languages show both a higher number of specialized neurons and less cross-language overlap, while typologically related Germanic and Romance languages share more neurons--reflecting linguistic proximity and possibly orthographic similarities in their internal representations. 

Building on these findings, we establish simple additive and multiplicative neuron intervention techniques, which we call \textbf{language arithmetics}\footnote{The naming follows conceptually similar works on task and prompt arithmetics \cite{ilharco2023editing,belanec2024task}.}, to test the causal role of language-specific neurons in controlling language use while preserving internal representations. Unlike prior approaches that rely on activation replacement \cite{tang2024language, kojima2024multilingual}, our approach adds target-language patterns to hidden states, while simultaneously multiplying unwanted language patterns with $0$. We demonstrate that language arithmetics outperforms both replacement-based methods \cite{tang2024language, kojima2024multilingual} and the widely used DiffMean \cite{marks2023geometry} approach on a novel \textit{language forcing} task, where models are expected to answer questions in a specific target language without being explicitly prompted, with especially strong results for languages better represented in pre-training. We further evaluate additive language arithmetic on four downstream tasks--machine translation, question answering, natural language inference, and machine comprehension--and show that targeted neuron activation improves performance without any task-specific fine-tuning, contrasting with the findings of \citet{mondal-etal-2025-language}. For example, our intervention improves translation scores by up to 10\% and enables cross-lingual transfer. Additionally, we find that when dominant language signals are suppressed, models fall back to the next most probable language, suggesting the presence of internal \textbf{"fallback"} mechanisms for language selection.

Overall, our work (1) reveals new patterns of language-specific neuron specialization and layer-wise processing, (2) applies a minimally invasive steering technique to test their causal influence, and (3) demonstrates practical benefits for multilingual model control and performance.

\section{Related Work}

Recent research has focused on uncovering language-specific mechanisms in LLMs. Several works have proposed methods to identify and intervene on neurons specialized for particular languages.

\citet{tang2024language} introduce the Language Activation Probability Entropy (LAPE) method, demonstrating that certain neurons are critical to multilingual capacity and can be manipulated to control language behavior via activation (setting neurons to their average values) or deactivation (zeroing them out). \citet{kojima2024multilingual}, building on \citet{cuadros2022self}, similarly identify neurons selectively active for one language and inactive for others, and show that setting these to their median activation can shift generation language. \citet{tan-etal-2024-neuron} further propose a frequency-based method to rank FFN neurons by activation counts on language-specific inputs, and show that fine-tuning these neurons improves downstream tasks such as machine translation. Expanding beyond FFNs, \citet{zhao2024largelanguagemodelshandle} introduce Parallel Language-specific Neuron Detection (PLND), which identifies language neurons in both attention and FFN modules. They show that deactivating these neurons reduces performance in the corresponding language, and that fine-tuning them on limited language data enhances multilingual ability.

Meanwhile, \citet{deng2025unveilinglanguagespecificfeatureslarge} highlight the problem of superposition--where neurons encode multiple concepts \citep{elhage2022toy}--and instead use sparse autoencoders (SAE) \citep{cunningham2023sparse} to extract latent dimensions linked to language identity. Ablating these features degrades performance, and they use them to guide steering vector construction for better control. Following this, \citet{chou2025causal} show that modifying a single SAE feature can steer generation language with high accuracy and semantic preservation, especially in mid-to-late layers. Similarly, \citet{andrylie2025sparse} introduce SAE-LAPE to identify interpretable language-specific features in the sparse weight space, enabling both language identification and control.

Our work extends this line of research along two key dimensions: (1) we analyze the internal structure of LLMs directly, without relying on auxiliary models such as SAEs, and (2) we examine a broader and more diverse set of languages. Specifically, we conduct a comprehensive study of language-specific neurons in Llama-3.1-8B \cite{grattafiori2024llama}, Mistral-Nemo-12B \cite{mistralnemo2407}, and Aya-Expanse-8B \& 32B \cite{dang2024ayaexpansecombiningresearch} across 21 languages. Our analysis employs the LAPE method \citep{tang2024language}, grounded in information theory, and systematically investigates both neuron identification and manipulation to evaluate their effectiveness for multilingual control and downstream performance improvement. 

\section{Language Neuron Identification}

\subsection{Identification Method}

Language Activation Probability Entropy (LAPE) \cite{tang2024language} identifies language-specific neurons within LLMs by analyzing activation patterns across different languages in the FFN modules of a transformer-based language model. 

For each neuron $j$ in layer $i$, the activation probability when processing text in language $k$ is computed as:

$$p^k_{i,j} = \mathbb{E}[\mathbf{I}(\text{act\_fn}(\tilde{h}_i W^i_1)_j > 0) \mid \text{language } k]$$

where $\mathbf{I}$ is the indicator function. The activation probabilities across all languages form a distribution $p_{i,j} = (p^1_{i,j}, p^2_{i,j}, \ldots, p^l_{i,j})$, which is L1-normalized to obtain $p'_{i,j}$. LAPE is then calculated using Shannon entropy \cite{shannon2001mathematical}:

 $$\text{LAPE}_{i,j} = -\sum_{k=1}^{l} p'^k_{i,j} \log(p'^k_{i,j})$$

Intuitively, neurons with low LAPE values are considered language-specific since their activation probabilities are concentrated on one or two languages, showing minimal activity for others.We select the bottom K\% (K=\{1..5\}) of neurons by LAPE score as candidate language-specific neurons through a three-step filtering process: (1) we exclude neurons with weak overall activity where none of their language activations exceed the 95th percentile (\textit{filter rate} = 0.95), (2) from the remaining neurons, we select the bottom K\% by LAPE score to identify those with the most language-specific activation patterns, and (3) we assign each selected neuron to all languages where its activation probability $p^k_{i,j}$ exceeds the 95th percentile threshold (\textit{activation threshold} = 0.95). While neurons can be assigned to multiple languages if they show strong activation across them, low LAPE scores typically result in assignments to only one or two languages.


\subsection{Per-layer Output Language}

We additionally analyze how language identity emerges across layers by computing three key statistics, using a method referred to as \emph{logit lens} \cite{Nostalgebraist2020}, similar to \citet{wendler2024llamas}. Specifically, we apply a FastText classifier \cite{joulin2016bag} to determine the language of the model’s output at each layer. First, we track the probability of generating the correct target language, revealing the depth at which language-specific behavior becomes prominent on the output level. Second, we measure the probability assigned to English regardless of the input, assessing the extent of English interference across layers. Third, we compute the entropy of the output language distribution to quantify the diversity and confidence of the model’s predictions.

\subsection{Models and Data}

We use the Llama-3.1-8B base model \cite{grattafiori2024llama} for all experiments. Notably, Llama-3.1 is a decoder-only model that was neither explicitly trained for multilingual tasks nor instruction-tuned. Its pre-training data includes approximately 5\% non-English content spanning over 30 languages. We also include Mistral-Nemo \cite{mistral2024nemo}, a slightly larger 12-B‑parameter base model, featuring strong multilingual, reasoning, and coding performance in its size class. Finally, we evaluate the Aya Expanse family \cite{dang2024ayaexpansecombiningresearch}: Aya-Expanse-8B, an open‑weight, instruction‑tuned multilingual model optimized via data arbitrage, preference training, and model‑merging; and its larger sibling, Aya-Expanse-32B--both supporting 23 languages and offering state‑of‑the‑art multilingual performance.

For neuron identification, we use the CulturaX corpus \cite{nguyen2023culturaxcleanedenormousmultilingual}, a large-scale multilingual dataset comprising over 6.3 trillion tokens across 167 languages. The data is sourced from Common Crawl \cite{wenzek2019ccnet} and Wikipedia \cite{wikidump} and has undergone extensive cleaning and language identification to ensure high quality. To compute language-specific activations and obtain LAPE values, we truncate the dataset for each language to 500MB (approx. 100M tokens) due to efficiency reasons. Our experiments focus on 21 representative languages, detailed in Appendix~\ref{app:langs}. For per-layer output experiments, we use 6 questions that are available in all languages, as described in Section \ref{sec:manip_data}.

\begin{figure}[t]
\centering
\begin{subfigure}[t]{1.0\linewidth}
    \centering
    \includegraphics[width=\linewidth]{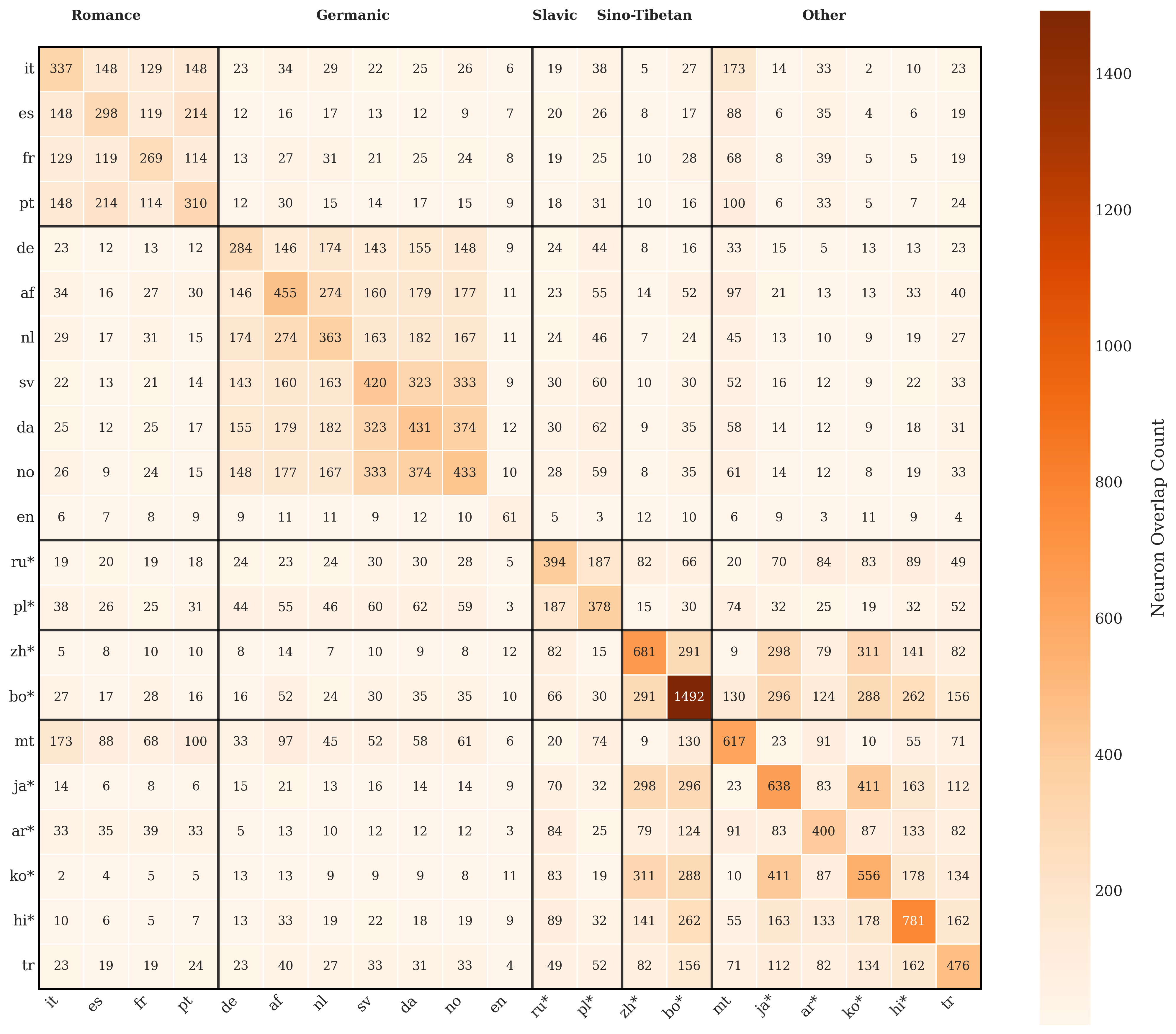}
    \label{fig:overlap_1}
\end{subfigure}

\caption{Neuron overlap between languages and language families in Llama-3.1-8B, based on the top 1\% of neurons identified as language-specific. Diagonals show counts per language; off-diagonals show overlaps. Asterisks mark non-Latin script languages.}
\label{fig:overlap}
\end{figure}

\subsection{Results}

\begin{figure}[h]
    \centering
    \includegraphics[width=1.0\linewidth]{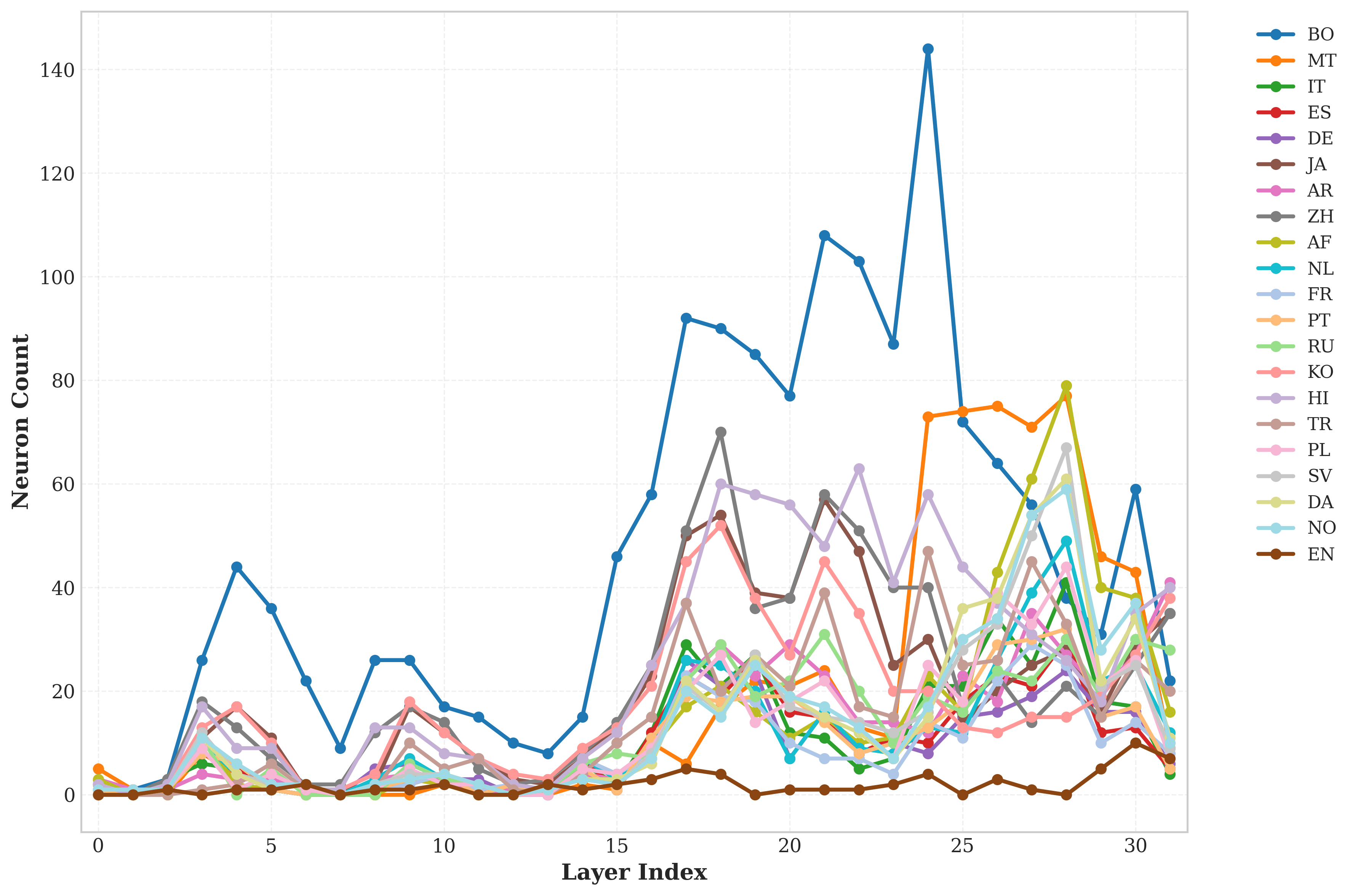}
    \caption{Layer-wise distribution of language-specific neurons for individual languages in \textbf{Llama-3.1-8B}. Other models exhibit similar patterns (Appendix \ref{appendix:language_neurons}).}
    \label{fig:distribution-plots}
\end{figure}

\textbf{Language Neuron Distribution:} 
Figure \ref{fig:distribution-plots} and Appendix \ref{appendix:language_neurons} show total neuron activation distributions with minimal early-layer activity and pronounced peaks in later layers across all 21 languages. Interestingly, language-specific processing in Llama-3.1, as well as the other tested models, demonstrates a strong concentration in layers 17-28, peaking at layers 27-28. Earlier Layers 0-15 show relatively minimal language-specific activity, with only modest peaks around layers 3-4 and 8-9. This contrasts with previous findings by \citet{tang2024language}, \citet{kojima2024multilingual}, and \citet{zhao2024largelanguagemodelshandle}, who report both early- and late-layer language neuron specialization.

Individual language distribution analysis (see Appendix \ref{appendix:language_neurons}) reveals typologically coherent clustering, with Germanic languages (German (de), Dutch (nl), Afrikaans (af), Danish (da), Norwegian (no), Swedish (sv)) showing similar language neuron distribution patterns concentrated around layers 24-26. Tibetan (bo) and Chinese (zh) (the only Sino-Tibetan languages we tested) demonstrate the most concentrated deep-layer activation, while Romance languages display more varied patterns. These patterns are observed across all tested models.

\textbf{Language Neuron Overlap:}
The language neuron overlap matrices in Figure \ref{fig:overlap} and Appendix \ref{appendix:language_neurons} show that language families exhibit internal cohesion with substantial neuron overlap, as also observed by \citet{tan-etal-2024-neuron}. In Llama-3.1, for instance, Germanic languages show particularly high overlap values (e.g., Swedish (sv) and Danish (da): 290 neurons representing 76.3\% and 74.7\% of their respective language-specific neurons; Dutch (nl) and Afrikaans (af): 250 neurons representing 76.5\% and 59.0\% respectively), possibly confirming that the similar activation patterns observed in Figure \ref{fig:individual-lang-distribution-llama} of Appendix \ref{appendix:language_neurons} reflect shared neurons rather than coincidental distributions.

Non-Latin script languages display notably higher neuron counts, with Chinese (zh*: 681), Tibetan (bo*: 1492), and Hindi (hi*: 781) having substantially more language-specific neurons than their Latin-script counterparts. Interestingly, these languages show minimal overlap with Latin-script languages, suggesting that orthographic complexity requires dedicated, non-transferable neural specialization. This is consistent across all models.

\textbf{Per-layer Language Analysis:} For more English-centric models (see Appendix~\ref{app:logit_lens}) such as Llama-3.1 and Mistral-Nemo, we observe that the target language tokens predominantly emerge in the final layers (25–30), aligning with the concentration of language-specific neurons in those layers. English tokens slightly begin to appear in the initial layers and persist through to the output. This pattern partly aligns with findings by \citet{wendler2024llamas} and \citet{schut2025multilingualllmsthinkenglish}. Language entropy in these models increases across layers: the model appears confident in its language prediction in early layers, but entropy rises from around layer 20 onward, suggesting that the decision about which language to generate is made in the final layers.

\begin{figure}[h]
    \centering
    \includegraphics[width=1\linewidth]{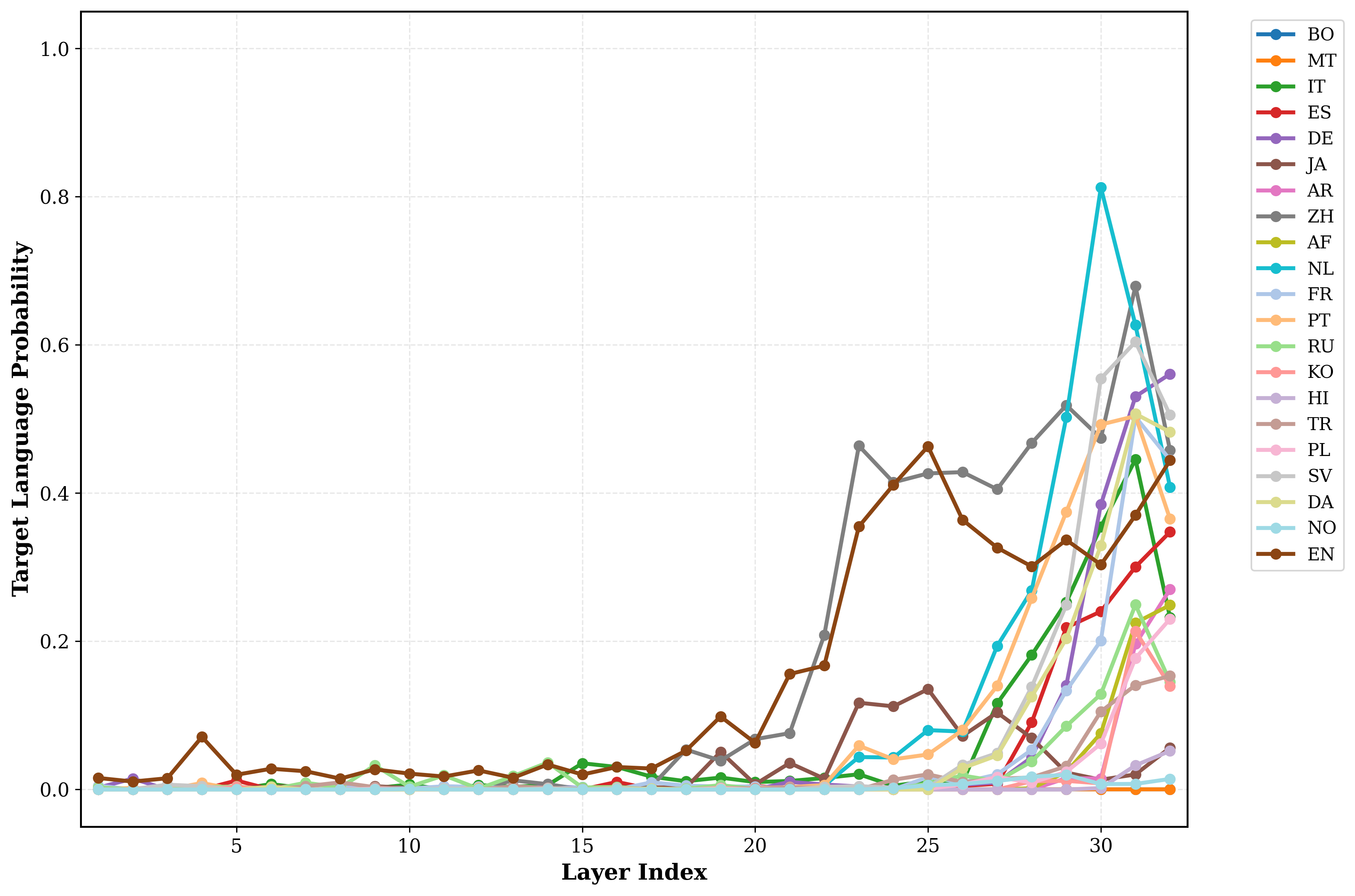}
    \caption{Predicted probabilities of target languages for Llama-3.1 using logit lens outputs from each layer and FastText for language identification. Results for other models are shown in Appendix~\ref{app:logit_lens}.}
    \label{fig:logitlens_target_llama}
\end{figure}

In contrast, more multilingual models such as Aya-Expanse-8B and 32B exhibit English token presence across nearly all layers, while the target language is also weakly activated throughout and peaks near the end. These models show an opposite entropy trend: entropy starts high, indicating early uncertainty about the generation language, and gradually decreases toward the final layers, where the model settles on a specific language.

\section{Language Neuron Manipulation}

\subsection{Method}

We adopt a manipulation approach inspired by \citet{tang2024language}, but instead of \textbf{replacing neuron activations with fixed values}, as done in \citet{tang2024language} (using mean activations) and \citet{kojima2024multilingual} (using median activations), we perform an \textbf{additive intervention}. This additive strategy is less destructive and allows the model to retain more of its original contextual dynamics while still shifting its behavior toward the target language.

Let \( \mathbf{a} \in \mathbb{R}^{L \times D} \) be the post-activation tensor after the gated projection layer for a single sequence, where \(L\) is the sequence length, and \(D\) is the hidden size. Let \( \mathcal{I} \subset \{1, \dots, D\} \) be the set of neuron indices identified as language-specific. For each \( i \in \mathcal{I} \), we define a boost value \( b_i \) corresponding to the average activation of neuron \( i \) when processing that language.

We apply the following update to the activation tensor:
\[
\mathbf{a}_{:, i} \leftarrow \mathbf{a}_{:, i} + b_i \quad \text{for all } i \in \mathcal{I}
\]

This additive manipulation shifts the activation patterns toward those typical of the desired language, while preserving the original contextual information encoded in the activations.

\begin{table*}[ht]
\small
\centering
\resizebox{\textwidth}{!}{
\begin{tabular}{l l l ccccc}
\toprule
\textbf{Model} & \textbf{Intervention} & \textbf{Strategy} & \textbf{Top 1\%} & \textbf{Top 2\%} & \textbf{Top 3\%} & \textbf{Top 4\%} & \textbf{Top 5\%} \\
\midrule
\multirow{4}{*}{Llama-3.1} 
              & Additive     & Activate              & 30.54\% & 36.62\% & 40.66\% & 42.67\% & 43.31\% \\
              & Additive     &  Deactivate + Activate & \cellcolor{mediumgreen}42.29\% & \cellcolor{mediumgreen}47.99\% & \cellcolor{mediumgreen}52.34\% & \cellcolor{mediumgreen}54.65\% & \cellcolor{mediumgreen}55.4\% \\
              & Replacement  & Activate              & 23.09\%  & 27.85\%      & 31.55\%      & 32.35\%      & 32.8\%      \\
              & Replacement  & Deactivate + Activate & 35.71\% & 38.85\%      & 42.78\%      & 42.97\%      & 42.78\%      \\
              & DiffMean & Activate & 28.45\% & 32.16\% & 35.52\% & 37.18\% & 37.49\% \\
              & DiffMean & Deactivate + Activate & \cellcolor{lightgreen}40.51\% & \cellcolor{lightgreen}44.48\% & \cellcolor{lightgreen}47.96\% & \cellcolor{lightgreen}50.15\% & \cellcolor{lightgreen}51.17\% \\
\bottomrule
\end{tabular}
}
\caption{Overall success rates (\%) of language forcing for Llama-3.1-8B using three intervention types and two manipulation strategies across different top-\textit{k}\% neuron thresholds. The results for Mistral-Nemo and Aya-Expanse follow similar patterns (see Appendix \ref{app:forcing}).}
\label{tab:language_forcing}
\end{table*}

\subsection{Tasks and Data}
\label{sec:manip_data}

We evaluate our neuron selection and manipulation method across five multilingual tasks to test its effectiveness in both generative and classification settings. For each task, we experiment with five different fractions of language-specific neurons.

First, we design a novel controlled language-forcing task using six simple questions ("How are you today?", "What is your name?", "What year is it now?", "What is your favorite color?", "What is the weather like?", "Where are you from?") translated into 20 languages using Google Translate \cite{wu2016googlesneuralmachinetranslation}. For each input question, we deactivate the neurons associated with the source language by setting their activations to zero (see Appendix \ref{app:forcing}) and activate the neurons associated with a desired target language by applying the additive intervention. We then use FastText \citep{joulin2016bag} to identify the language of the generated output. This setup evaluates the model's ability to override the input language and generate responses in the specified target language. 

We further evaluate our neuron manipulation method across four multilingual downstream tasks covering both generation and classification outputs. For generation, we use FLORES-200 \citep{costa2022no} for machine translation, activating only target-language neurons during decoding to test whether neuron-level steering can enforce correct-language outputs, and XQuAD \citep{Artetxe_2020} for extractive question answering. For classification, we use XNLI \citep{conneau-etal-2018-xnli} for natural language inference and Belebele \citep{bandarkar-etal-2024-belebele} for multiple-choice reading comprehension. We vary the ratio of manipulated neurons to assess how language-specific activation impacts multilingual performance. All evaluations are performed in a zero-shot setting, with prompts and generation hyperparameters detailed in Appendix \ref{app:generation}.

Additionally, we investigate whether models exhibit a \textbf{"fallback"} mechanism in language selection during generation. To this end, we use a set of 70 English questions from Vicuna \cite{vicuna2023} and progressively deactivate neurons associated with high-resource languages one by one.

\subsection{Results}

\textbf{Language Forcing:} For the language forcing task, we compare two neuron manipulation strategies: (1) deactivating neurons associated with the source language and simultaneously activating neurons for the target language, and (2) only activating the neurons corresponding to the desired target language. Our results show that the first approach more reliably steers the model to generate output in the target language (Table \ref{tab:language_forcing} and Appendix \ref{app:forcing}). We additionally compare our additive language arithmetic intervention with the simple replacement methods \cite{tang2024language, kojima2024multilingual} and the widely used DiffMean approach \cite{marks2023geometry, panickssery2023steering}\footnote{For DiffMean, we construct the steering vector as the difference between the mean activation for the target language (positive class) and that of all 20 other languages (negative class).} (see Table~\ref{tab:language_forcing} and Appendix~\ref{app:forcing}). We find that the additive method performs best across all tested models, likely because it is less disruptive to the model’s internal representations.

Figure~\ref{fig:manip} illustrates the effects of this manipulation across 21 typologically diverse languages in Llama-3.1-8B with the top 5\% neuron ratio. The results indicate that higher-resourced languages such as Chinese (zh), Korean (ko), Japanese (ja), German (de), etc. are more susceptible to successful language forcing. In contrast, languages with likely lower representation in the pretraining corpus are more difficult to control, showing inconsistent output. 

Below, we provide qualitative examples demonstrating successful language forcing, where the model generates output in a specified target language despite the input being in a different source language:

\begin{tcolorbox}[colback=gray!5!white, colframe=black!75!white,
                  title=Forcing Japanese, fonttitle=\bfseries,
                  boxrule=0.5pt, arc=4pt, left=4pt, right=4pt, top=4pt, bottom=4pt]
\small
\textbf{Input:} Q: Wie geht es dir heute? A: \\
\textbf{Output:} \begin{CJK}{UTF8}{min}今日は元気です。\end{CJK}
\end{tcolorbox}

\begin{tcolorbox}[colback=gray!5!white, colframe=black!75!white,
                  title=Forcing German, fonttitle=\bfseries,
                  boxrule=0.5pt, arc=4pt, left=4pt, right=4pt, top=4pt, bottom=4pt]
\small
\textbf{Input:} Q: \foreignlanguage{russian}{Откуда ты?} A:\\
\textbf{Output:} Ich komme aus Deutschland.
\end{tcolorbox}

\begin{tcolorbox}[colback=gray!5!white, colframe=black!75!white,
                  title=Forcing Korean, fonttitle=\bfseries,
                  boxrule=0.5pt, arc=4pt, left=4pt, right=4pt, top=4pt, bottom=4pt]
\small
\textbf{Input:} Q: Comment tu t'appelles? A:\\
\textbf{Output:} \begin{CJK}{UTF8}{mj}저는 김민수입니다.\end{CJK}.
\end{tcolorbox}

\begin{figure*}[t!]
  \centering
  \begin{minipage}[t]{0.48\textwidth}
    \centering
    \includegraphics[width=\linewidth]{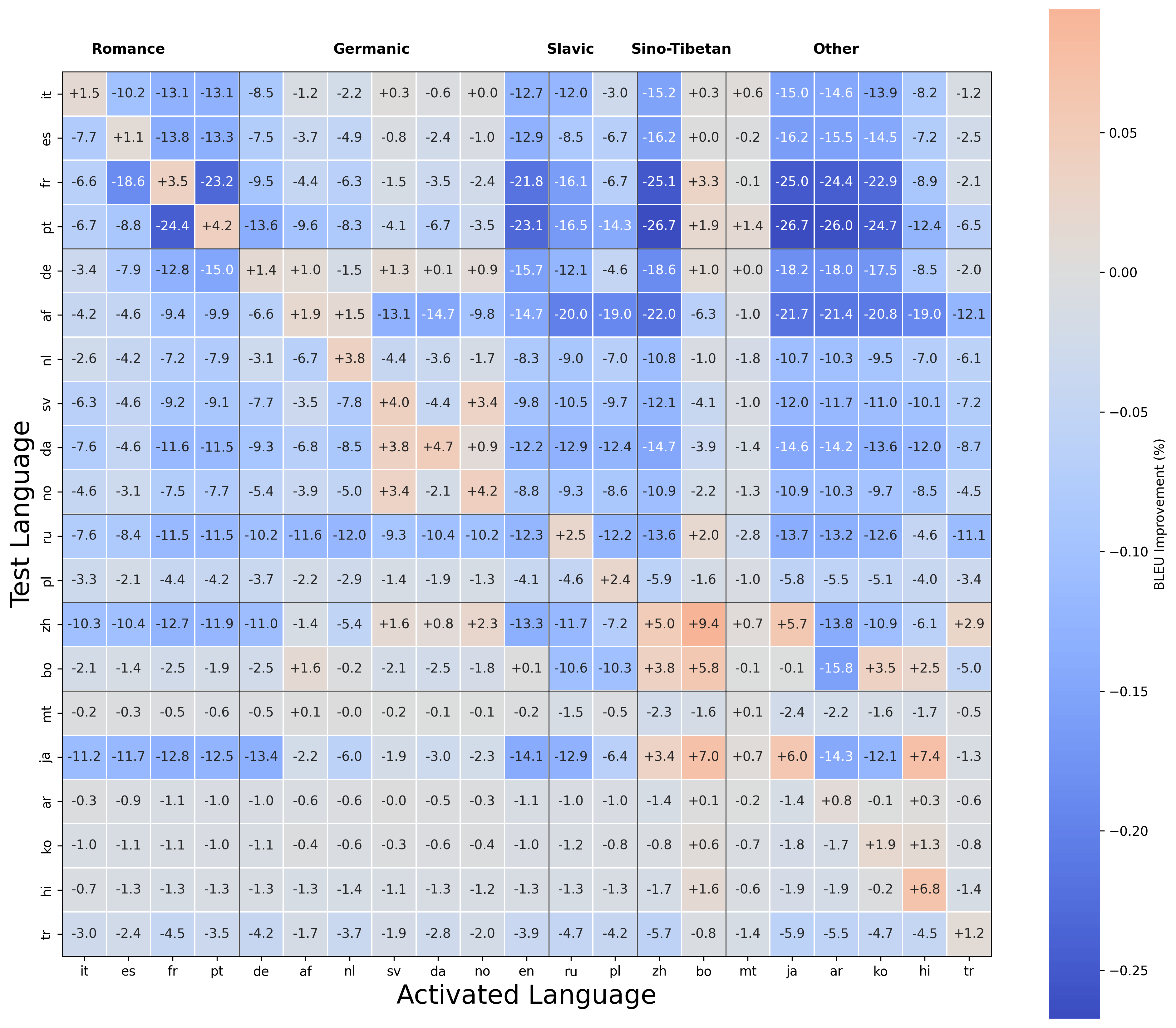}
    \caption{FLORES performance changes over the baseline (measured by BLEU score) when activating language-specific neurons for Mistral-Nemo (5\%).}
    \label{fig:flores}
  \end{minipage}
  \hfill
  \begin{minipage}[t]{0.48\textwidth}
    \centering
    \includegraphics[width=\linewidth]{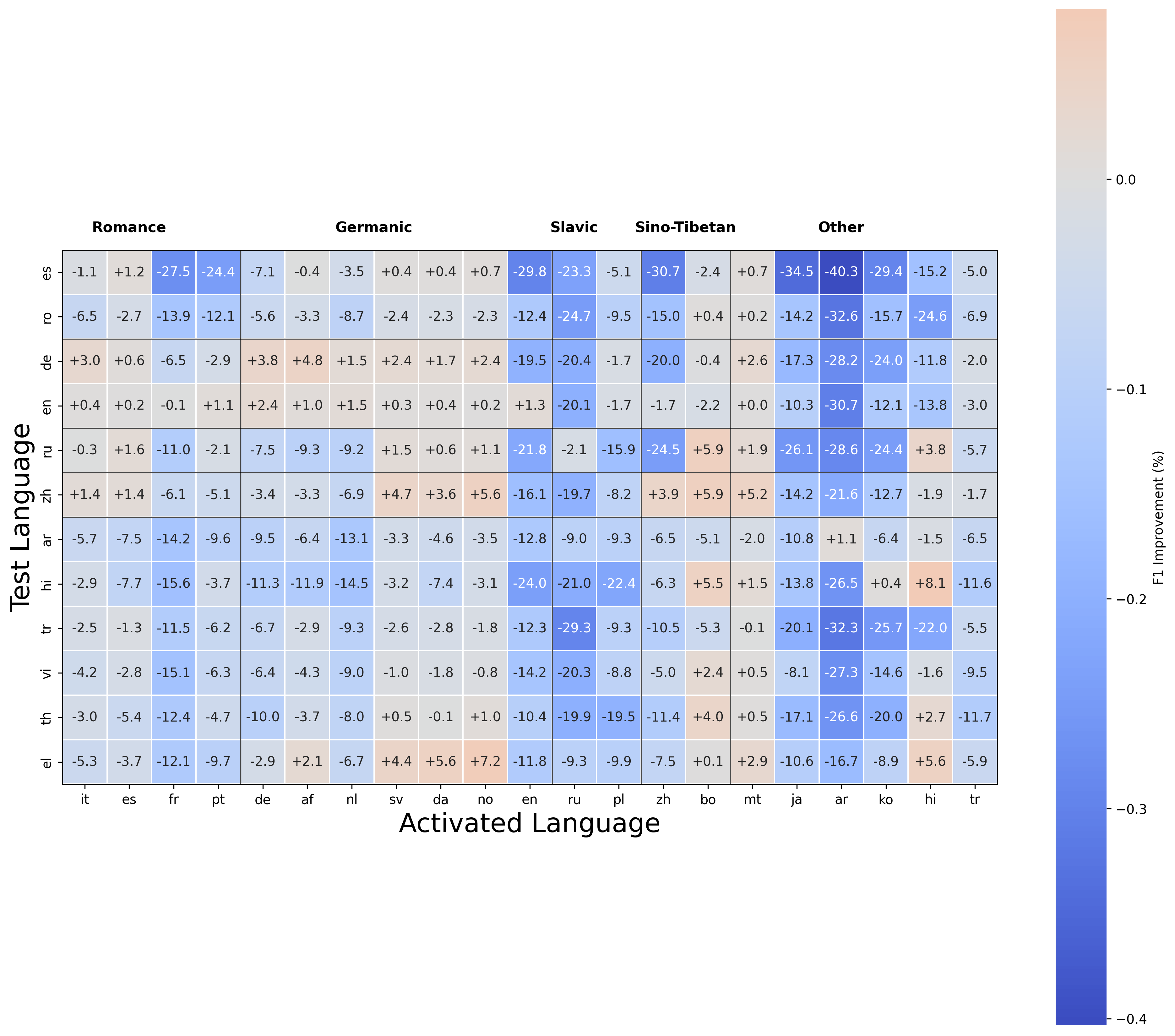}
    \caption{QXuAD performance changes over the baseline (measured by F1-score) when activating language-specific neurons for Mistral-Nemo (5\%).}
    \label{fig:xquad}
  \end{minipage}
  
  \vspace{5pt}

  \begin{minipage}[t]{0.48\textwidth}
    \centering
    \includegraphics[width=\linewidth]{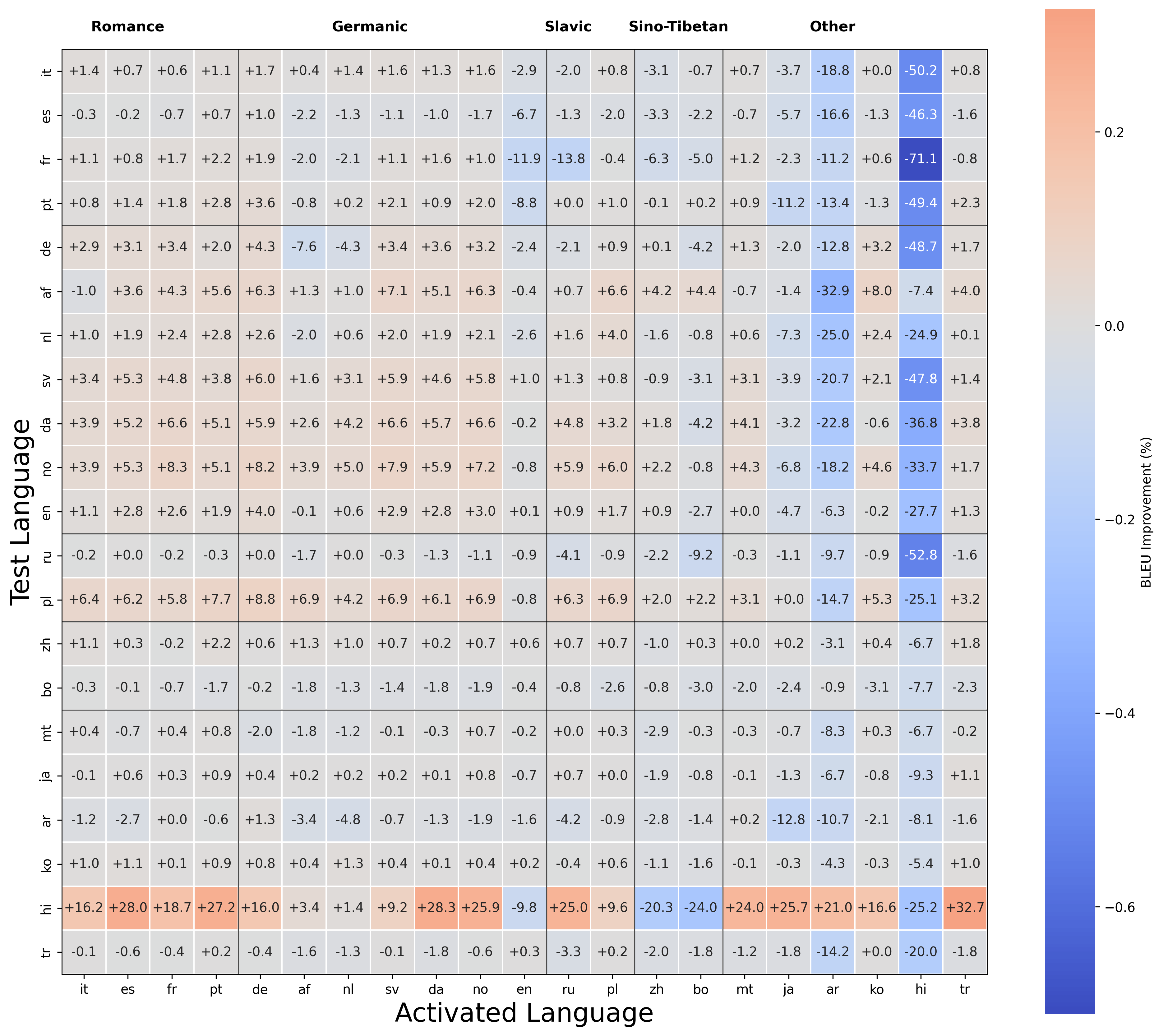}
    \caption{BELEBELE changes over the baseline (measured by accuracy score) when activating language-specific neurons for Mistral-Nemo (5\%).}
    \label{fig:belebele}
  \end{minipage}
  \hfill
  \begin{minipage}[t]{0.48\textwidth}
    \centering
    \includegraphics[width=\linewidth]{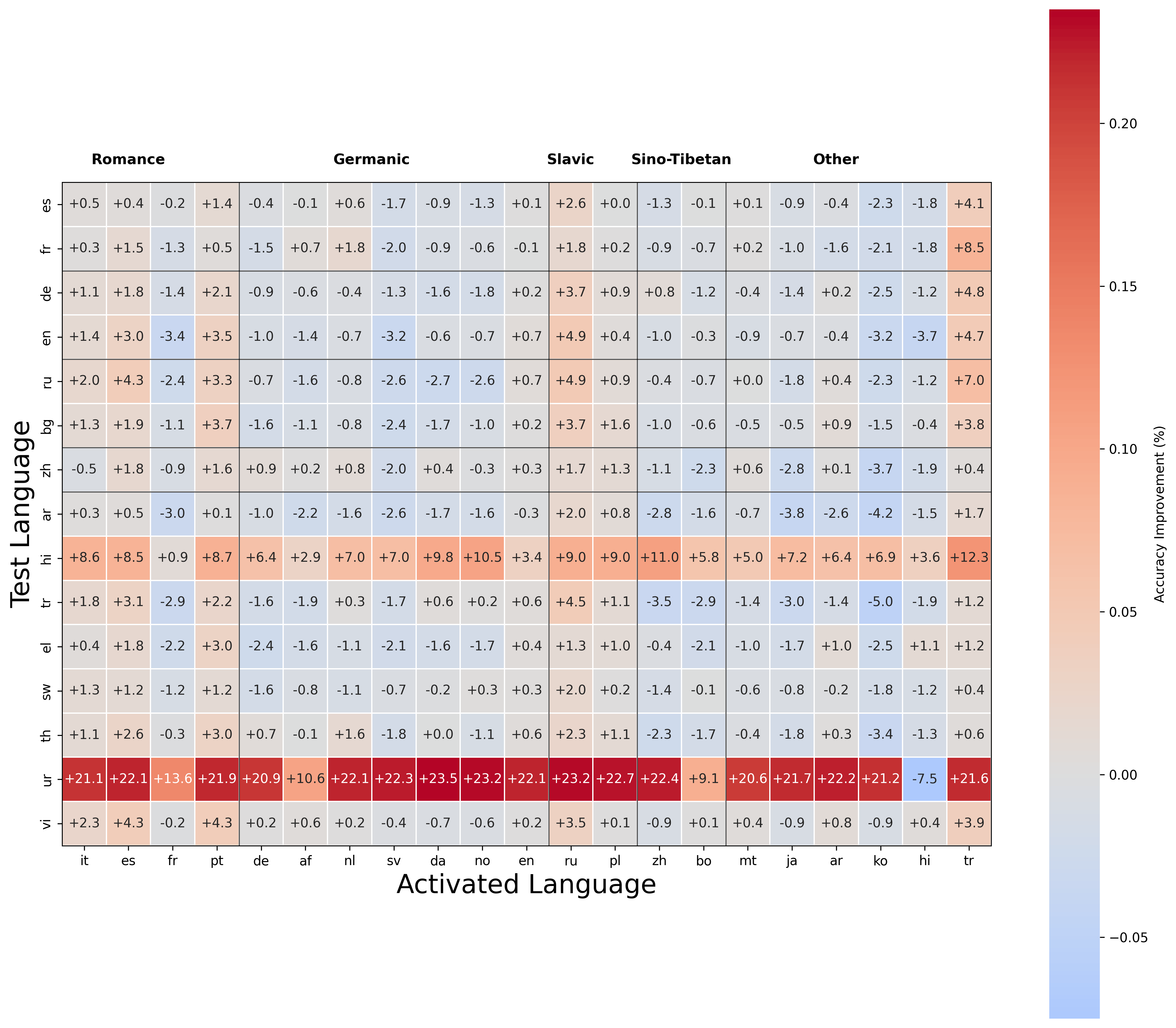}
    \caption{XNLI changes over the baseline (measured by accuracy score) when activating language-specific neurons for Llama-3.1 (5\%).}
    \label{fig:xnli}
  \end{minipage}
\end{figure*}

\textbf{Downstream Tasks:} We evaluate the impact of activating language-specific neurons during inference on four multilingual benchmarks: FLORES-200 and Belebele using all languages covered in our study, XNLI and XQuAD using all available languages. We conduct systematic evaluations on Llama-3.1 and Mistral-Nemo across five neuron ratios. Selected heatmap results are shown in Figures \ref{fig:flores}, \ref{fig:belebele}, \ref{fig:xnli}, and \ref{fig:xquad}.

\begin{figure*}[hbt]
\centering

\begin{subfigure}{\linewidth}
    \centering
    \includegraphics[width=0.9\linewidth]{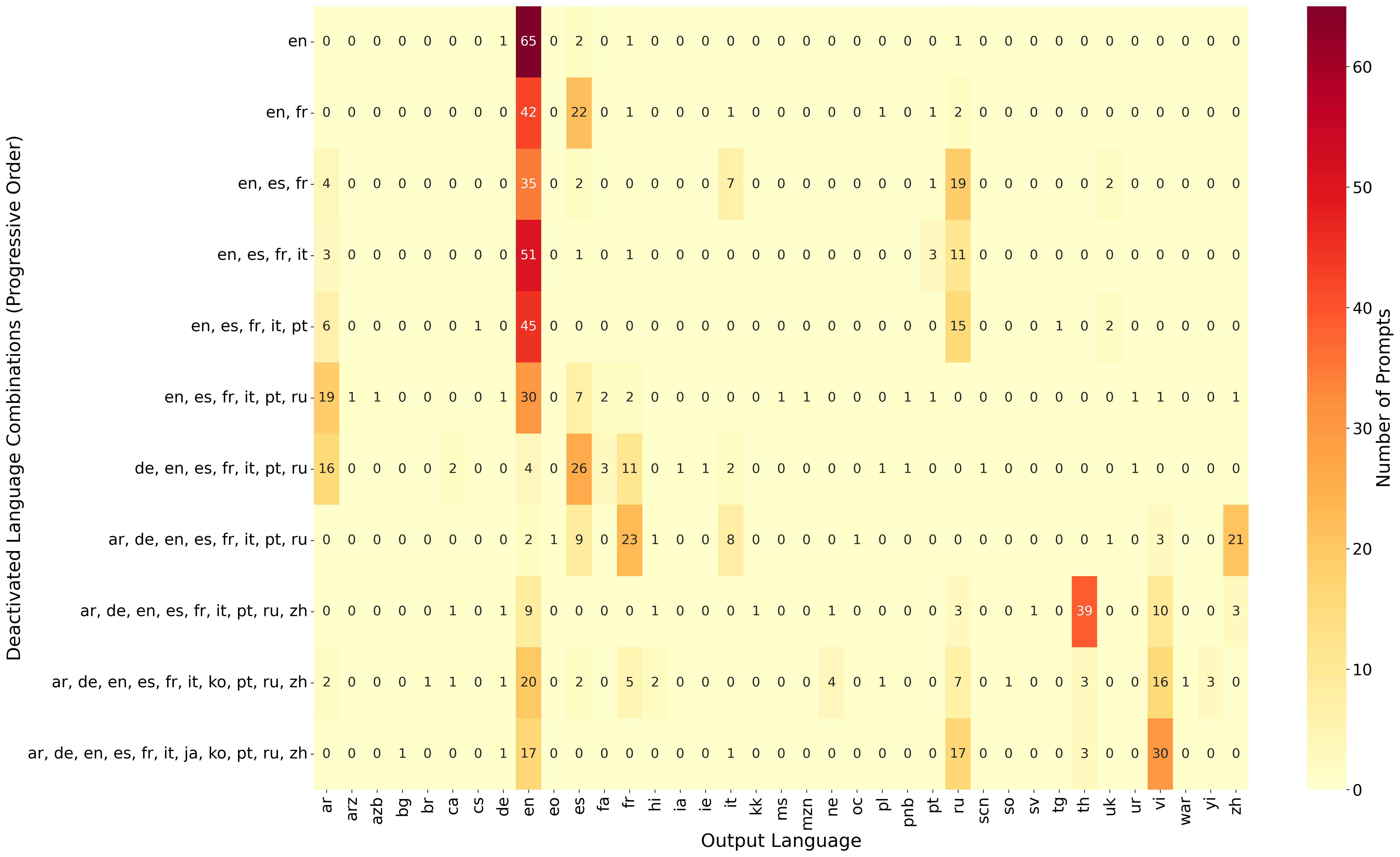}
\end{subfigure}

\caption{
Progressive deactivation of language-specific neurons for high-resource languages in Llama-3.1-8B. In this setting, language neurons are deactivated by setting their activations to $-1$; using $0$ does not yield the desired effect. After each deactivation round, the model is prompted with 70 English questions, and we then identify the output language of its responses.
}
\label{fig:fallbacks_llama_5}
\end{figure*}

Activating target language neurons improves performance in nearly all cases across all tasks and languages, with gains typically ranging from 5-15\%, as opposed to the negative effects reported by \citet{mondal-etal-2025-language}. Cross-lingual transfer effects vary substantially by task type. For generative tasks--FLORES translation and XQuAD question answering--we observe minimal cross-lingual improvements and frequent performance degradations when activating non-target languages. This pattern is particularly pronounced in FLORES, where activating neurons from different languages often causes the model to generate outputs in the activated language rather than the target language, explaining the consistent negative transfer. XQuAD exhibits similar behavior, though somewhat less severe, as the generative nature of the task makes it susceptible to language confusion.
In contrast, discriminative tasks--Belebele reading comprehension and XNLI natural language inference--show markedly different patterns. Here, activating target language neurons remains beneficial, but we also observe substantial positive cross-lingual transfer, particularly from related languages within the same family. Romance language activations benefit other Romance languages, and Germanic activations similarly help Germanic languages, with improvements of 5-10\% in many cases. Notably, Hindi and Urdu demonstrate exceptional robustness, maintaining or improving performance under nearly all language activations across both task types, possibly due to polysemantic neurons, units encoding multiple concepts \citep{elhage2022toy}.

\textbf{Language "Fallbacks":} By progressively deactivating high-resource language neurons in Llama‑3.1‑8B (see Figure \ref{fig:fallbacks_llama_5} and Appendix~\ref{app:fallbacks}), we observe that the model exhibits internal "fallback" strategies for language generation. When English (en) neurons are deactivated, the model defaults to Spanish (es), French (fr), and Russian (ru) in a few cases. Deactivating English and French shifts responses primarily to Russian (ru), with some in Italian (it), Spanish (es), and Arabic (ar). Further removal of Latin-script languages--Spanish, Italian, and Portuguese (pt)--promotes generation in Russian, Arabic, and Chinese. Deactivating German, Russian, and Arabic leads to "fallback" responses in Thai, Vietnamese, and Chinese. Notably, some English, Spanish, and French responses persist despite deactivation, reflecting the model’s strong priors for these languages. Similar patterns are observed for Mistral-Nemo, but with a slightly different language hierarchy. We leave a more detailed investigation of these dynamics to future work.

\section{Discussion}

\subsection{Identification Insights}
The layer-wise distribution of language-specific neurons in Llama-3.1, Mistral-Nemo, and Aya-Expanse partially aligns with prior findings \cite{tang2024language, kojima2024multilingual, zhao2024largelanguagemodelshandle}. We observe that these neurons are predominantly concentrated in the later layers, whereas \citet{tang2024language} and \citet{kojima2024multilingual} report a dual-peaked distribution in older models, with concentrations in both early and late layers. Notably, \citet{tang2024language} also show that the more multilingual BLOOM model \cite{le2023bloom} exhibits a similar late-layer concentration, a trend further supported by \citet{mondal-etal-2025-language} with recent multilingual models. This divergence likely reflects architectural differences or variations in pre-training data composition and training procedures. Interestingly, \citet{chou2025causal} demonstrate that modifying a small number of SAE features from mid-to-late layers enables high-accuracy language steering for high-resource languages, while \citet{andrylie2025sparse} similarly find that their SAE-LAPE method identifies language-specific features concentrated in the same layers. These converging results, obtained through entirely different methodologies, reinforce our observation that language-specific representations are predominantly localized in the mid-to-late transformer layers of modern multilingual LLMs.

Layer-wise logit lens analysis complements these findings: in all tested models, English token probabilities emerge early on and persist, while target language tokens appear more sharply in the very final layers, which is in line with \citet{wendler2024llamas}, \citet{schut2025multilingualllmsthinkenglish}, and \citet{wang2025lost}. Entropy trends in more English-centric models rise toward the end, suggesting that language generation decisions are finalized in the later layers. In terms of cross-linguistic neuron overlap, we observe clear clustering patterns based on language families. This supports findings from \citet{tan-etal-2024-neuron}, who report that related languages share substantial neural components while maintaining distinct processing features.

\subsection{Manipulation Insights}

Our findings show that neuron-level interventions can reliably steer multilingual model behavior, with our additive language arithmetic method outperforming activation replacement \cite{tang2024language, kojima2024multilingual} and DiffMean-based approaches \cite{marks2023geometry}. Activating language-specific neurons improves performance across both generative and discriminative tasks, with the strongest gains observed when the activated language matches the target or is typologically related, yielding positive transfer within language families for some tasks, but more limited benefits for structurally distant languages. Interestingly, Hindi and Urdu show disproportionate improvements across many interventions, potentially due to neurons encoding multiple overlapping functions \citep{elhage2022toy}. Moreover, compared to \citet{mondal-etal-2025-language}, who report limited improvements from activation replacement or language neuron fine-tuning on tasks like XNLI \cite{conneau2018xnli} and XQuAD \cite{artetxe2019cross}, our method achieves performance gains due to three key differences. First, we evaluate across a broader and more typologically diverse set of languages. Second, our additive interventions using mean activation values preserve more of the original representational context. Third, all evaluations are conducted in a strict zero-shot setting, without fine-tuning on downstream tasks in any source language. 

The effects are especially pronounced in our language forcing experiments. Deactivating neurons for the input language while simultaneously activating neurons for a desired target language significantly increases the likelihood of generating output in the target language--even when the input and output languages are typologically distant. These results suggest that target-language neurons contain sufficient signal to override source-language priors, especially for languages better represented in pre-training. In contrast, low-resource languages are harder to control, likely due to weaker or less distinct neural representations resulting from limited pretraining exposure. Importantly, manipulation also reveals internal "fallback" mechanisms: when dominant language neurons like English or French are deactivated, the model switches to secondary high-resource languages such as Russian or Arabic. This "fallback" hierarchy varies slightly across models and language deactivation orders but consistently exposes the model's internal language preferences. The persistence of certain languages despite deactivation--particularly English--highlights the robustness of high-priority language priors in LLMs.


\section{Conclusion}

Our large-scale study demonstrates that language-specific neurons encode meaningful linguistic features that can be identified, analyzed, and manipulated to steer model behavior across languages. By employing a novel additive intervention method, we show that activating these neurons allows for generation language control and improves performance on various multilingual tasks. These findings underscore the potential of neuron-level control as an interpretable mechanism for guiding multilingual model outputs.

\section*{Limitations}

Our study has several limitations. First, while we demonstrate the effectiveness of neuron-level interventions across multiple tasks, a broader evaluation spanning more diverse downstream settings--such as dialogue, summarization, or code generation--is needed to assess the generality and robustness of our approach. Second, although our interventions often produce outputs in the desired language, the quality and fluency of these forced outputs across different languages--especially lower-resourced ones--remain underexplored and require more rigorous evaluation. Third, our current analysis primarily focuses on isolated neurons with high language specificity. This overlooks the possibility that language representations may be distributed across larger sub-networks or circuits; future work should investigate the interactions and dependencies between such neurons to better understand the network-level mechanisms supporting multilingual processing.

Additionally, due to computational constraints, not all experiments were run across every model in our study. In particular, some analyses (large-scale downstream evaluation) were limited to smaller models--LLama-3.1-8B and Mistral-Nemo-14B. Expanding these evaluations to include larger models could reveal further insights but requires significantly more resources.

\section*{Acknowledgments}
This research was supported by \textit{DisAI - Improving scientific excellence and creativity in combating disinformation with artificial intelligence and language technologies}, a project funded by Horizon Europe under \href{https://doi.org/10.3030/101079164}{GA No.101079164}, by \textit{lorAI - Low Resource Artificial Intelligence}, a project funded by the European Union under \href{https://doi.org/10.3030/101136646}{GA No.101136646}, and by the German Federal Ministry of Research, Technology and Space (BMFTR) as part of the project TRAILS (01IW24005).
\bibliography{custom}

\appendix
\clearpage
\newpage

\section*{Appendix}
\section{Prompt Templates}
\label{app:generation}

\subsection{FLORES-200 Machine Translation}
For the machine translation task, we use the following prompt format to instruct the model to translate from a source language into a target language:

\begin{quote}
\texttt{Translate this \{source\_name\} sentence into \{target\_name\}: \{source\_text\}. Translation:}
\end{quote}

Here, \texttt{\{source\_name\}} and \texttt{\{target\_name\}} are replaced by the names of the source and target languages (e.g., \texttt{French}, \texttt{Hindi}), and \texttt{\{source\_text\}} is the input sentence.

\subsection{XQuAD Question Answering}
For extractive question answering, we prompt the model to find an answer span in a given context:

\begin{quote}
\texttt{Answer the question based on the \{language\_name\} context provided. Extract the exact answer from the context.}

\texttt{Context: \{context\}}

\texttt{Question: \{question\}}

\texttt{Answer:}
\end{quote}

Here, \texttt{\{language\_name\}} is the name of the language (e.g., \texttt{Hindi}), and \texttt{\{context\}}, \texttt{\{question\}} are the inputs.

\subsection{Belebele Machine Comprehension}
For the multiple-choice machine comprehension task, we use:

\begin{quote}
\texttt{Read the following \{language\_name\} passage and answer the question.}

\texttt{Passage: \{passage\}}

\texttt{Question: \{question\}}

\texttt{A. \{option\_A\}}

\texttt{B. \{option\_B\}}

\texttt{C. \{option\_C\}}

\texttt{D. \{option\_D\}}

\texttt{Answer:}
\end{quote}

Each \texttt{\{option\_X\}} is a possible answer choice, and the model is expected to return one of \texttt{A}, \texttt{B}, \texttt{C}, or \texttt{D}.

\subsection{XNLI Natural Language Inference}
To classify the relationship between a premise and hypothesis, we use:

\begin{quote}
\texttt{Given the following \{language\_name\} premise and hypothesis, determine the relationship between them.}

\texttt{Premise: \{premise\}}

\texttt{Hypothesis: \{hypothesis\}}

\texttt{Options:}

\texttt{1. Entailment}

\texttt{2. Neutral}

\texttt{3. Contradiction}

\texttt{Answer:}
\end{quote}

The model is expected to return one of the listed option numbers or labels.

\subsection{Generation Settings}

We use the same decoding configuration across all tasks unless specified otherwise. The generation is performed using the following sampling parameters:

\begin{quote}
\texttt{SamplingParams(}\
\texttt{\hspace{0.5cm}temperature=0,}\
\texttt{\hspace{0.5cm}repetition\_penalty=1.1,}\
\texttt{\hspace{0.5cm}stop\_token\_ids=[eos\_token\_id] if eos\_token\_id is not None else [],}\
\texttt{\hspace{0.5cm}skip\_special\_tokens=True}\
\texttt{)}
\end{quote}

We vary the \texttt{max\_tokens} parameter depending on the task:

\begin{itemize}
\item \textbf{FLORES-200 (Machine Translation):} \texttt{max\_tokens = 128}
\item \textbf{XQUAD (Question Answering):} \texttt{max\_tokens = 64}
\item \textbf{XNLI (Natural Language Inference):} \texttt{max\_tokens = 32}
\item \textbf{BELEBELE (Machine Comprehension):} \texttt{max\_tokens = 32}
\item \textbf{Language Forcing Experiments:} \texttt{max\_tokens = 256}
\end{itemize}

These settings ensure deterministic generation (due to \texttt{temperature=0}) while reducing repetition and enabling flexible truncation across task types.

\begin{table*}[t!]
\section{Selected Languages}
\label{app:langs}
\centering
\begin{tabular}{lll}
\toprule
\textbf{Language Code} & \textbf{Language Name} & \textbf{Language Family} \\
\midrule
bo & Tibetan & Sino-Tibetan \\
mt & Maltese & Afro-Asiatic \\
it & Italian & Indo-European (Romance) \\
es & Spanish & Indo-European (Romance) \\
de & German & Indo-European (Germanic) \\
ja & Japanese & Japonic \\
ar & Arabic & Afro-Asiatic (Semitic) \\
zh & Chinese & Sino-Tibetan \\
af & Afrikaans & Indo-European (Germanic) \\
nl & Dutch & Indo-European (Germanic) \\
fr & French & Indo-European (Romance) \\
pt & Portuguese & Indo-European (Romance) \\
ru & Russian & Indo-European (Slavic) \\
ko & Korean & Koreanic \\
hi & Hindi & Indo-European (Indo-Aryan) \\
tr & Turkish & Turkic \\
pl & Polish & Indo-European (Slavic) \\
sv & Swedish & Indo-European (Germanic) \\
da & Danish & Indo-European (Germanic) \\
no & Norwegian & Indo-European (Germanic) \\
en & English & Indo-European (Germanic) \\
\bottomrule
\end{tabular}
\caption{
Language codes, names, and families of the 21 languages used in our experiments. 
Language family classification follows \textit{Glottolog} \citep{glottolog}.
}
\label{tab:languages}
\end{table*}

\clearpage

\begin{table}[h]
\section{Neuron Counts}
\small
\centering
\begin{tabular}{lccccc}
\toprule
\textbf{Lang.} & \textbf{1\%} & \textbf{2\%} & \textbf{3\%} & \textbf{4\%} & \textbf{5\%} \\
\midrule
bo & 1492 & 2687 & 3723 & 4708 & 5687 \\
mt & 617 & 1578 & 2687 & 3853 & 5043 \\
it & 337 & 761 & 1198 & 1653 & 2104 \\
es & 298 & 667 & 1007 & 1358 & 1717 \\
de & 284 & 651 & 1065 & 1509 & 1965 \\
ja & 638 & 1084 & 1503 & 1887 & 2286 \\
ar & 400 & 708 & 1093 & 1472 & 1913 \\
zh & 681 & 1150 & 1564 & 1929 & 2286 \\
af & 455 & 1174 & 1966 & 2852 & 3803 \\
nl & 363 & 901 & 1432 & 2040 & 2680 \\
fr & 269 & 615 & 988 & 1376 & 1786 \\
pt & 310 & 698 & 1072 & 1449 & 1844 \\
ru & 394 & 717 & 985 & 1248 & 1546 \\
ko & 556 & 830 & 1054 & 1264 & 1484 \\
hi & 781 & 1456 & 2071 & 2644 & 3199 \\
tr & 476 & 1031 & 1681 & 2307 & 2916 \\
pl & 378 & 900 & 1455 & 2065 & 2744 \\
sv & 420 & 990 & 1614 & 2358 & 3089 \\
da & 431 & 1055 & 1771 & 2557 & 3344 \\
no & 433 & 1041 & 1743 & 2499 & 3268 \\
en & 61 & 88 & 116 & 140 & 151 \\
\bottomrule
\end{tabular}
\caption{Number of language-specific neurons at different top-\textit{k} thresholds in \textbf{Llama-3.1}.}
\label{tab:neuron_counts_llama}
\end{table}

\begin{table}[h!]
\small
\centering
\begin{tabular}{lccccc}
\toprule
\textbf{Lang.} & \textbf{1\%} & \textbf{2\%} & \textbf{3\%} & \textbf{4\%} & \textbf{5\%} \\
\midrule
bo & 1153 & 1941 & 2547 & 3059 & 3554 \\
mt & 250 & 717 & 1283 & 1913 & 2552 \\
it & 155 & 343 & 539 & 766 & 969 \\
es & 154 & 343 & 528 & 730 & 907 \\
de & 130 & 291 & 460 & 656 & 859 \\
ja & 384 & 770 & 1113 & 1446 & 1808 \\
ar & 275 & 499 & 703 & 912 & 1131 \\
zh & 346 & 717 & 1076 & 1380 & 1723 \\
af & 219 & 567 & 1012 & 1549 & 2115 \\
nl & 184 & 458 & 791 & 1207 & 1646 \\
fr & 156 & 323 & 475 & 650 & 812 \\
pt & 175 & 388 & 589 & 799 & 1001 \\
ru & 213 & 387 & 520 & 653 & 764 \\
ko & 335 & 623 & 852 & 1079 & 1299 \\
hi & 310 & 619 & 924 & 1193 & 1443 \\
tr & 213 & 586 & 1027 & 1510 & 1982 \\
pl & 186 & 436 & 752 & 1106 & 1472 \\
sv & 160 & 451 & 807 & 1211 & 1637 \\
da & 171 & 486 & 886 & 1335 & 1812 \\
no & 148 & 464 & 836 & 1237 & 1643 \\
en & 32 & 55 & 78 & 100 & 117 \\
\bottomrule
\end{tabular}
\caption{Number of language-specific neurons at different top-\textit{k} thresholds in \textbf{Mistral-Nemo-Base-2407}.}
\label{tab:neuron_counts_nemo}
\end{table}

\begin{table}[h!]
\small
\centering
\begin{tabular}{lccccc}
\toprule
\textbf{Lang.} & \textbf{1\%} & \textbf{2\%} & \textbf{3\%} & \textbf{4\%} & \textbf{5\%} \\
\midrule
bo & 1636 & 3343 & 4947 & 6551 & 8085 \\
mt & 433  & 1169 & 2108 & 3079 & 4150 \\
it & 231  & 591  & 981  & 1393 & 1850 \\
es & 277  & 621  & 1016 & 1388 & 1745 \\
de & 211  & 509  & 874  & 1245 & 1674 \\
ja & 636  & 1171 & 1580 & 1983 & 2393 \\
ar & 383  & 662  & 962  & 1251 & 1566 \\
zh & 702  & 1348 & 1933 & 2512 & 3030 \\
af & 296  & 781  & 1440 & 2181 & 2925 \\
nl & 181  & 485  & 859  & 1243 & 1633 \\
fr & 211  & 489  & 842  & 1229 & 1636 \\
pt & 280  & 621  & 1007 & 1408 & 1811 \\
ru & 381  & 708  & 1052 & 1356 & 1671 \\
ko & 597  & 1051 & 1405 & 1763 & 2110 \\
hi & 459  & 880  & 1269 & 1668 & 2063 \\
tr & 322  & 734  & 1126 & 1512 & 1927 \\
pl & 260  & 633  & 1003 & 1364 & 1746 \\
sv & 282  & 837  & 1595 & 2437 & 3331 \\
da & 303  & 886  & 1688 & 2628 & 3569 \\
no & 290  & 866  & 1672 & 2576 & 3496 \\
en & 55   & 80   & 115  & 141  & 169 \\
\bottomrule
\end{tabular}
\caption{Number of language-specific neurons at different top-\textit{k} thresholds in \textbf{Aya-Expanse-8B}.}
\label{tab:neuron_counts_aya}
\end{table}

\begin{table}[h!]
\small
\centering
\begin{tabular}{lccccc}
\toprule
\textbf{Lang.} & \textbf{1\%} & \textbf{2\%} & \textbf{3\%} & \textbf{4\%} & \textbf{5\%} \\
\midrule
bo & 1107 & 2241 & 3314 & 4403 & 5541 \\
mt & 568  & 1431 & 2403 & 3427 & 4461 \\
it & 353  & 758  & 1192 & 1629 & 2076 \\
es & 344  & 711  & 1100 & 1483 & 1805 \\
de & 240  & 631  & 1105 & 1557 & 1986 \\
ja & 676  & 1249 & 1703 & 2152 & 2563 \\
ar & 361  & 643  & 902  & 1163 & 1436 \\
zh & 656  & 1306 & 1920 & 2507 & 3032 \\
af & 364  & 978  & 1672 & 2368 & 3021 \\
nl & 235  & 607  & 990  & 1371 & 1691 \\
fr & 280  & 593  & 943  & 1332 & 1717 \\
pt & 362  & 709  & 1134 & 1548 & 1940 \\
ru & 355  & 757  & 1138 & 1514 & 1860 \\
ko & 549  & 1000 & 1383 & 1746 & 2091 \\
hi & 498  & 956  & 1296 & 1604 & 1899 \\
tr & 331  & 728  & 1138 & 1524 & 1902 \\
pl & 276  & 656  & 1036 & 1467 & 1823 \\
sv & 322  & 897  & 1585 & 2273 & 2925 \\
da & 343  & 980  & 1739 & 2488 & 3162 \\
no & 329  & 968  & 1705 & 2448 & 3102 \\
en & 51   & 85   & 122  & 148  & 171 \\
\bottomrule
\end{tabular}
\caption{Number of language-specific neurons at different top-\textit{k} thresholds in \textbf{Aya-Expanse-32B}.}
\label{tab:neuron_counts_aya32}
\end{table}

\clearpage

\begin{figure}[t]
\section{Language Neuron Distributions}
\label{appendix:language_neurons}
    \centering
    \includegraphics[width=0.9\linewidth]{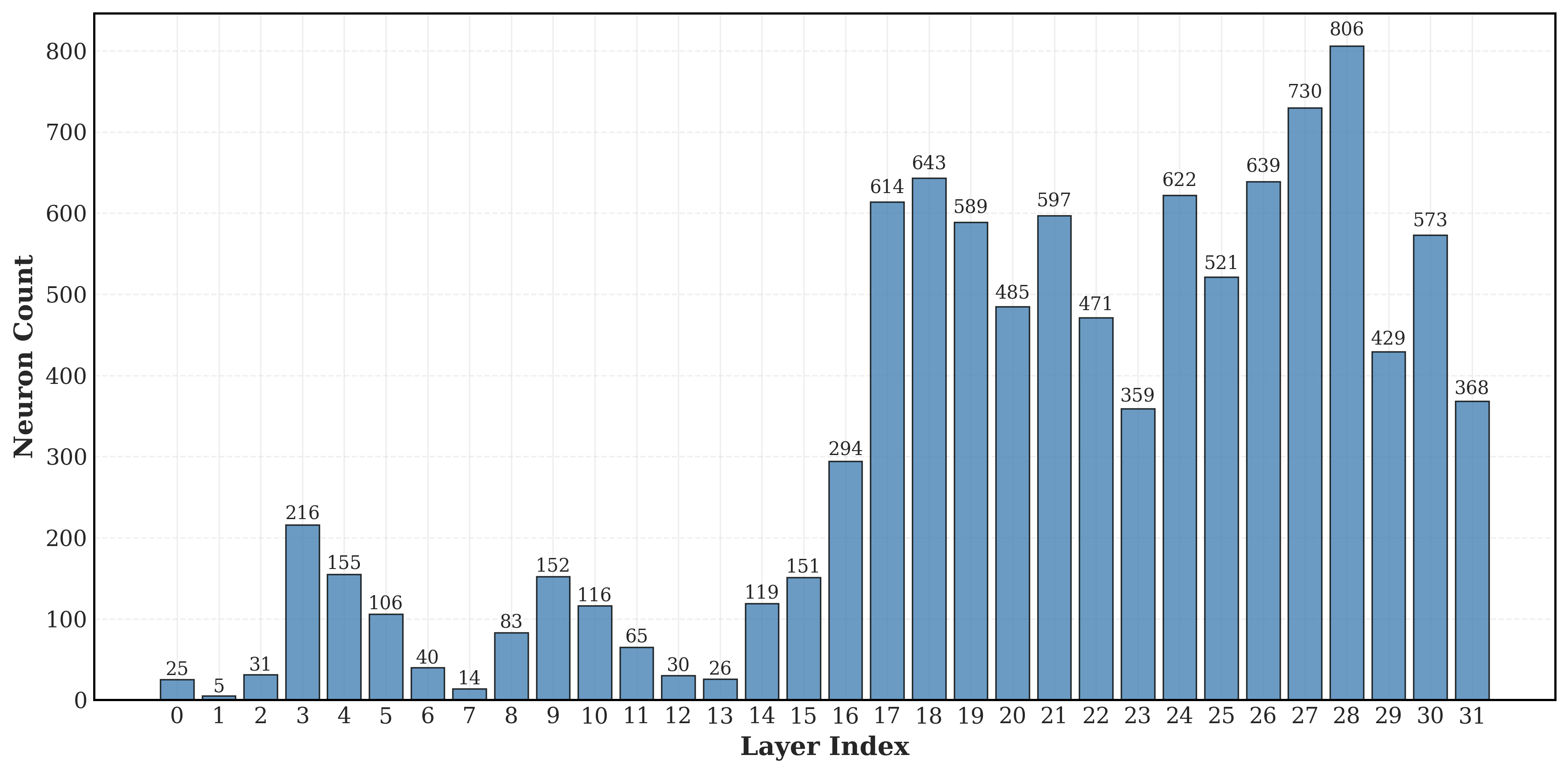}
    \caption{Distribution of identified language-specific neurons across \textbf{Llama-3.1-8B} layers for all 21 evaluated languages. The neuron distributions for individual languages are further in the Appendix.}
    \label{fig:distribution-plots-llama}
\end{figure}

\begin{figure}[h]
    \centering
    \includegraphics[width=0.9\linewidth]{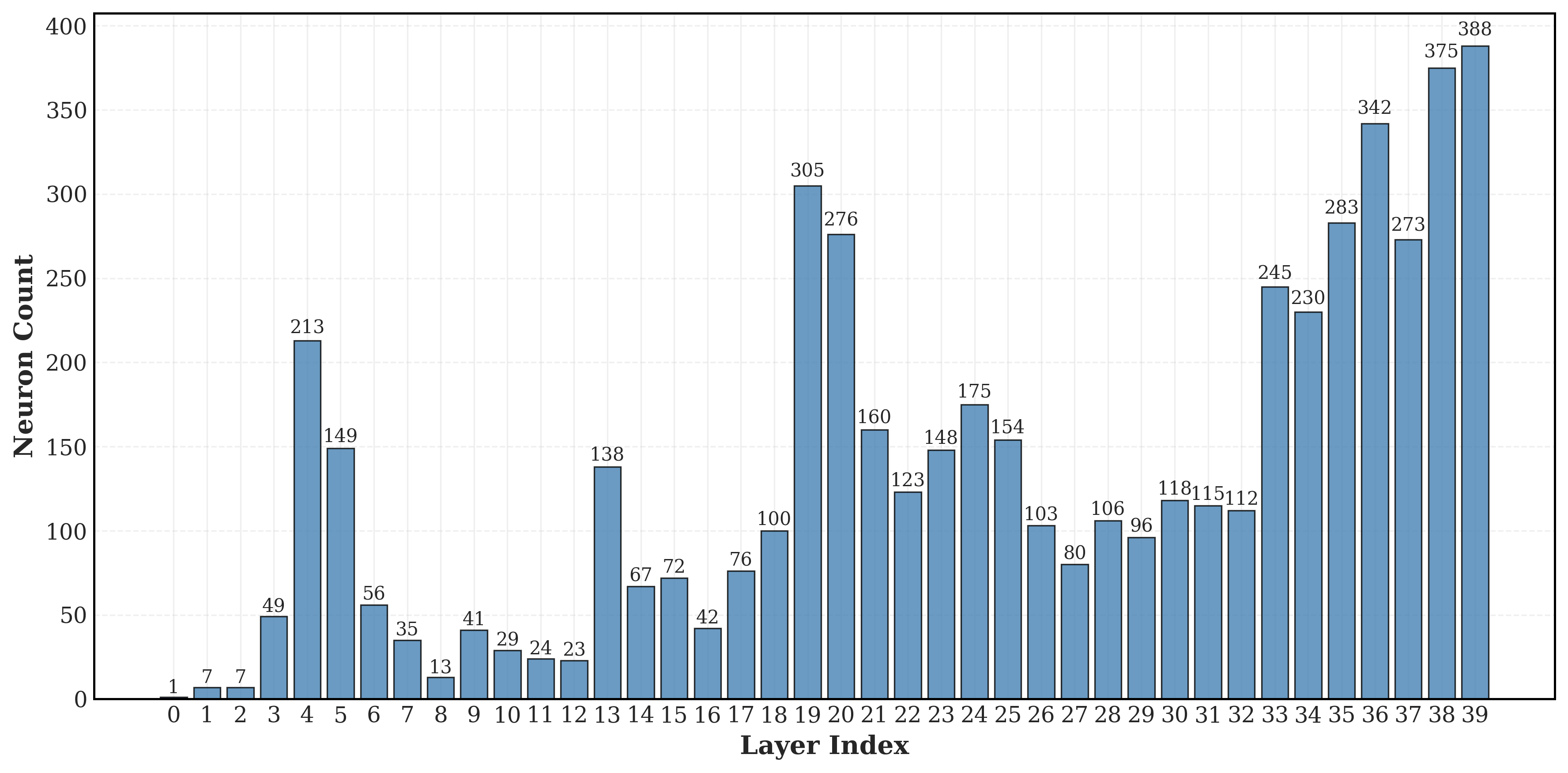}
    \caption{Distribution of identified language-specific neurons across \textbf{Mistral-Nemo} layers for all 21 evaluated languages. The neuron distributions for individual languages are further in the Appendix.}
    \label{fig:distribution-plots-nemo}
\end{figure}

\begin{figure}[h]
    \centering
    \includegraphics[width=0.9\linewidth]{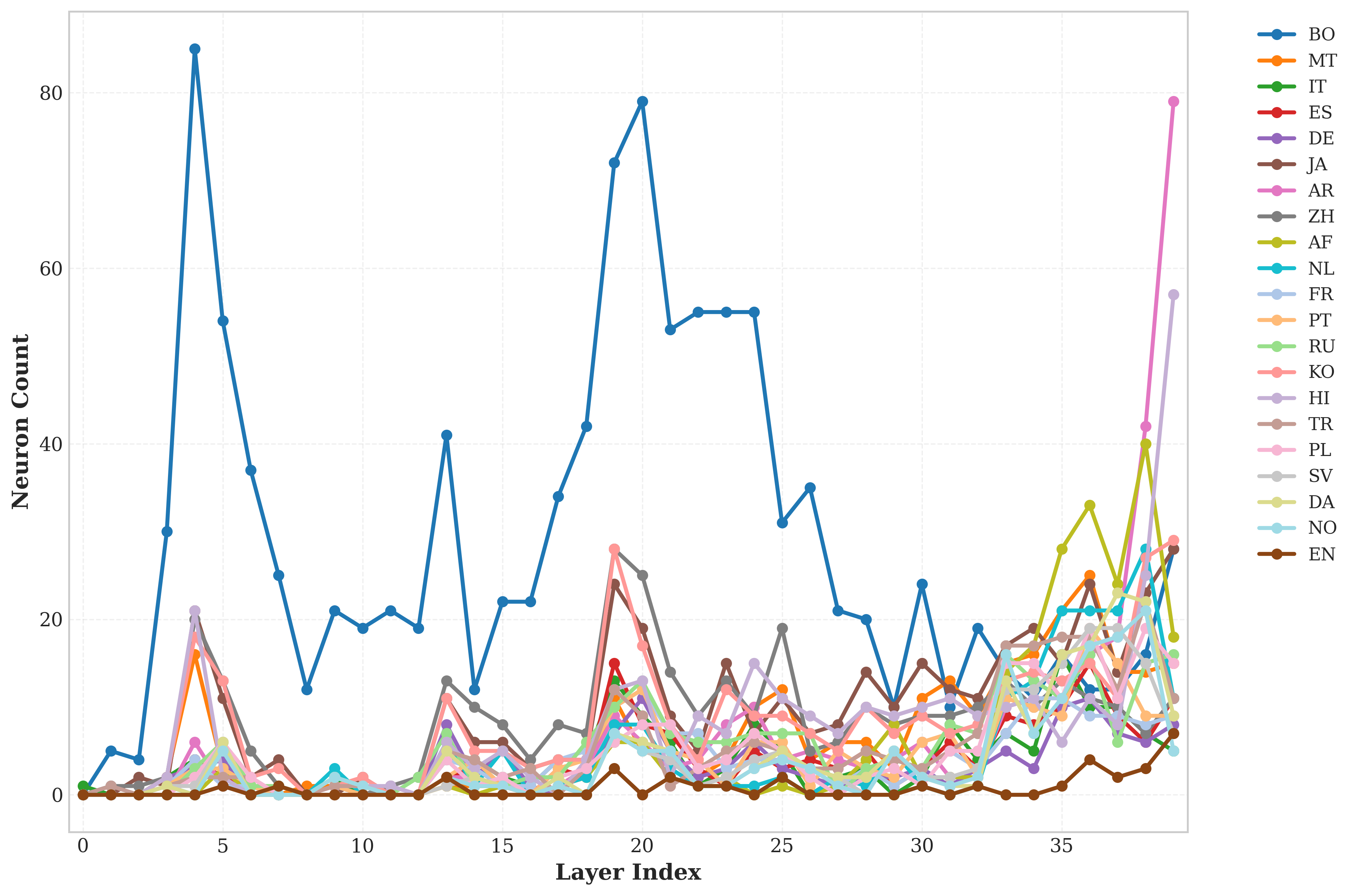}
    \caption{Distribution of \textbf{individual} language-specific neurons across \textbf{Mistral-Nemo} layers for all 21 languages.}
    \label{fig:all-langs-distributions-nemo}
\end{figure}

\begin{figure}[h]
    \centering
    \includegraphics[width=0.9\linewidth]{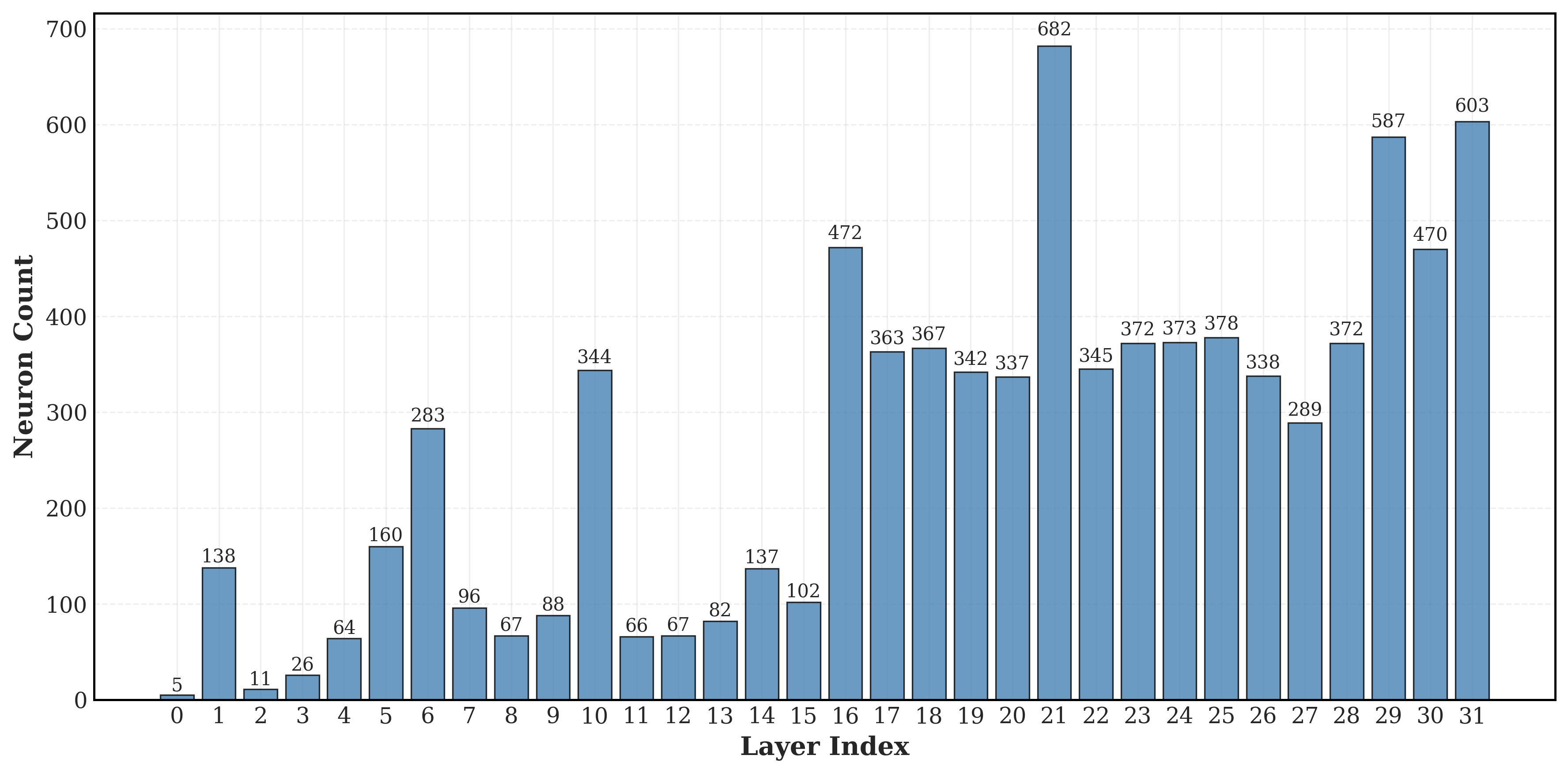}
    \caption{Distribution of identified language-specific neurons across \textbf{Aya-Expanse-8B} layers for all 21 evaluated languages.}
    \label{fig:distribution-plots-aya-8}
\end{figure}

\begin{figure}[h]
    \centering
    \includegraphics[width=0.9\linewidth]{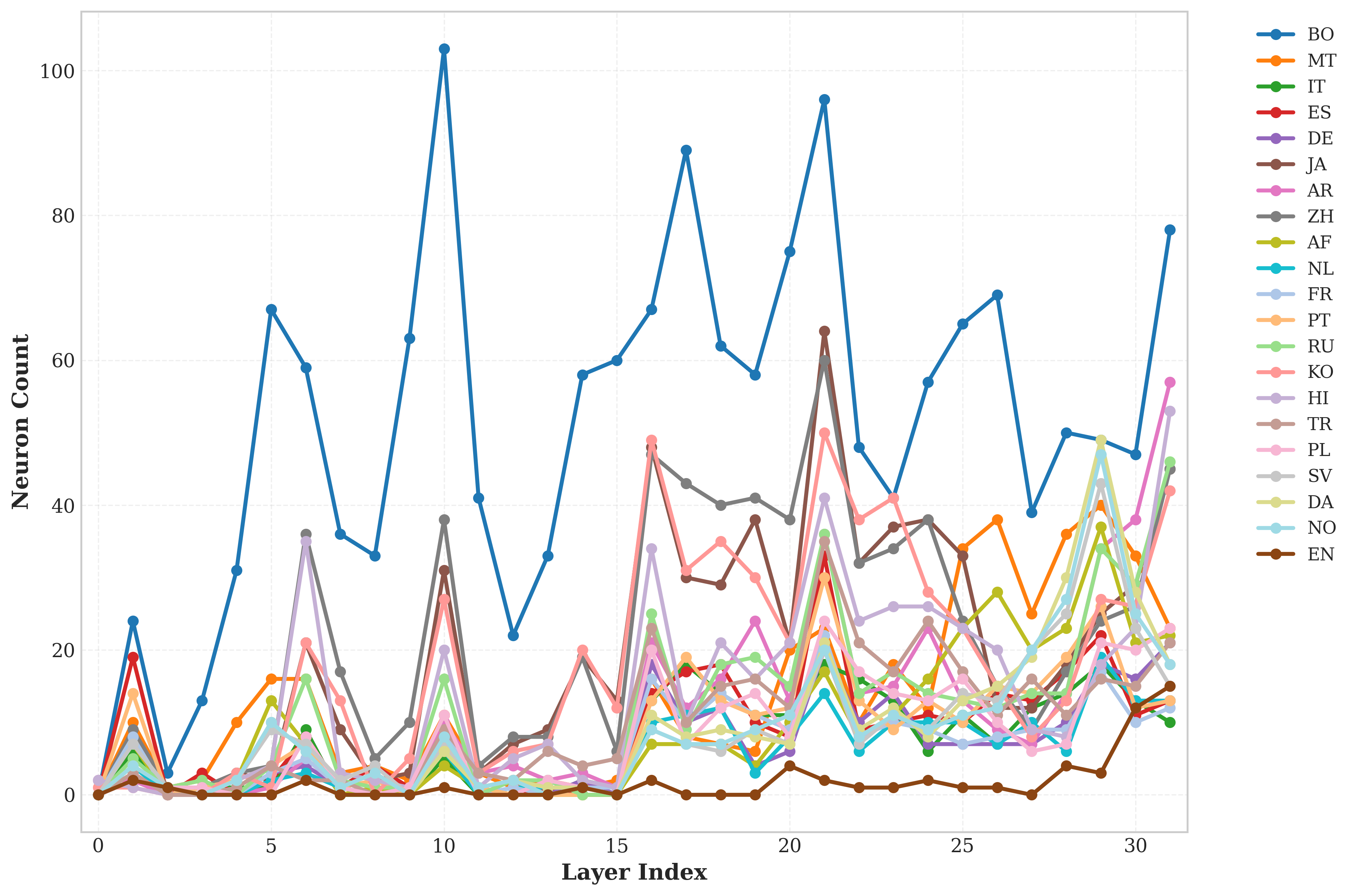}
    \caption{Distribution of \textbf{individual} language-specific neurons across \textbf{Aya-Expanse-8B} layers for all 21 languages.}
    \label{fig:all-langs-distributions-aya-8}
\end{figure}

\begin{figure}[h]
    \centering
    \includegraphics[width=0.9\linewidth]{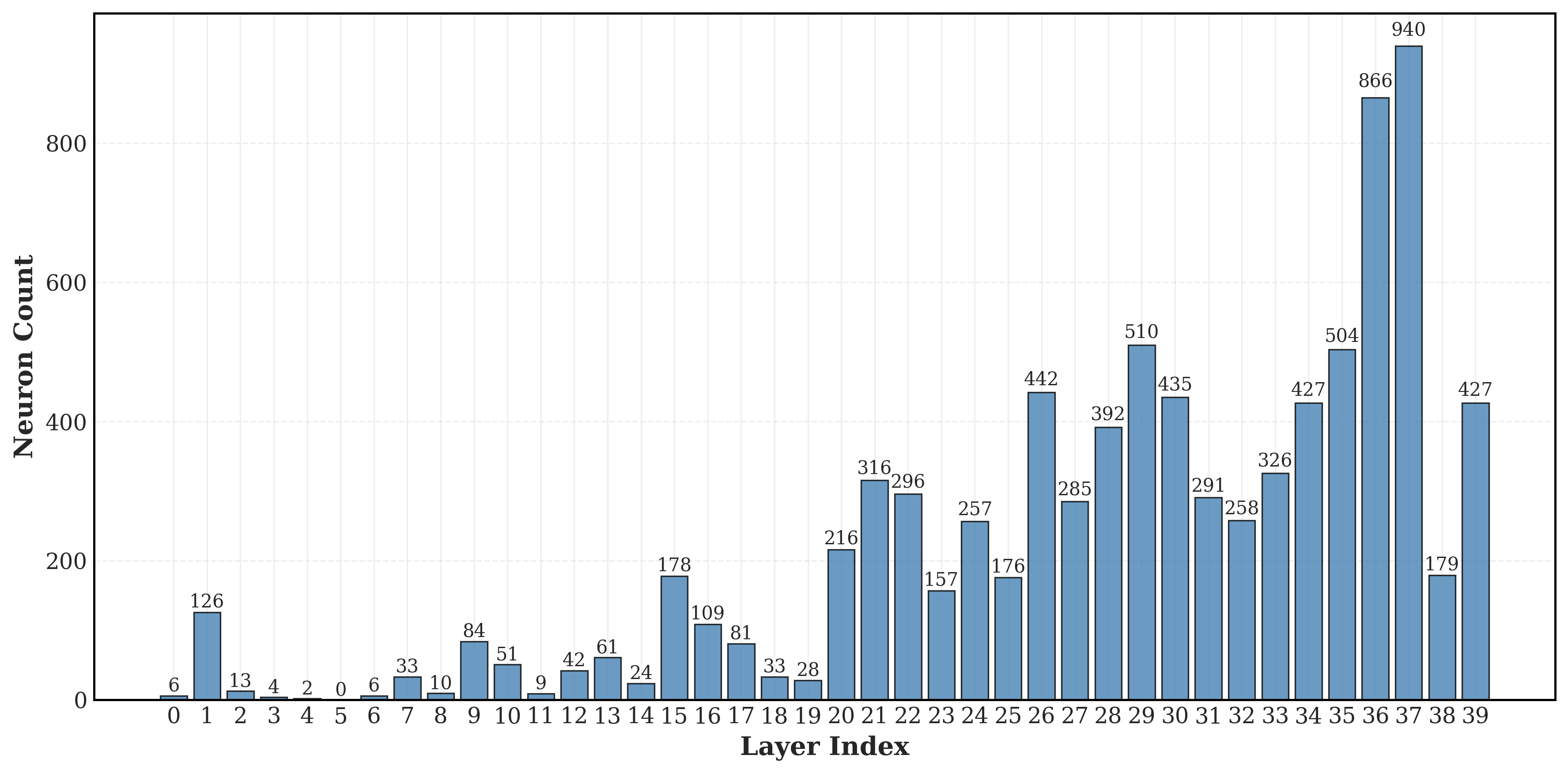}
    \caption{Distribution of identified language-specific neurons across \textbf{Aya-Expanse-32B} layers for all 21 evaluated languages. The neuron distributions for individual languages are further in the Appendix.}
    \label{fig:distribution-plots-aya-32}
\end{figure}

\begin{figure}[h]
    \centering
    \includegraphics[width=0.9\linewidth]{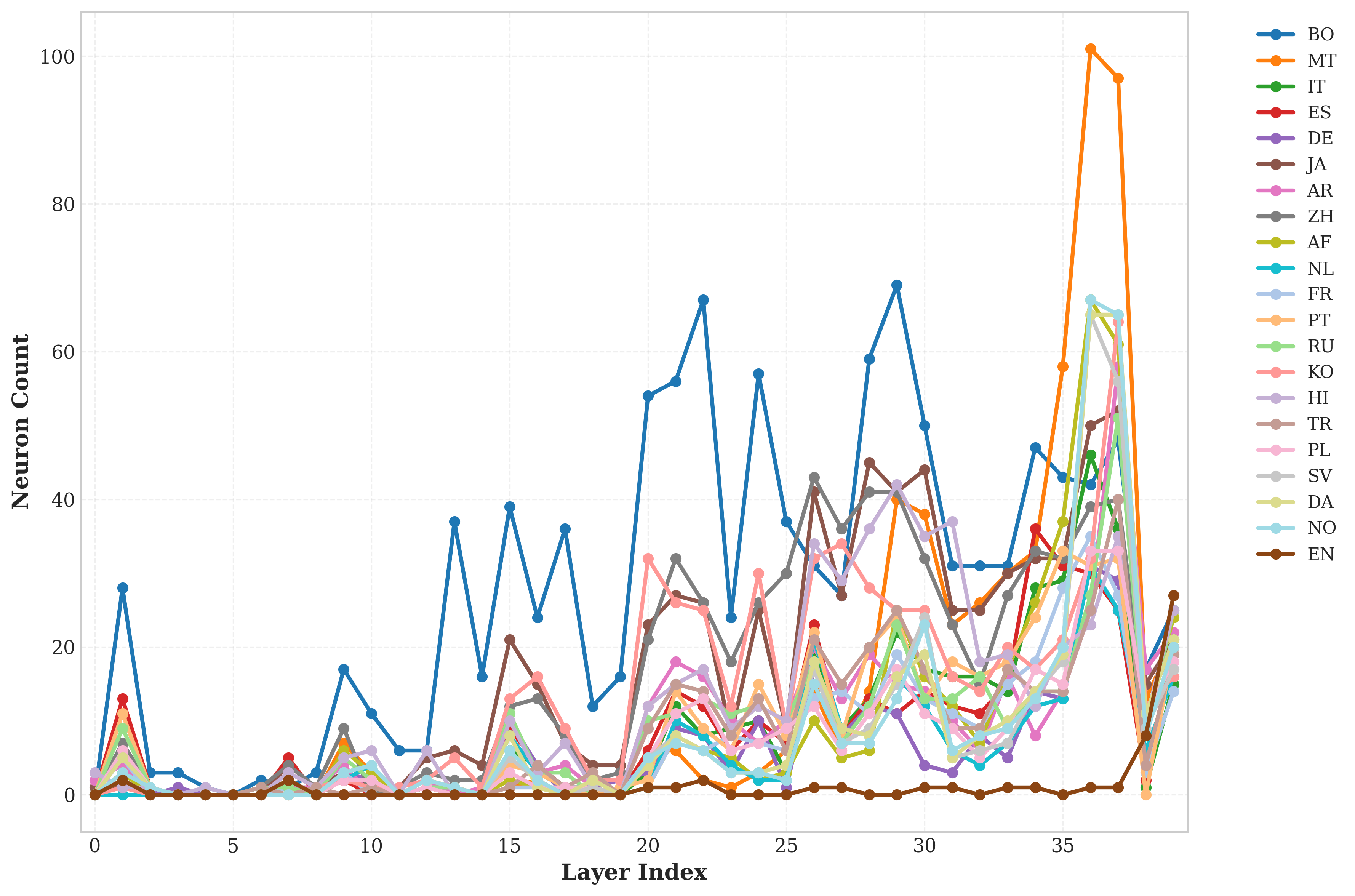}
    \caption{Distribution of \textbf{individual} language-specific neurons across \textbf{Aya-Expanse-32B} layers for all 21 languages.}
    \label{fig:all-langs-distributions-aya-32}
\end{figure}

\begin{figure*}[hbt]
\centering
\begin{subfigure}[t]{0.75\linewidth}
    \centering
    \includegraphics[width=\linewidth]{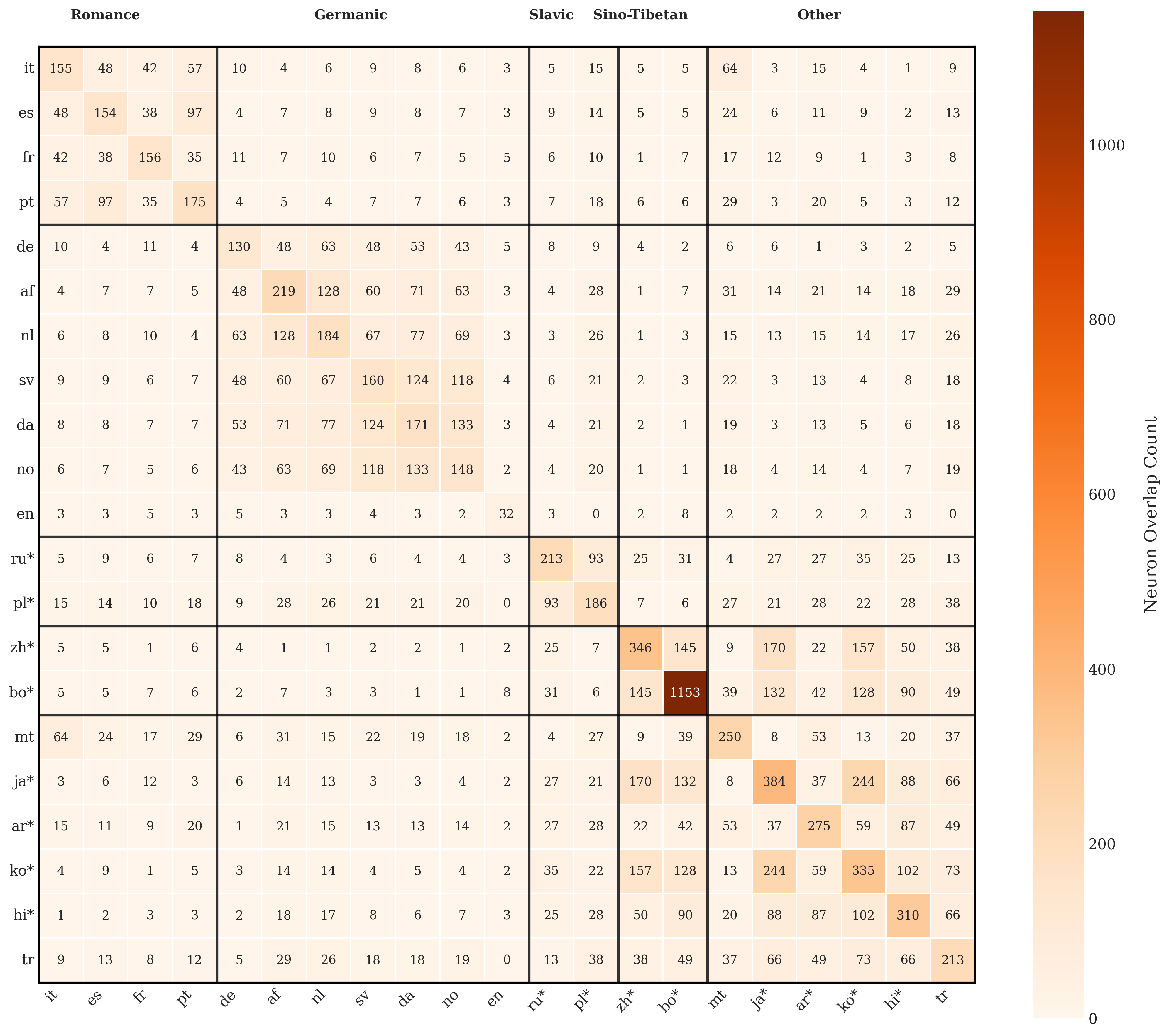}
\end{subfigure}

\caption{Overlap of language-specific neurons between individual languages and language families in \textbf{Mistral-Nemo} when considering top 1\% of neurons as potentially language-specific. Diagonal values indicate the number of language-specific neurons for each language; off-diagonal values indicate the number of overlapping neurons. Asterisks denote languages with non-Latin scripts.}
\label{fig:overlap_nemo}
\end{figure*}

\begin{figure*}[hbt]
\centering
\begin{subfigure}[t]{0.75\linewidth}
    \centering
    \includegraphics[width=\linewidth]{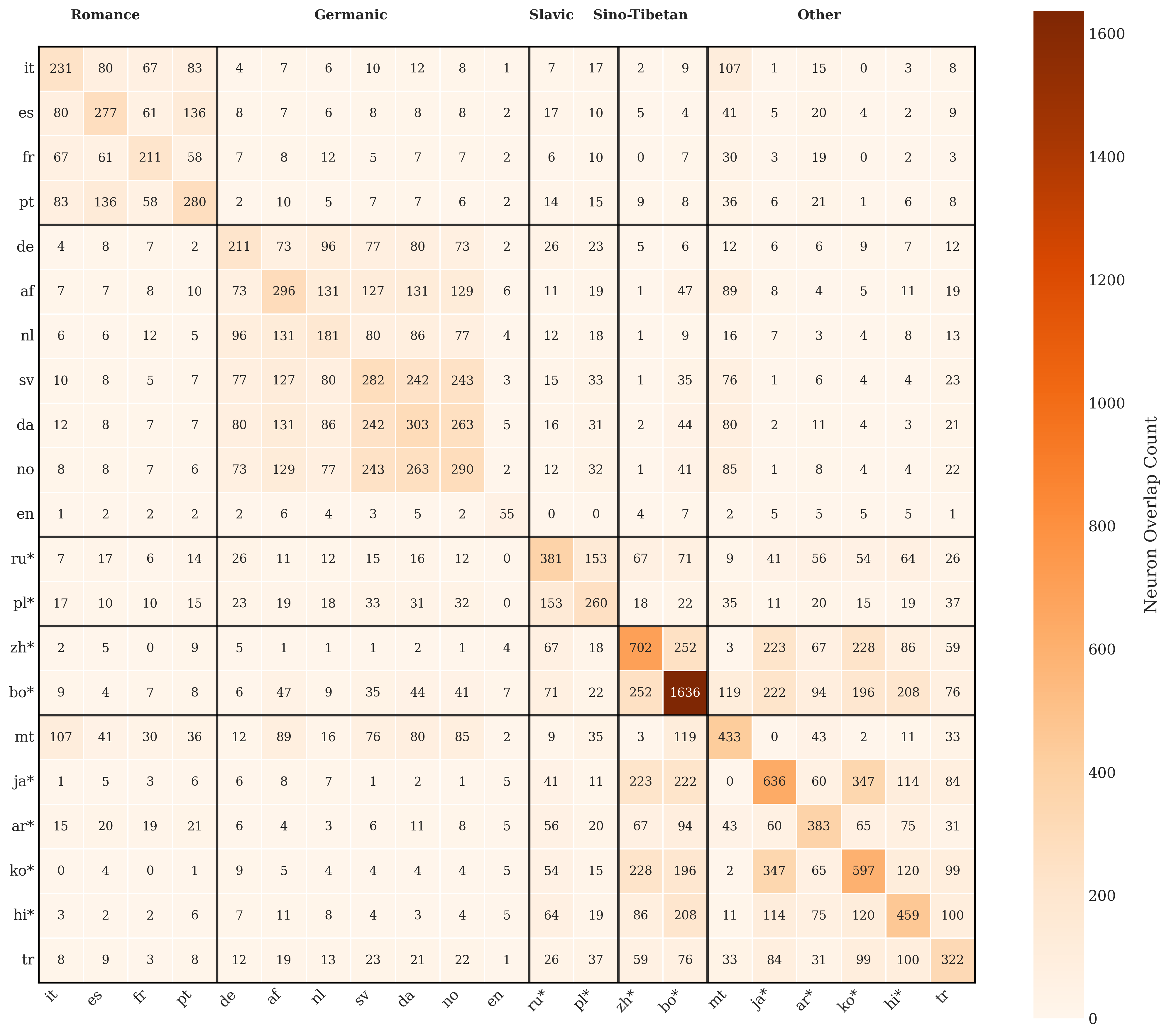}
    \label{fig:overlap_aya-8}
\end{subfigure}
\hfill
\begin{subfigure}[t]{0.75\linewidth}
    \centering
    \includegraphics[width=\linewidth]{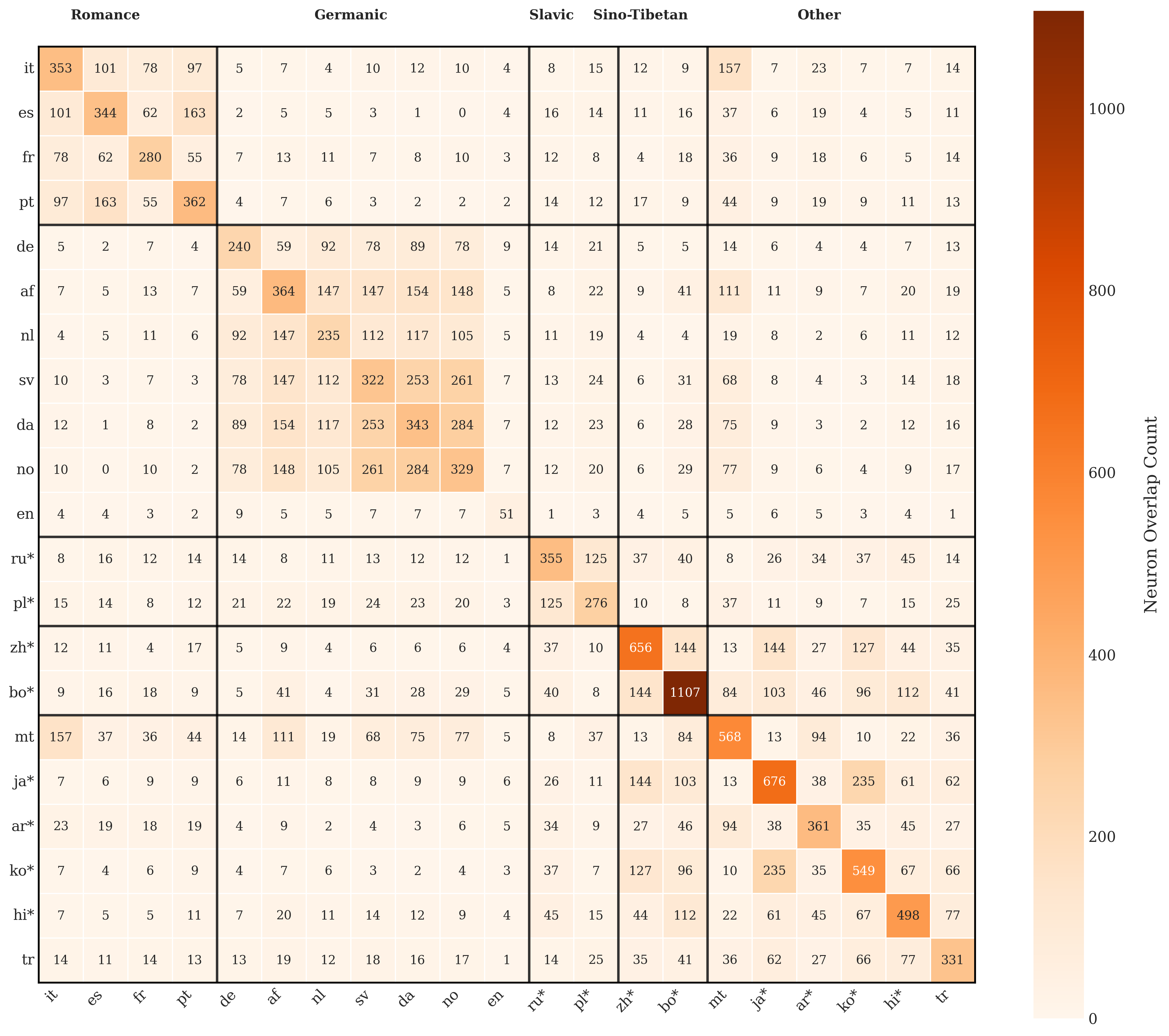}
    \label{fig:overlap_aya-32}
\end{subfigure}

\caption{Overlap of language-specific neurons between individual languages and language families in \textbf{Aya-Expanse-8B and 32B} when considering top 1\% of neurons as potentially language-specific. Diagonal values indicate the number of language-specific neurons for each language; off-diagonal values indicate the number of overlapping neurons. Asterisks denote languages with non-Latin scripts.}
\label{fig:overlap_aya}
\end{figure*}

\begin{figure*}
    \centering
    \includegraphics[width=1\linewidth]{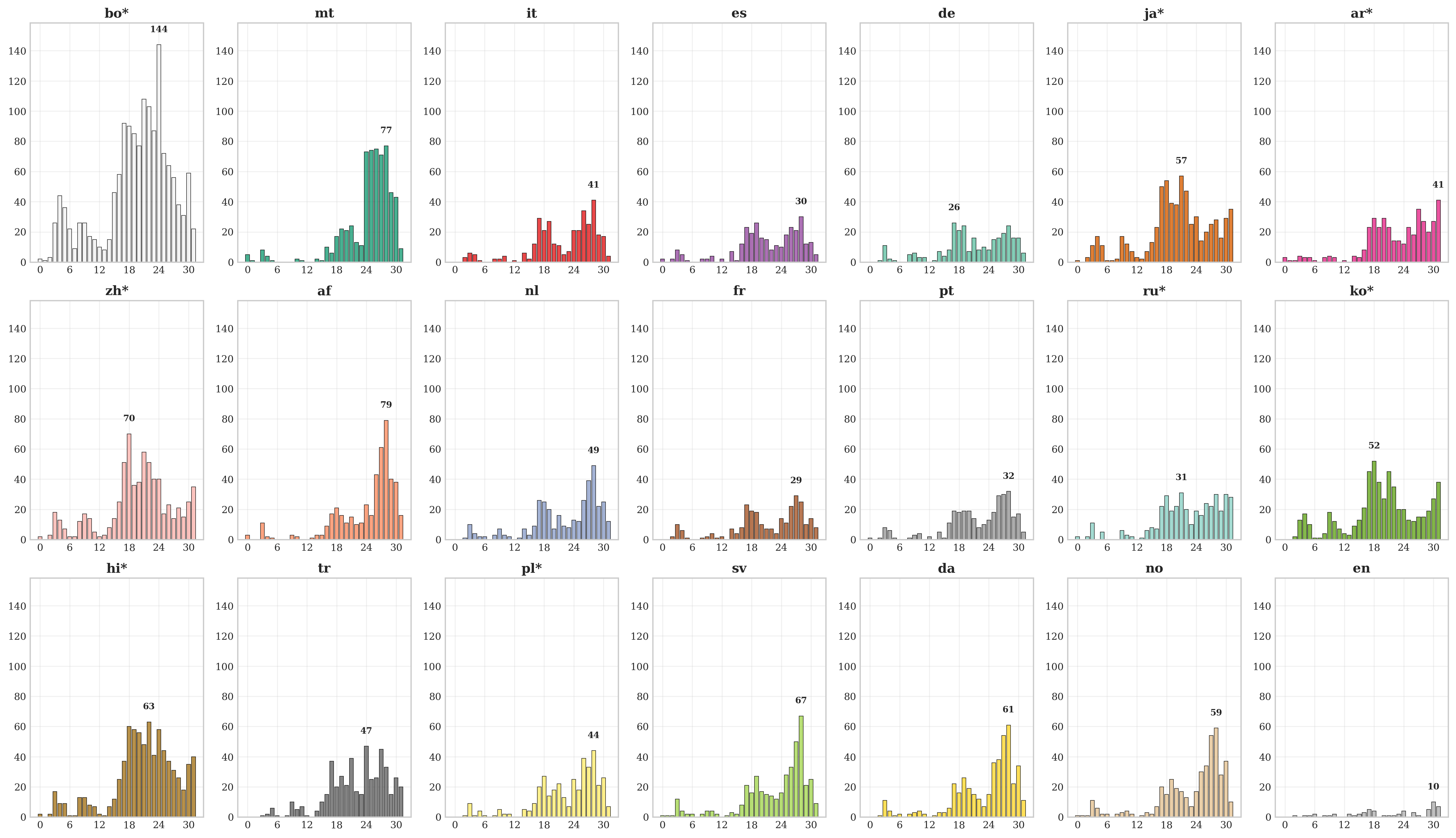}
    \caption{Distribution of \textbf{individual} language-specific neurons across \textbf{Llama-3.1-8B} layers for all 21 languages.}
    \label{fig:individual-lang-distribution-llama}
\end{figure*} 

\begin{figure*}
    \centering
    \includegraphics[width=1\linewidth]{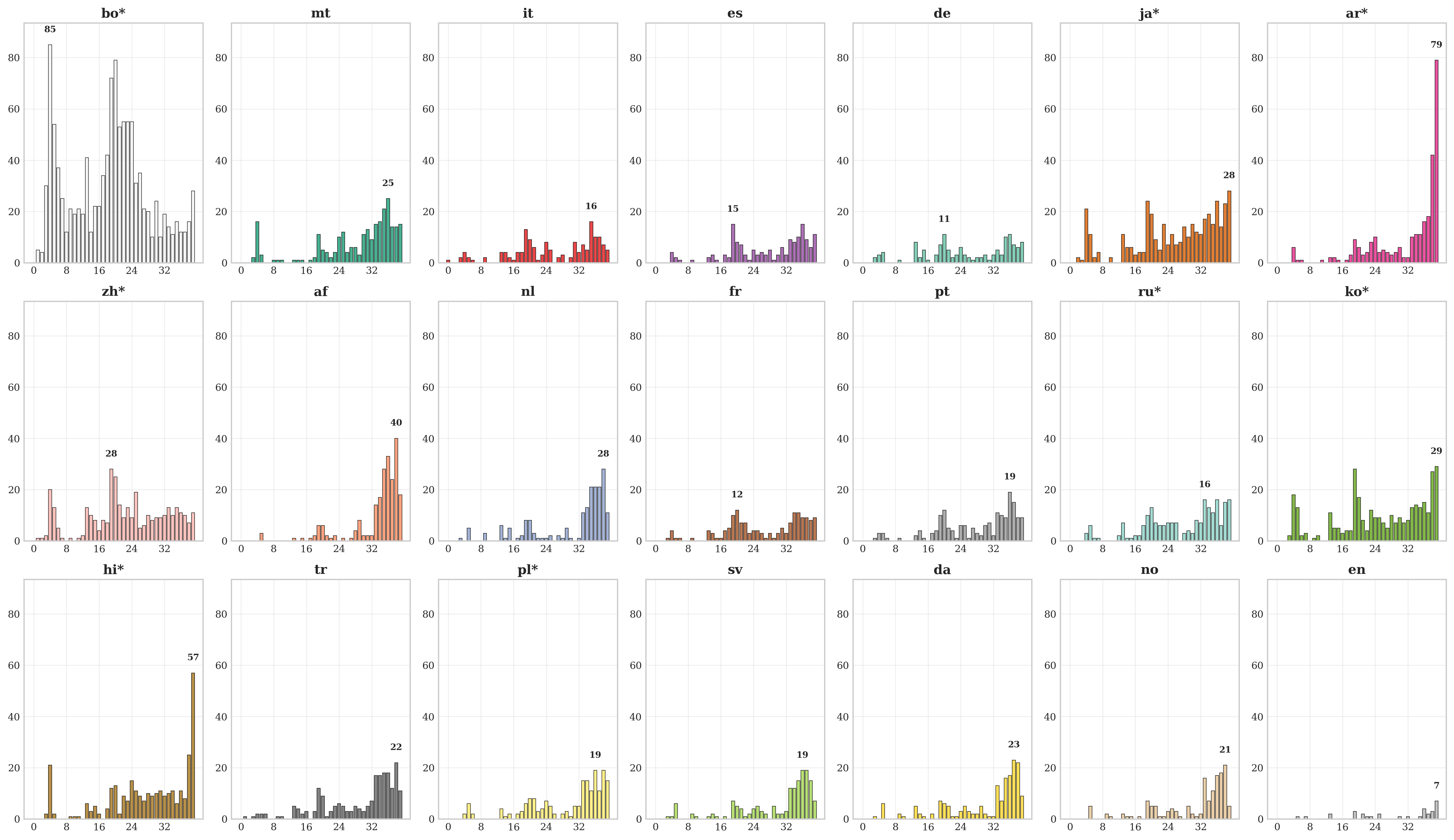}
    \caption{Distribution of \textbf{individual} language-specific neurons across \textbf{Mistral-Nemo} layers for all 21 languages.}
    \label{fig:individual-lang-distribution-nemo}
\end{figure*}

\begin{figure*}
    \centering
    \includegraphics[width=1\linewidth]{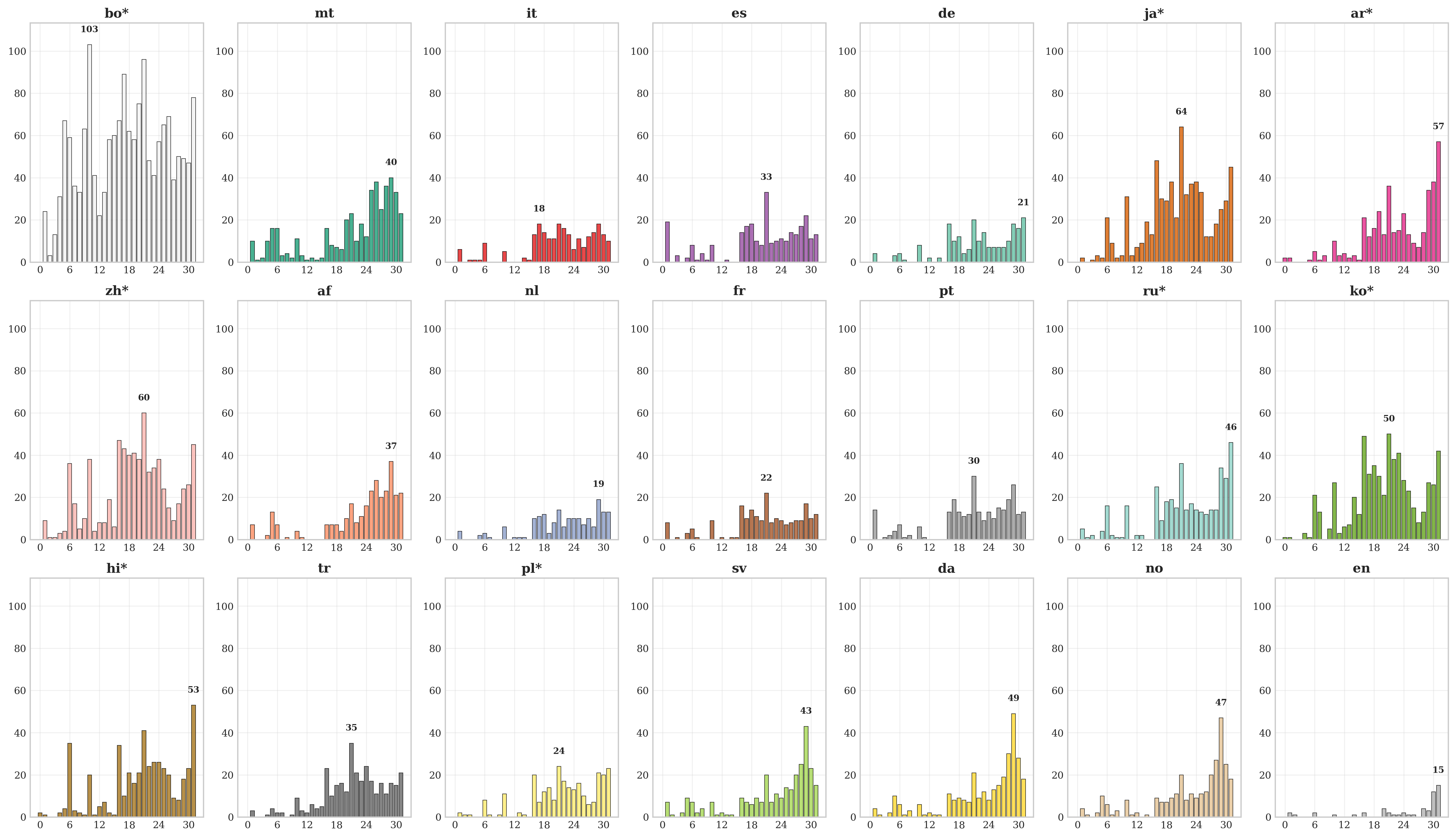}
    \caption{Distribution of \textbf{individual} language-specific neurons across \textbf{Aya-Expanse--8B} layers for all 21 languages.}
    \label{fig:individual-lang-distribution-aya-8}
\end{figure*} 

\begin{figure*}
    \centering
    \includegraphics[width=1\linewidth]{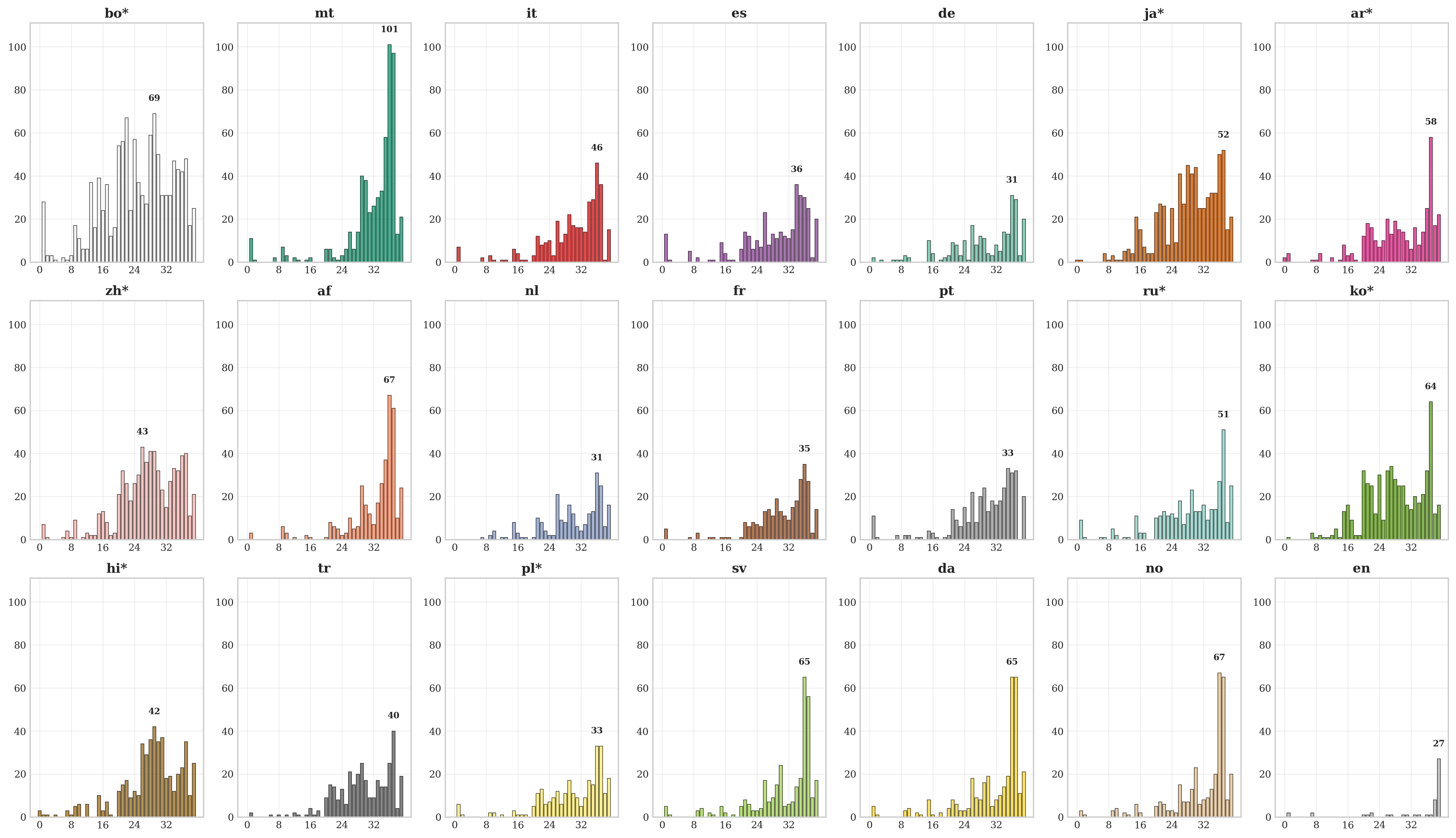}
    \caption{Distribution of \textbf{individual} language-specific neurons across \textbf{Aya-Expanse-32B} layers for all 21 languages.}
    \label{fig:individual-lang-distribution-aya-32}
\end{figure*}

\clearpage


\begin{figure*}[t]
\section{Output Analysis with Logit Lens}
\label{app:logit_lens}

    \centering
    \begin{subfigure}[t]{0.48\textwidth}
        \includegraphics[width=\linewidth]{assets/peek_target_llama.png}
        \caption{Probabilities of the target languages.}
        \label{fig:logitlens_target_llama_app}
    \end{subfigure}
    \hfill
    \begin{subfigure}[t]{0.48\textwidth}
        \includegraphics[width=\linewidth]{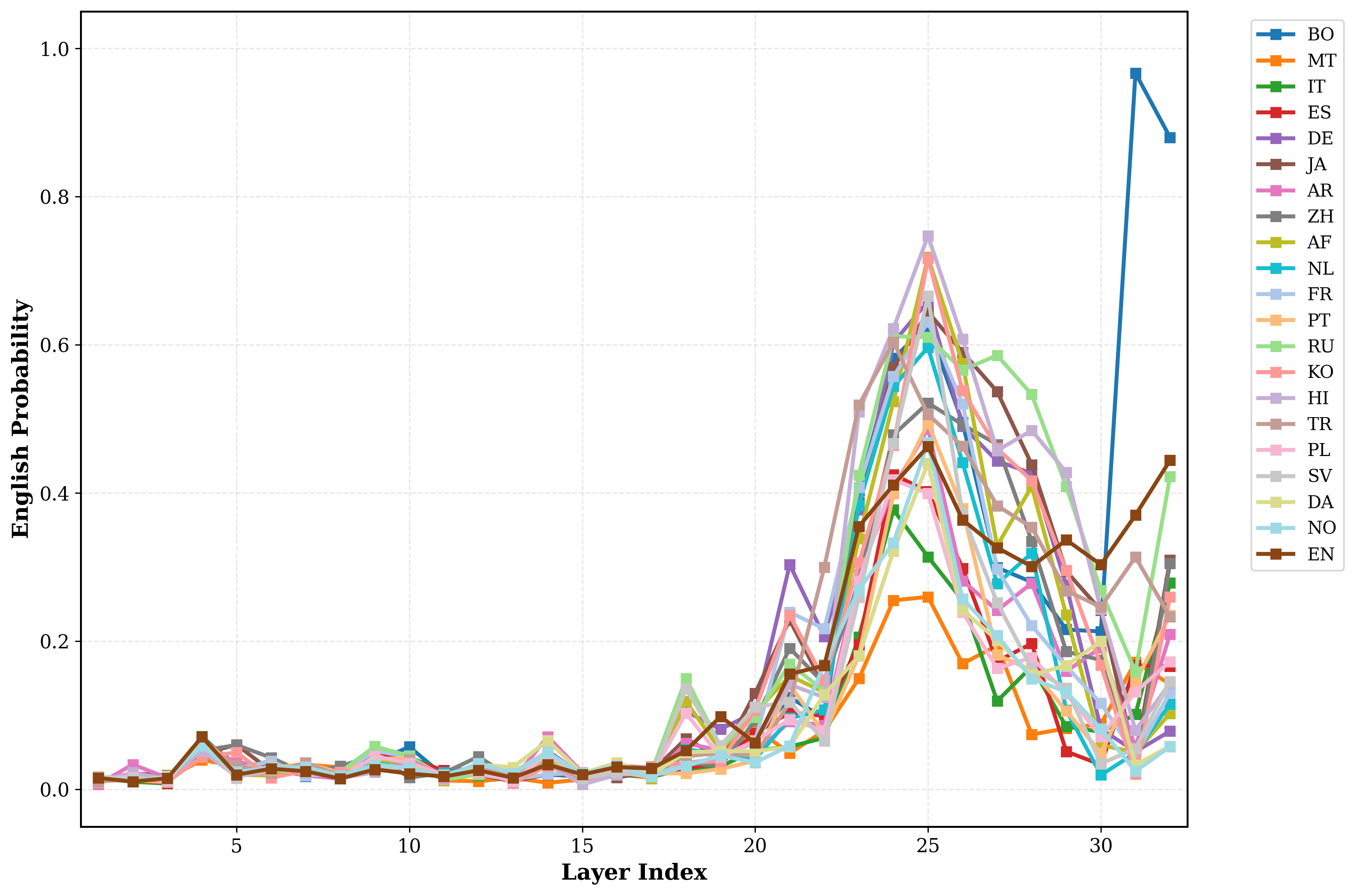}
        \caption{Probabilities of English.}
        \label{fig:logitlens_english_llama}
    \end{subfigure}
    
    \vspace{0.5em}
    
    \begin{subfigure}[t]{0.48\textwidth}
        \includegraphics[width=\linewidth]{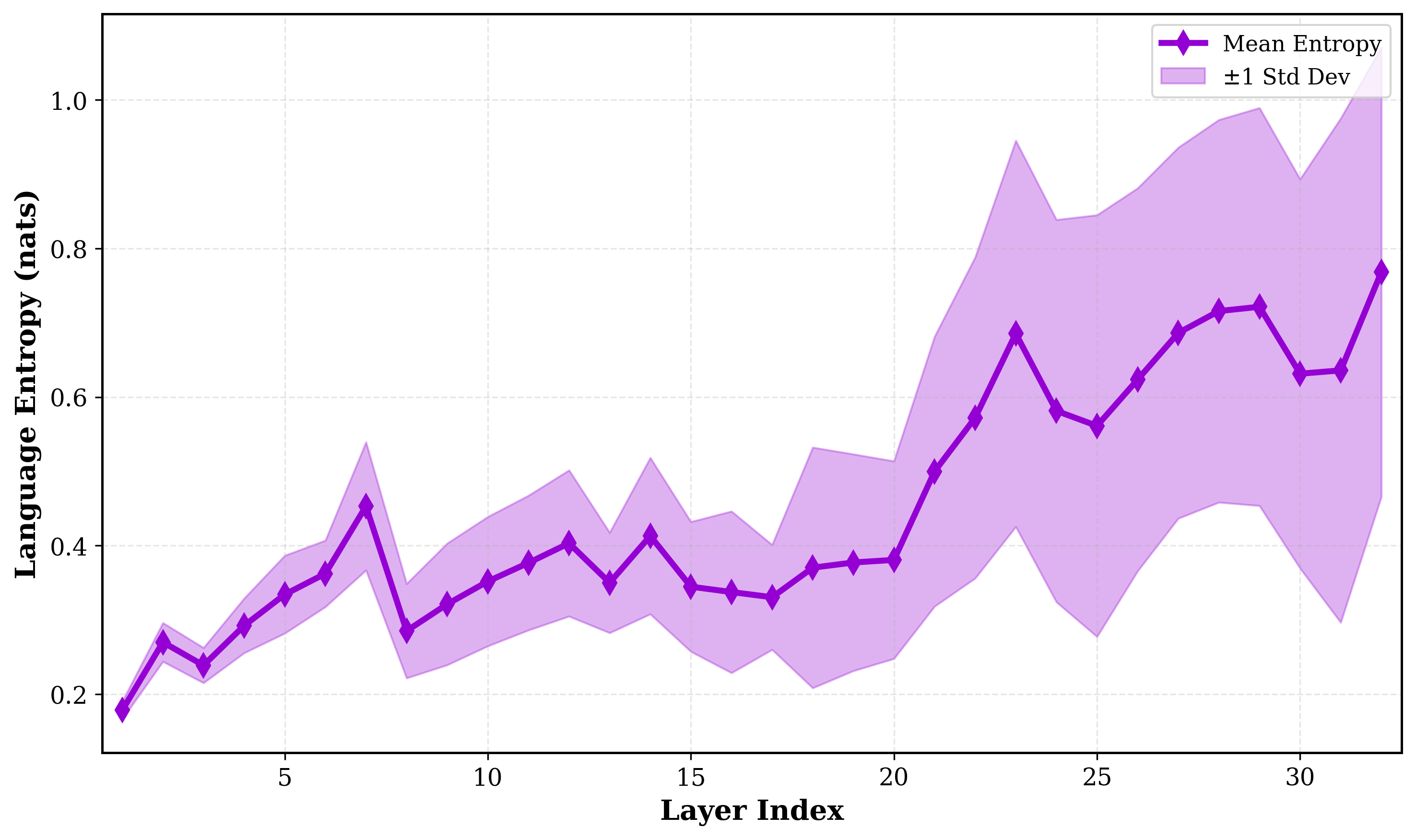}
        \caption{Evolution of target language probability.}
        \label{fig:logitlens_evolution_llama}
    \end{subfigure}
    \hfill
    \begin{subfigure}[t]{0.48\textwidth}
        \includegraphics[width=\linewidth]{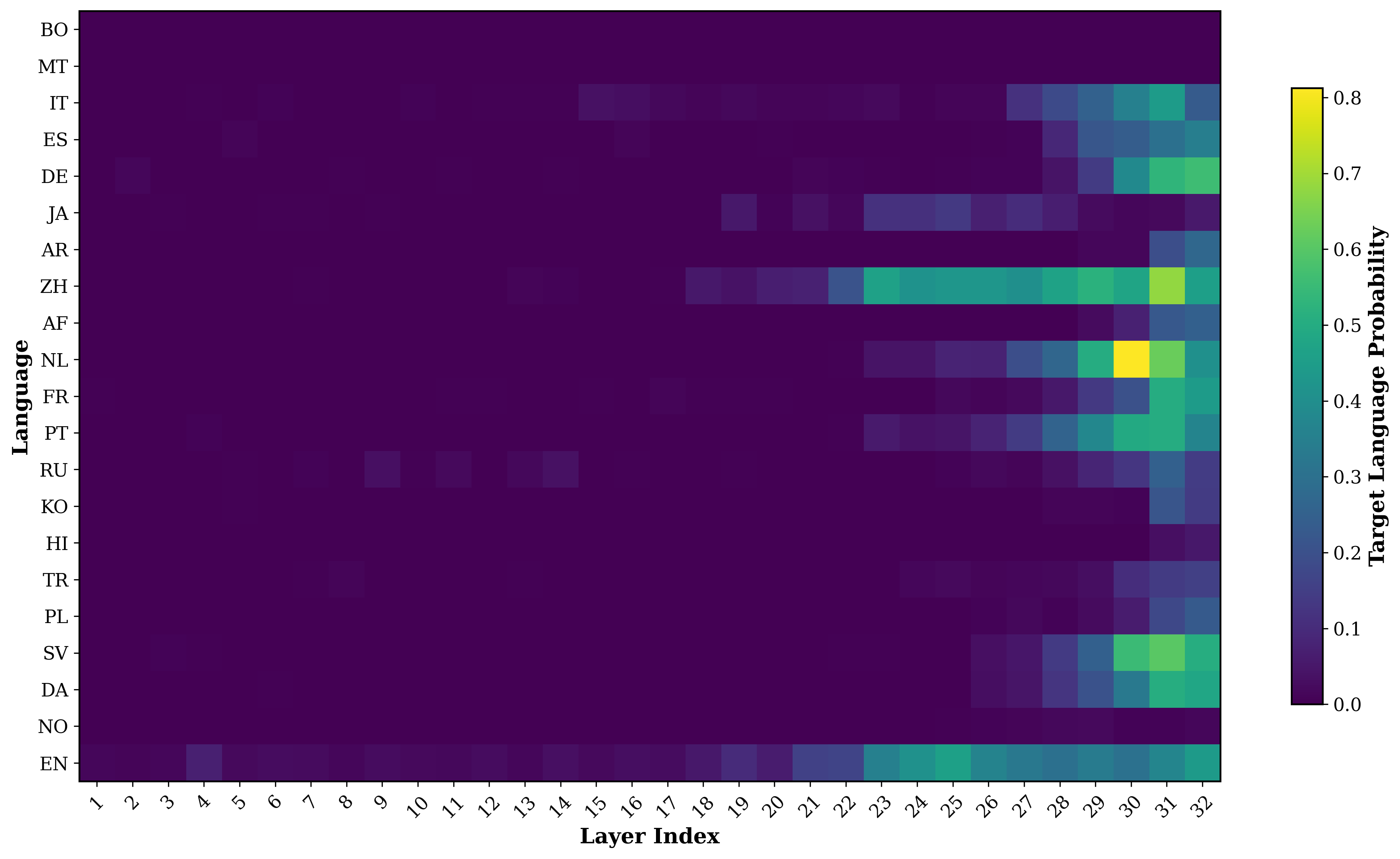}
        \caption{Heatmap of English probabilities.}
        \label{fig:logitlens_heatmap_llama}
    \end{subfigure}
    
    \caption{Peeking into language prediction across layers using the logit lens for \textbf{Llama-3.1-8B}.}
    \label{fig:logitlens_combined_llama}
\end{figure*}

\begin{figure*}[t]
    \centering
    \begin{subfigure}[t]{0.48\textwidth}
        \includegraphics[width=\linewidth]{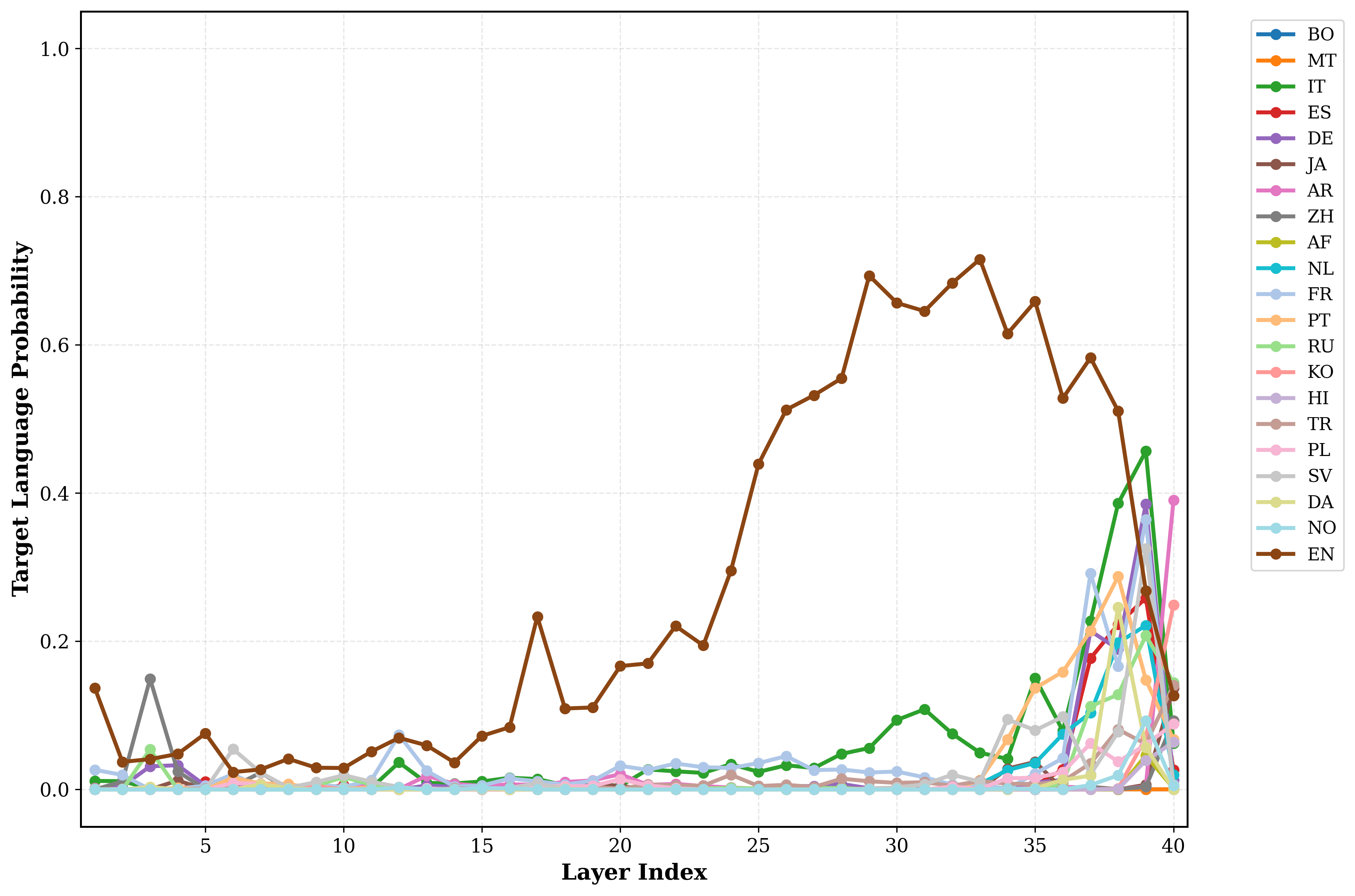}
        \caption{Probabilities of the target languages.}
        \label{fig:logitlens_target_nemo}
    \end{subfigure}
    \hfill
    \begin{subfigure}[t]{0.48\textwidth}
        \includegraphics[width=\linewidth]{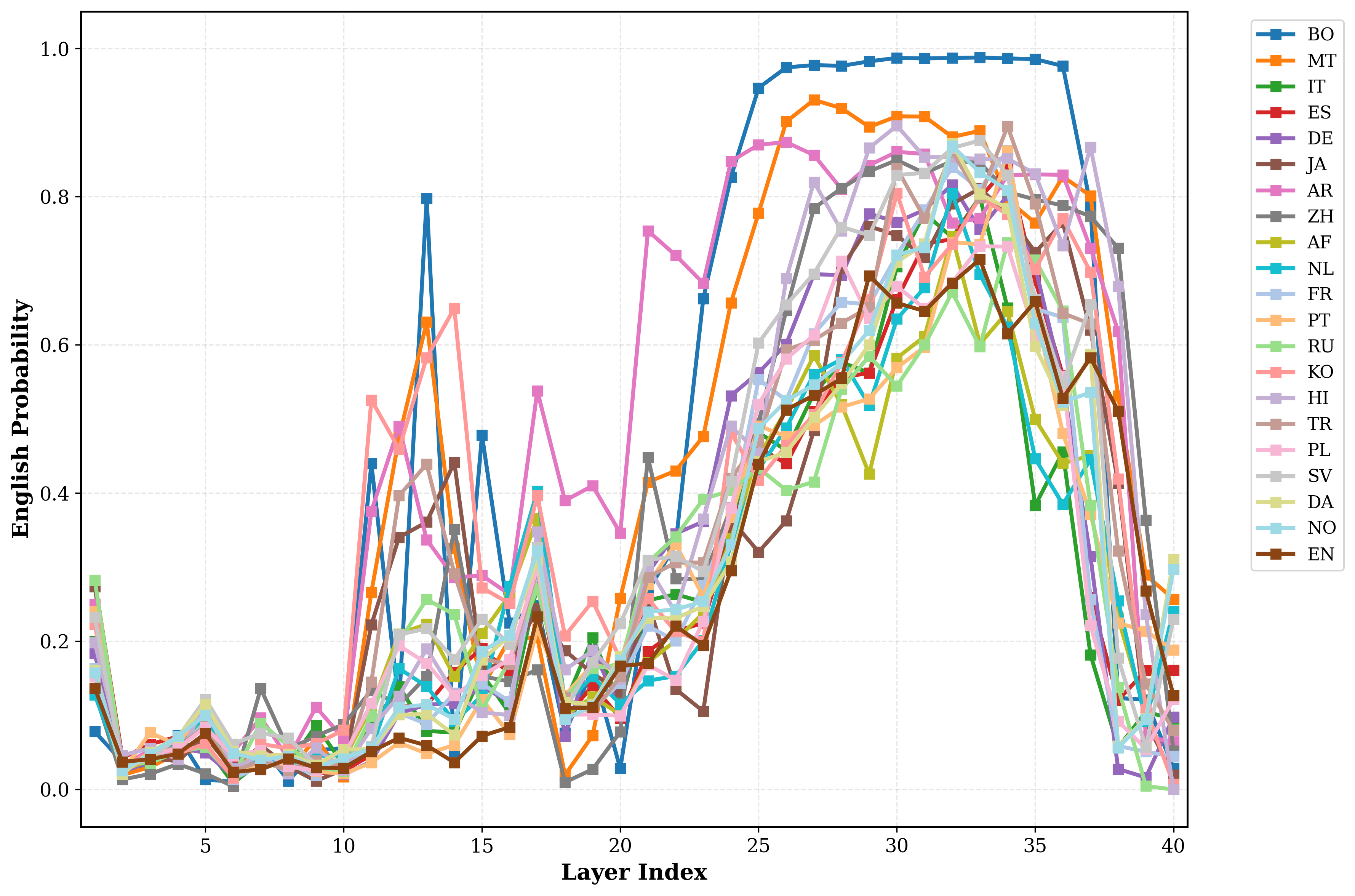}
        \caption{Probabilities of English.}
        \label{fig:logitlens_english_nemo}
    \end{subfigure}
    
    \vspace{0.5em}
    
    \begin{subfigure}[t]{0.48\textwidth}
        \includegraphics[width=\linewidth]{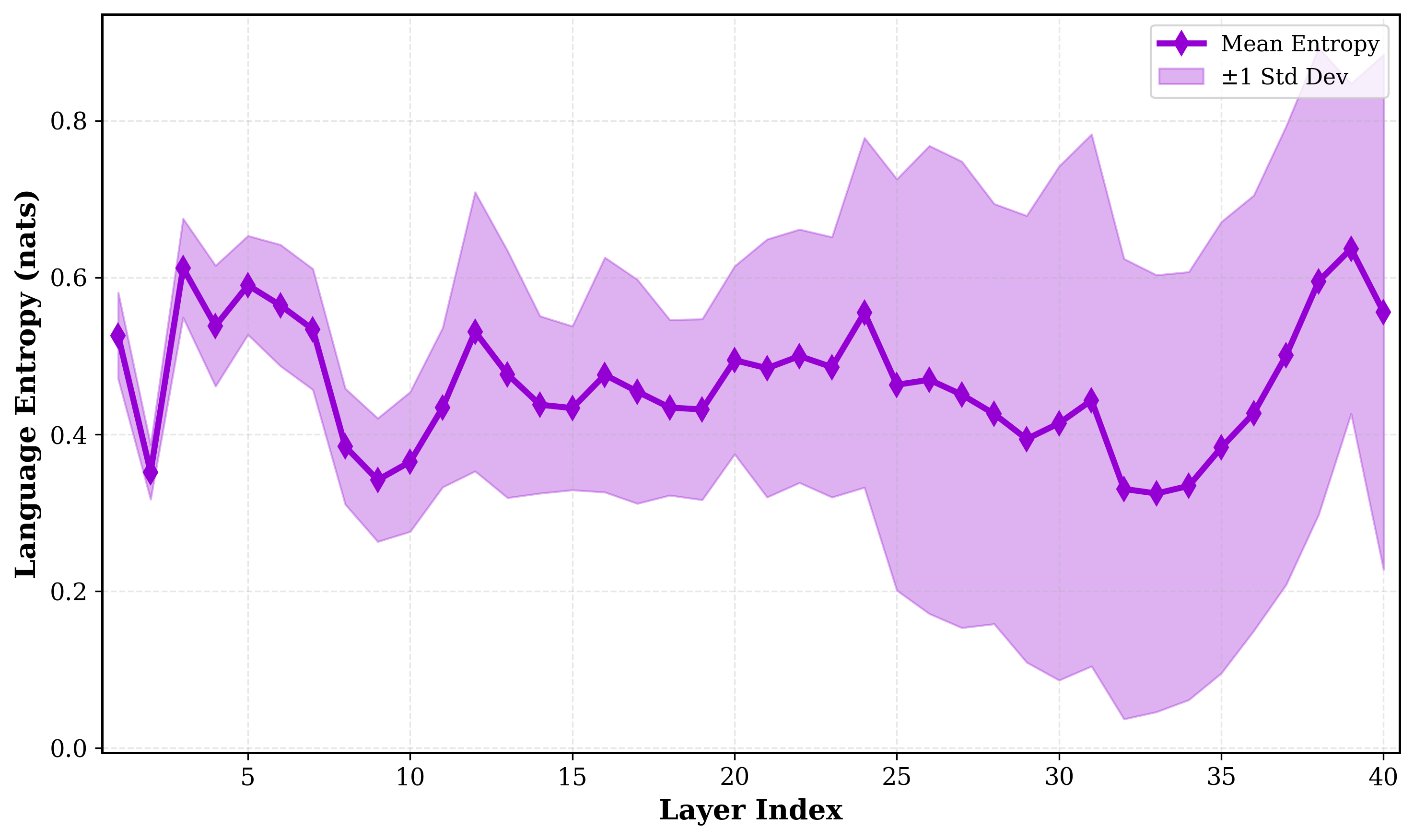}
        \caption{Evolution of target language probability.}
        \label{fig:logitlens_evolution_nemo}
    \end{subfigure}
    \hfill
    \begin{subfigure}[t]{0.48\textwidth}
        \includegraphics[width=\linewidth]{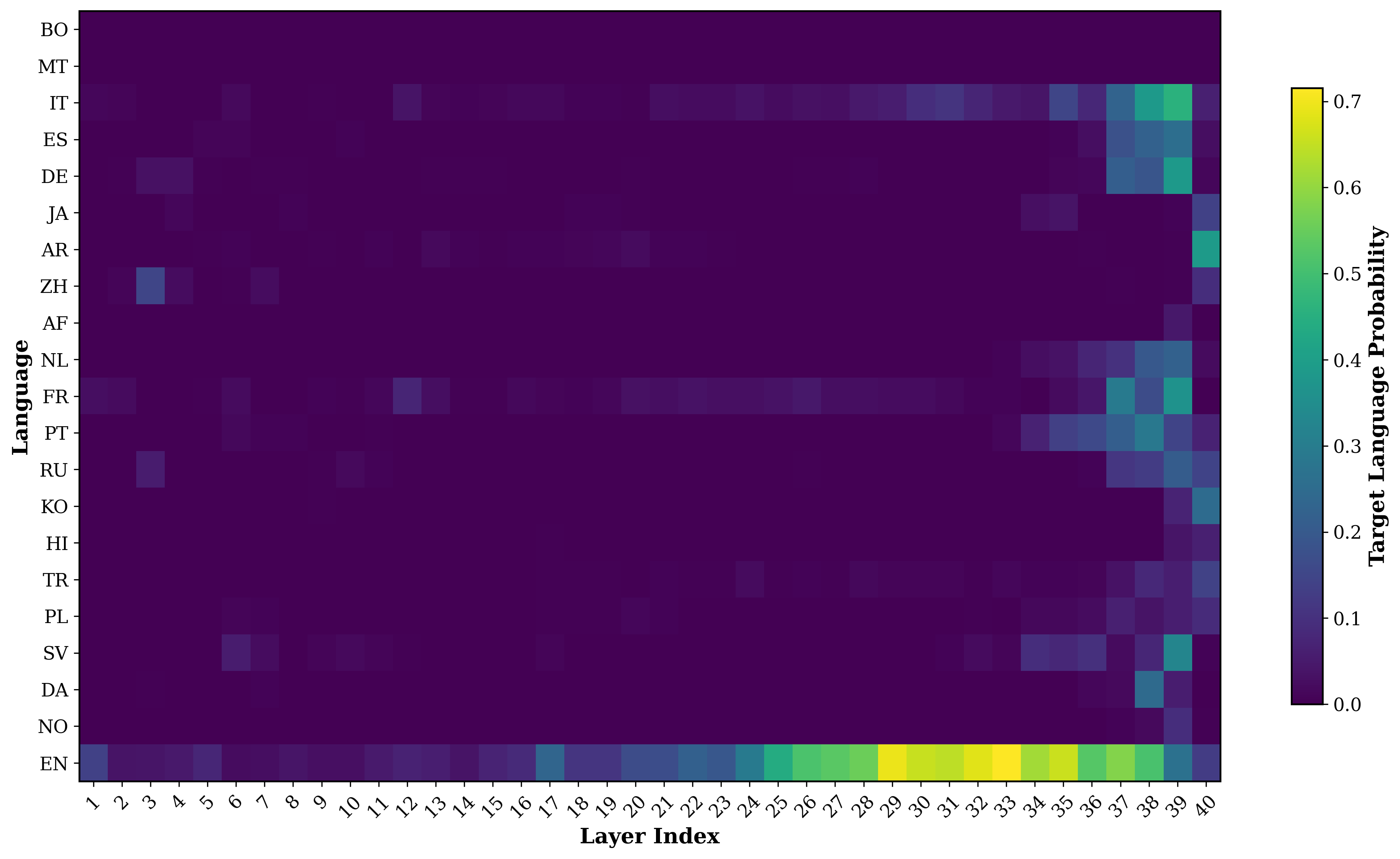}
        \caption{Heatmap of English probabilities.}
        \label{fig:logitlens_heatmap_nemo}
    \end{subfigure}
    
    \caption{Peeking into language prediction across layers using the logit lens for \textbf{Mistral-Nemo}.}
    \label{fig:logitlens_combined_nemo}
\end{figure*}

\clearpage

\begin{figure*}[h]
    \centering
    \begin{subfigure}[t]{0.48\textwidth}
        \includegraphics[width=\linewidth]{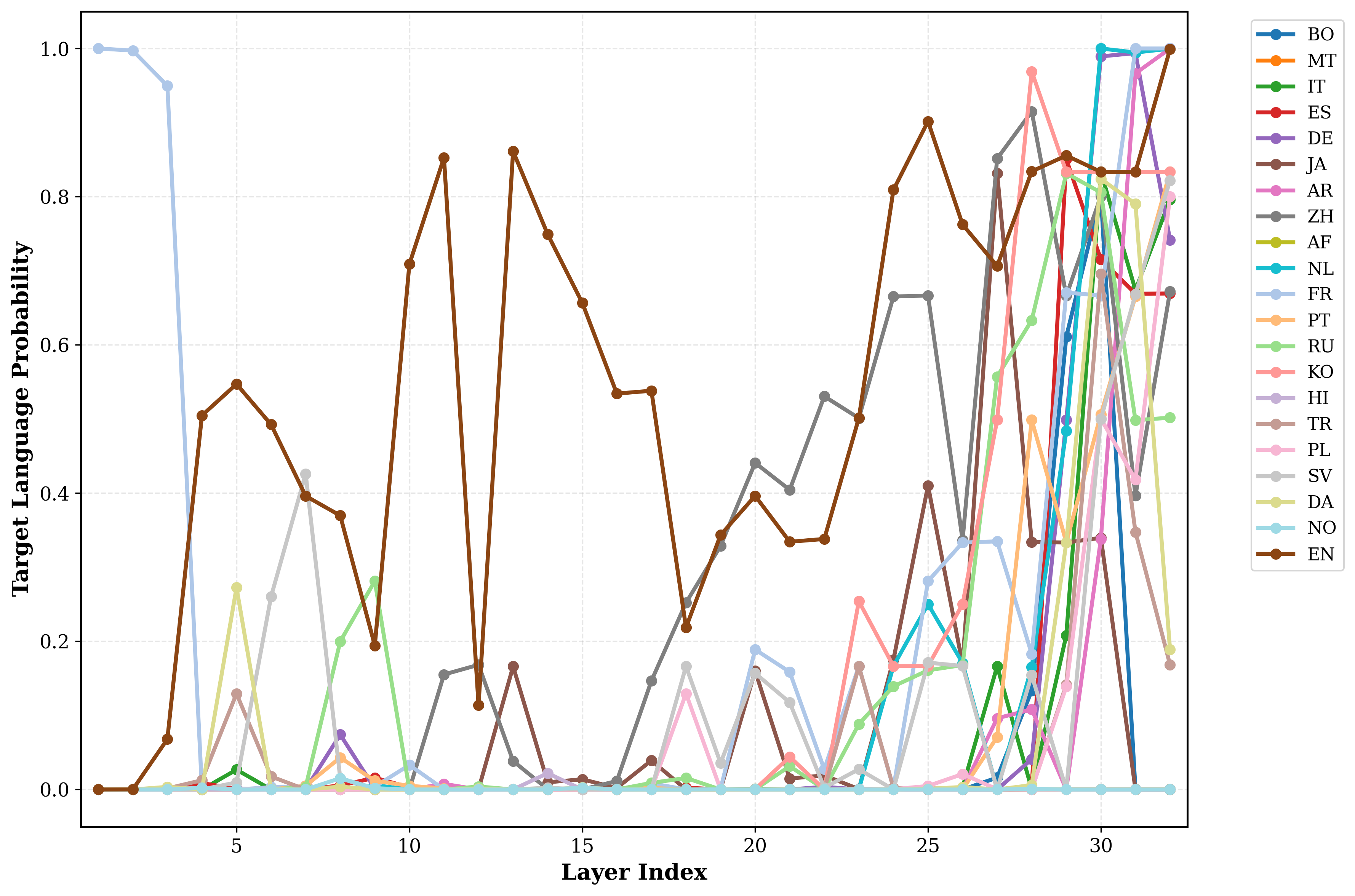}
        \caption{Probabilities of the target languages.}
        \label{fig:logitlens_target_aya-8}
    \end{subfigure}
    \hfill
    \begin{subfigure}[t]{0.48\textwidth}
        \includegraphics[width=\linewidth]{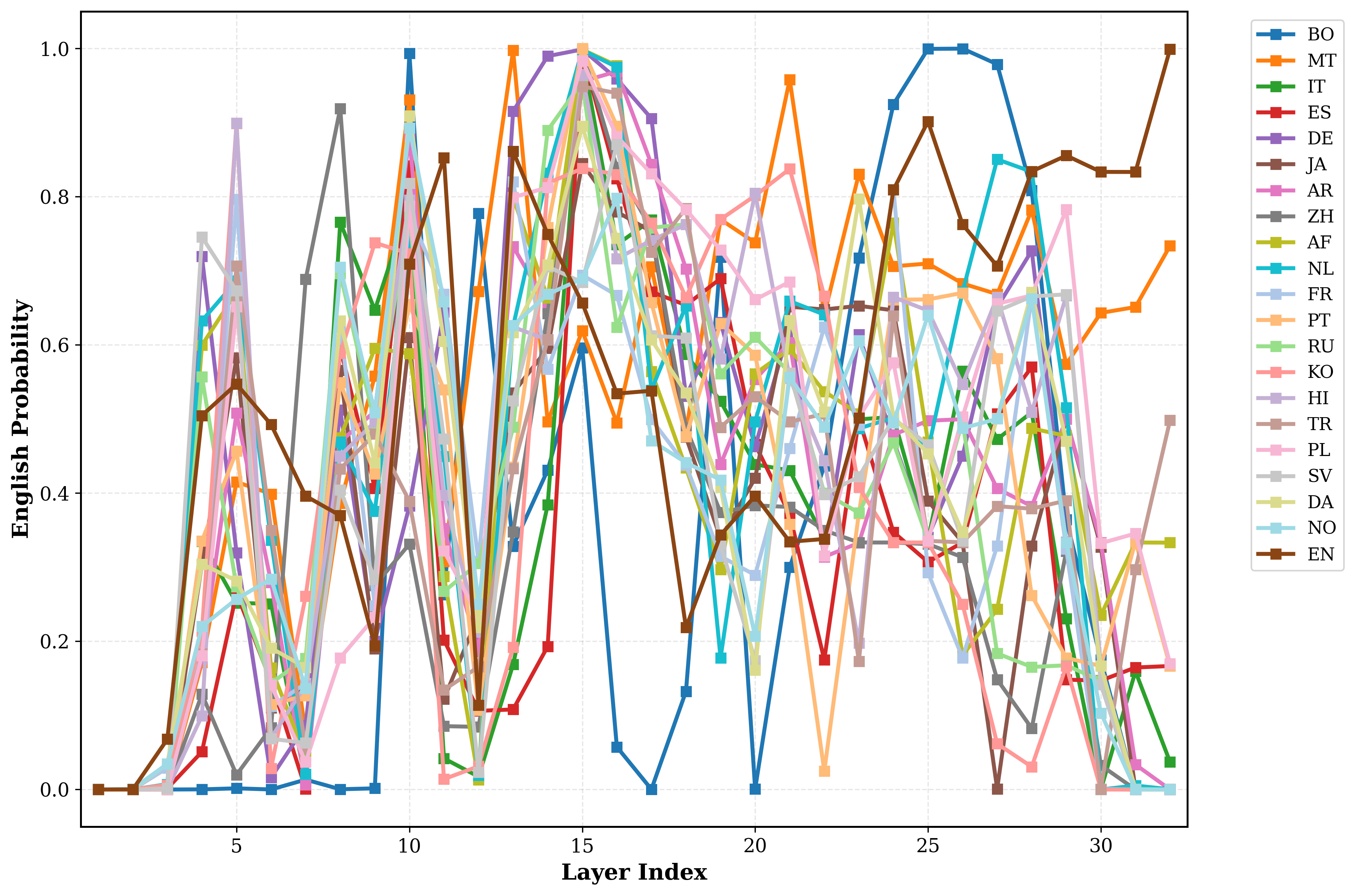}
        \caption{Probabilities of English.}
        \label{fig:logitlens_english_aya-8}
    \end{subfigure}
    
    \vspace{0.5em}
    
    \begin{subfigure}[t]{0.48\textwidth}
        \includegraphics[width=\linewidth]{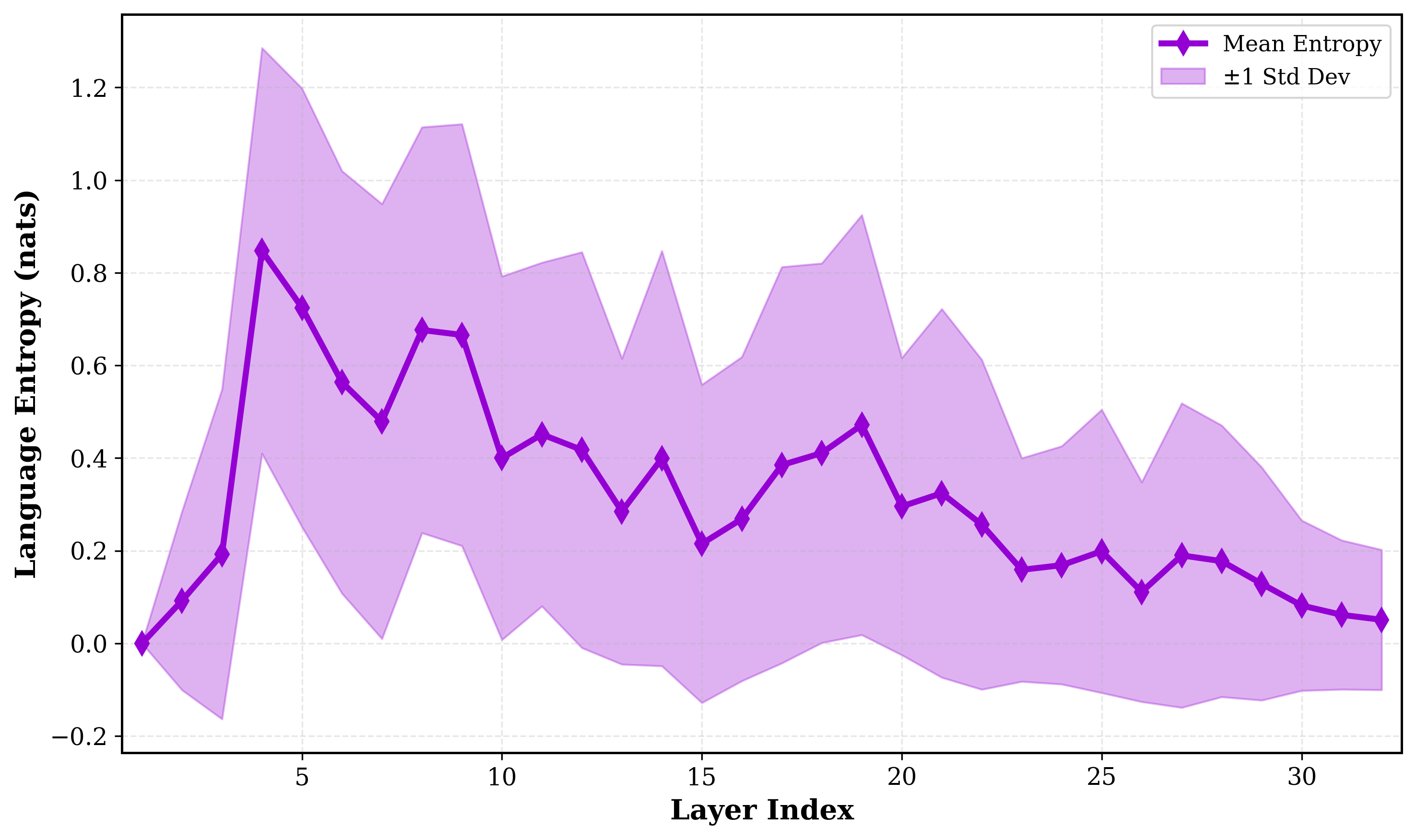}
        \caption{Evolution of target language probability.}
        \label{fig:logitlens_evolution_aya-8}
    \end{subfigure}
    \hfill
    \begin{subfigure}[t]{0.48\textwidth}
        \includegraphics[width=\linewidth]{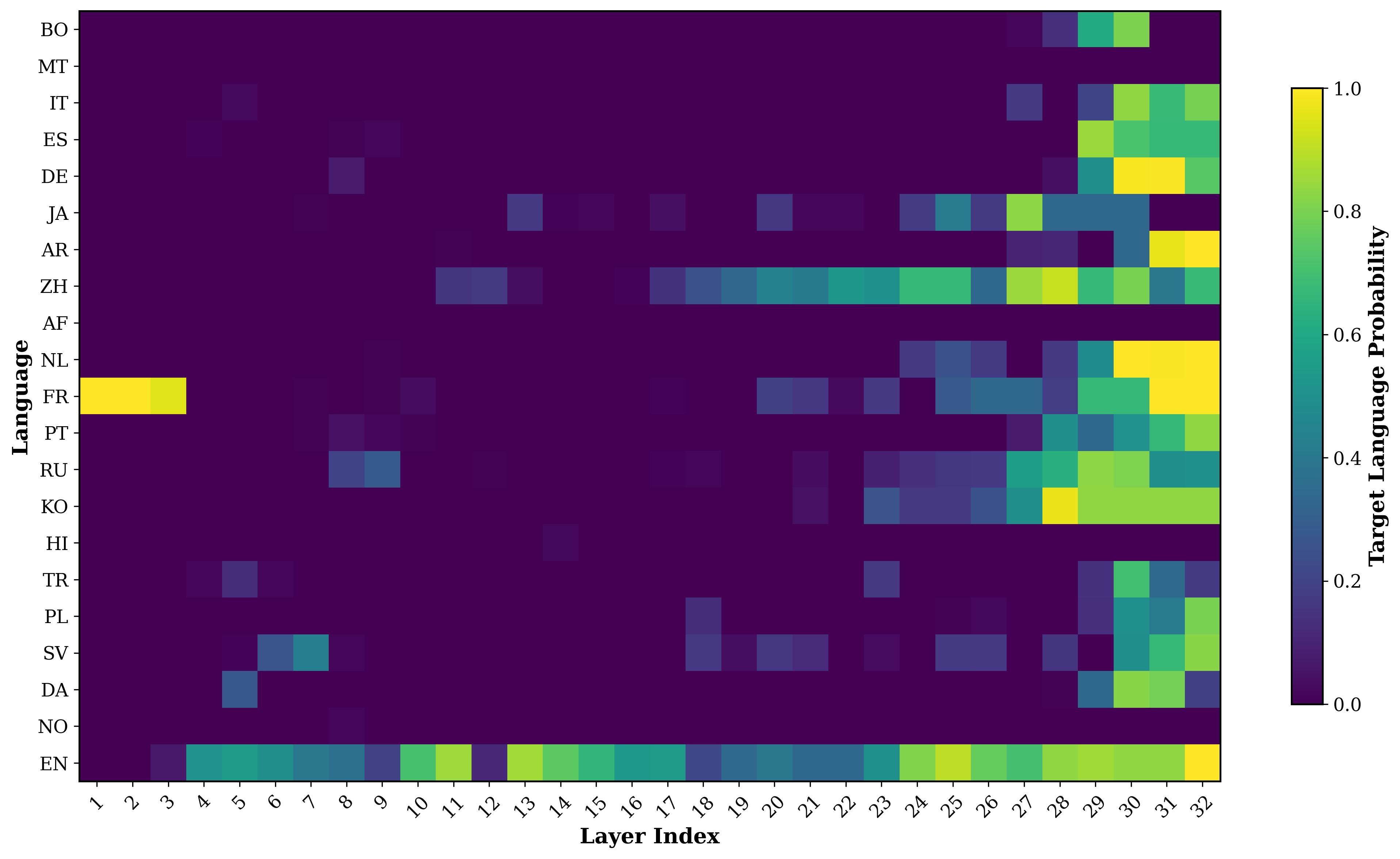}
        \caption{Heatmap of English probabilities.}
        \label{fig:logitlens_heatmap_aya-8}
    \end{subfigure}
    
    \caption{Peeking into language prediction across layers using the logit lens for \textbf{Aya-Expanse-8B}.}
    \label{fig:logitlens_combined_aya-8}
\end{figure*}

\begin{figure*}[t]
    \centering
    \begin{subfigure}[t]{0.48\textwidth}
        \includegraphics[width=\linewidth]{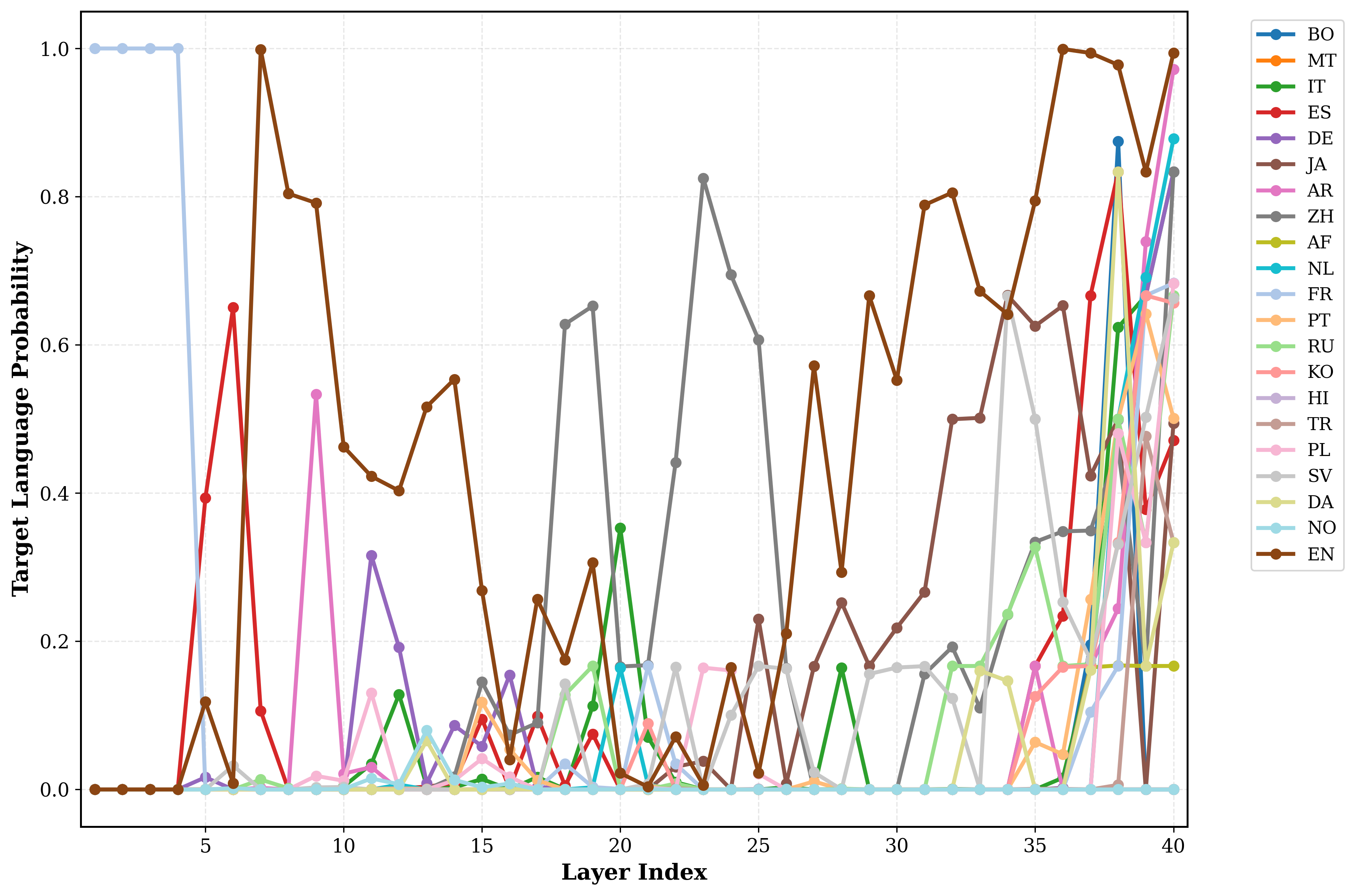}
        \caption{Probabilities of the target languages.}
        \label{fig:logitlens_target_aya-32}
    \end{subfigure}
    \hfill
    \begin{subfigure}[t]{0.48\textwidth}
        \includegraphics[width=\linewidth]{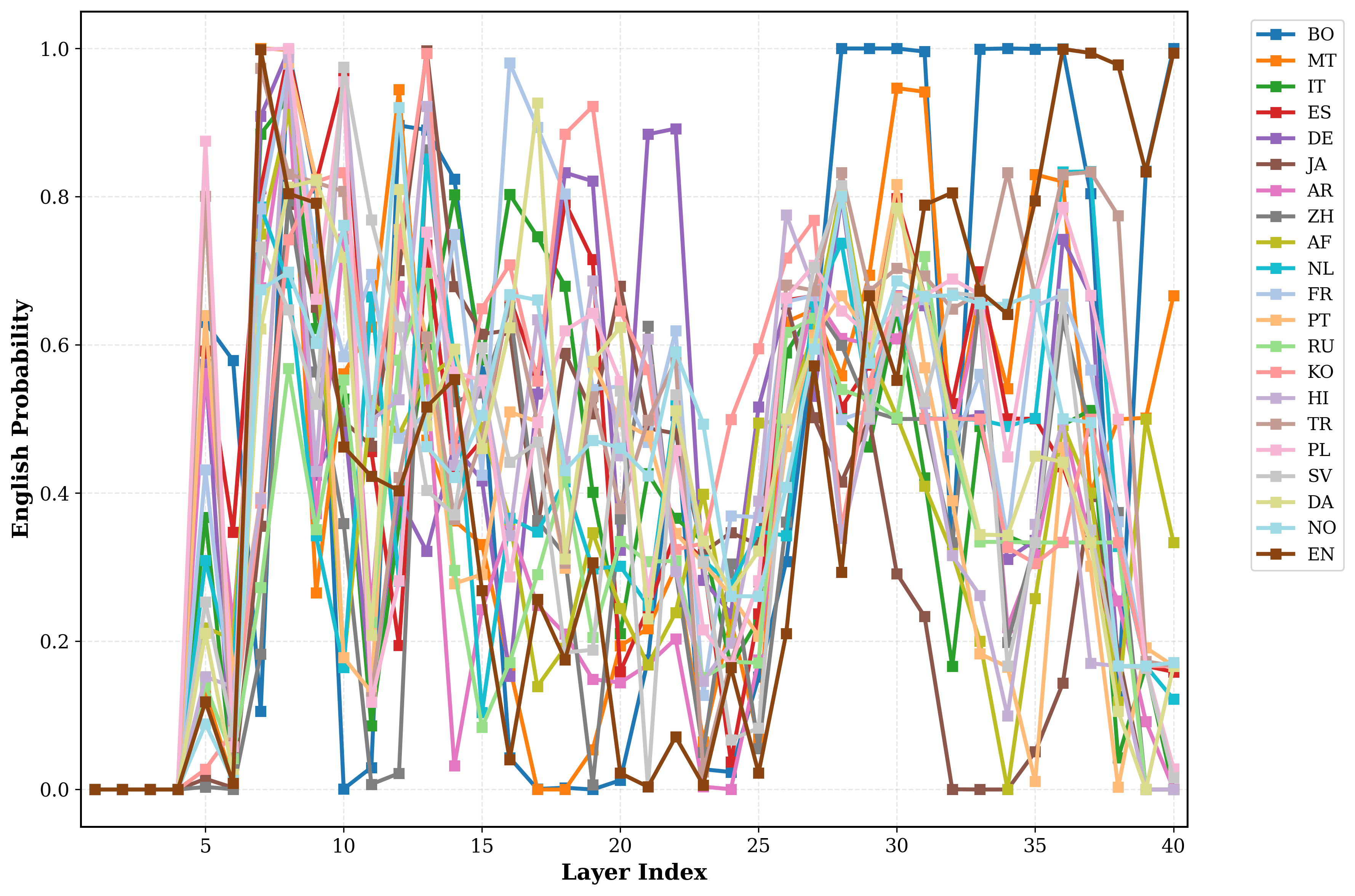}
        \caption{Probabilities of English.}
        \label{fig:logitlens_english_aya-32}
    \end{subfigure}
    
    \vspace{0.5em}
    
    \begin{subfigure}[t]{0.48\textwidth}
        \includegraphics[width=\linewidth]{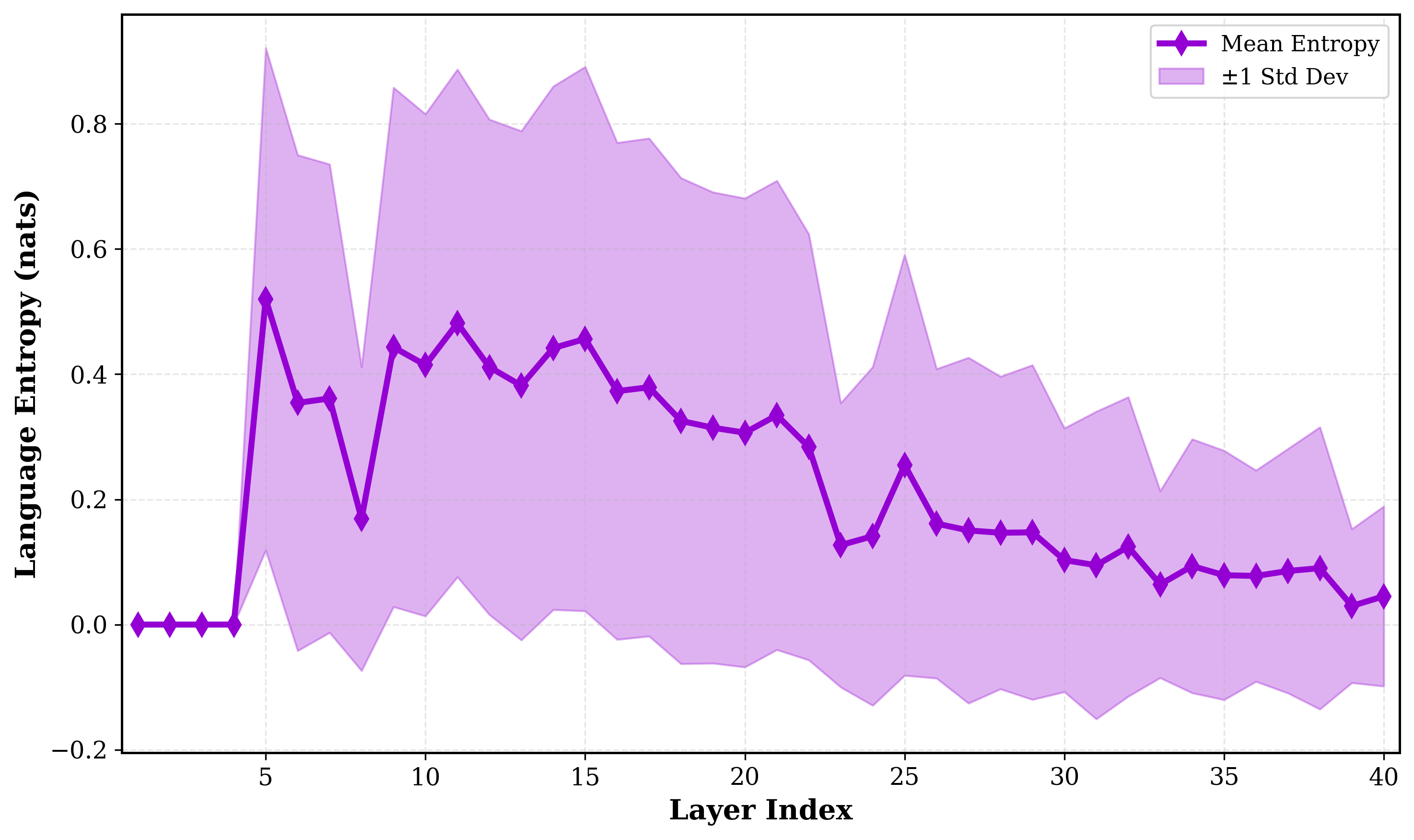}
        \caption{Evolution of target language probability.}
        \label{fig:logitlens_evolution_aya-32}
    \end{subfigure}
    \hfill
    \begin{subfigure}[t]{0.48\textwidth}
        \includegraphics[width=\linewidth]{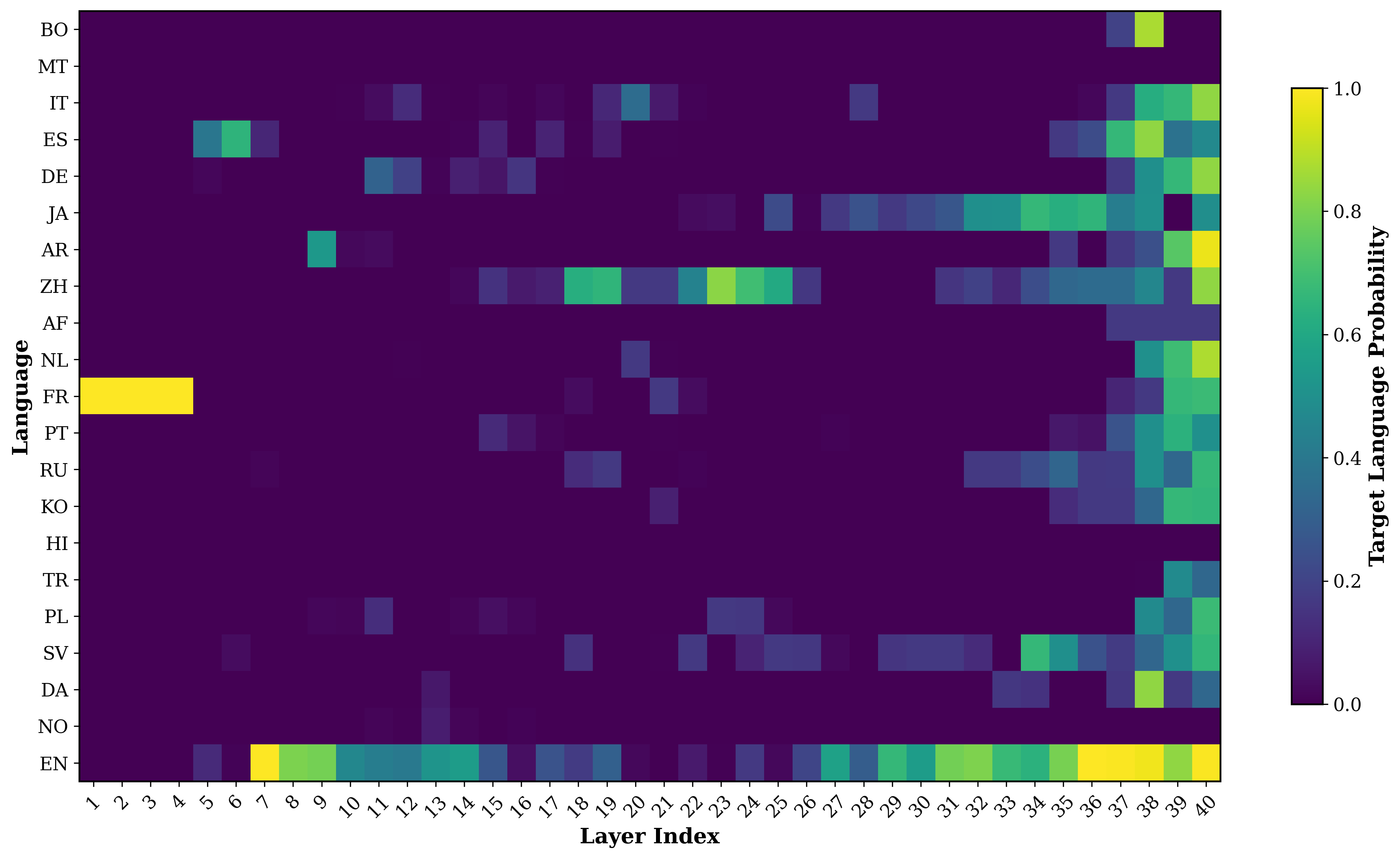}
        \caption{Heatmap of English probabilities.}
        \label{fig:logitlens_heatmap_aya-32}
    \end{subfigure}
    
    \caption{Peeking into language prediction across layers using the logit lens for \textbf{Aya-Expanse-32B}.}
    \label{fig:logitlens_combined_aya-32}
\end{figure*}

\clearpage


\begin{figure*}[hbt]
\section{Language Forcing}
\label{app:forcing}
\centering
\begin{minipage}[t]{0.48\textwidth}
  \centering
  \begin{subfigure}[t]{\linewidth}
    \includegraphics[width=\linewidth]{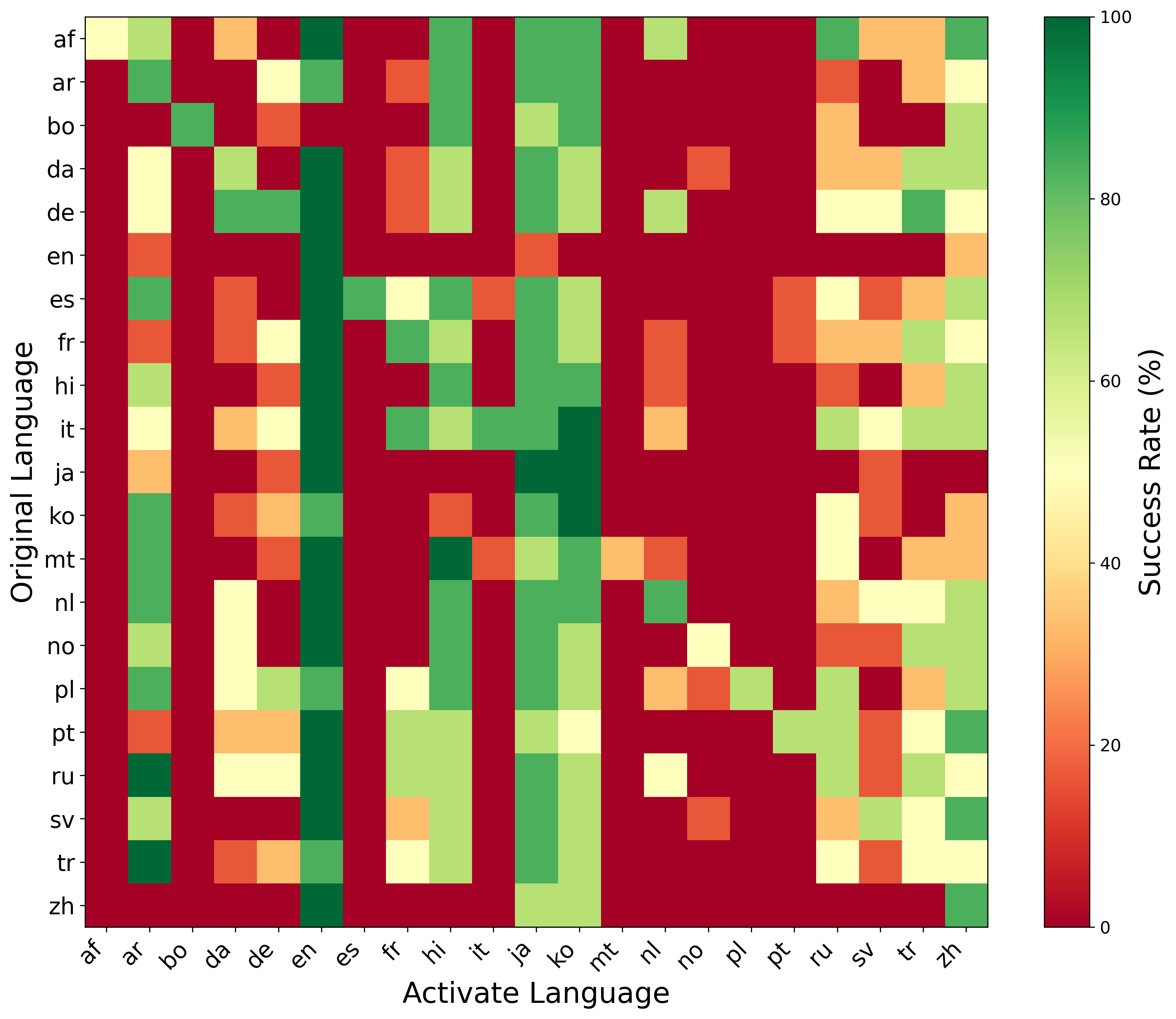}
    \caption{Activate-only - Top 1\%}
  \end{subfigure}
  \vspace{0.2cm}
  \begin{subfigure}[t]{\linewidth}
    \includegraphics[width=\linewidth]{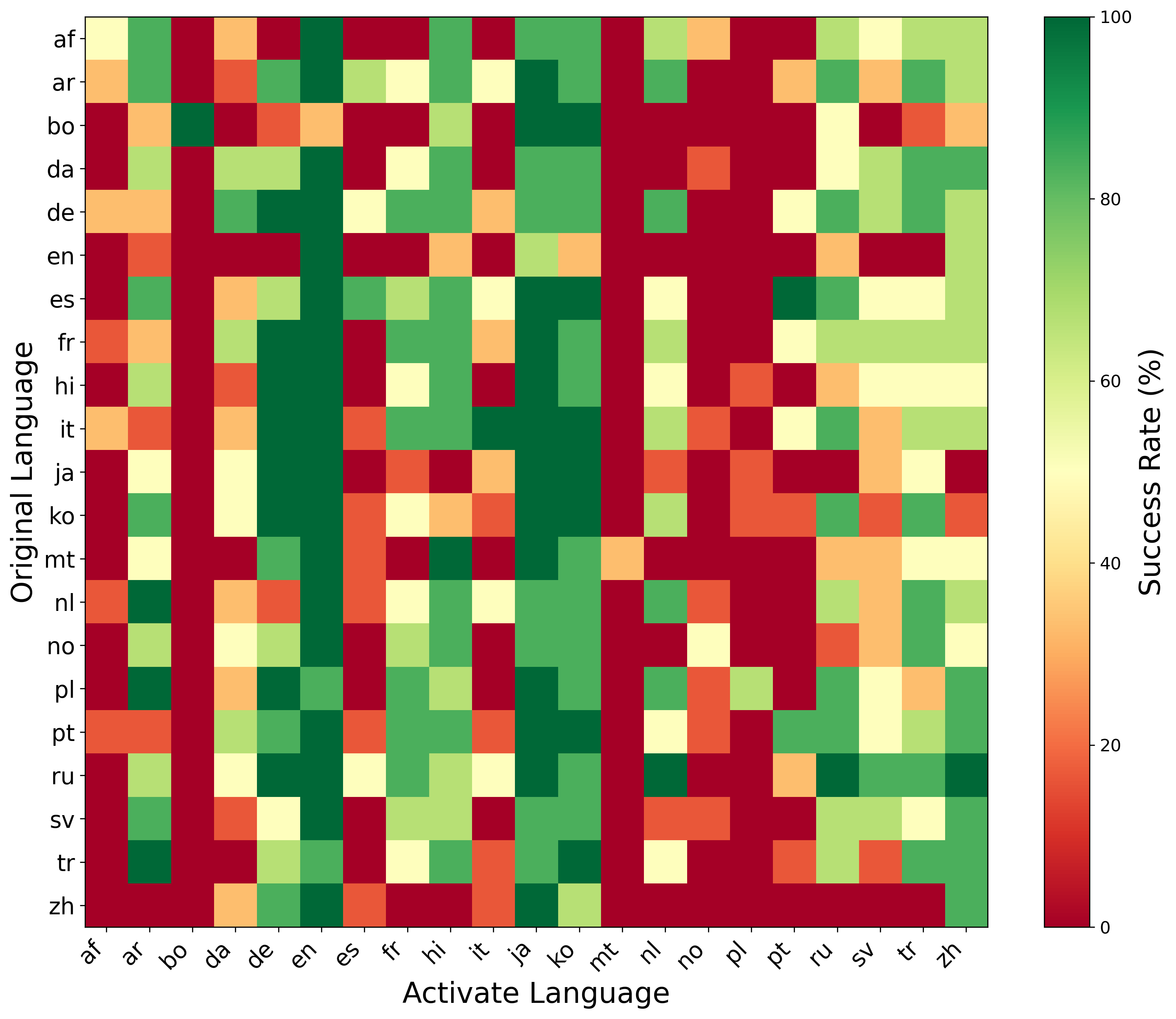}
    \caption{Activate-only - Top 5\%}
  \end{subfigure}
\end{minipage}
\hfill
\begin{minipage}[t]{0.48\textwidth}
  \centering
  \begin{subfigure}[t]{\linewidth}
    \includegraphics[width=\linewidth]{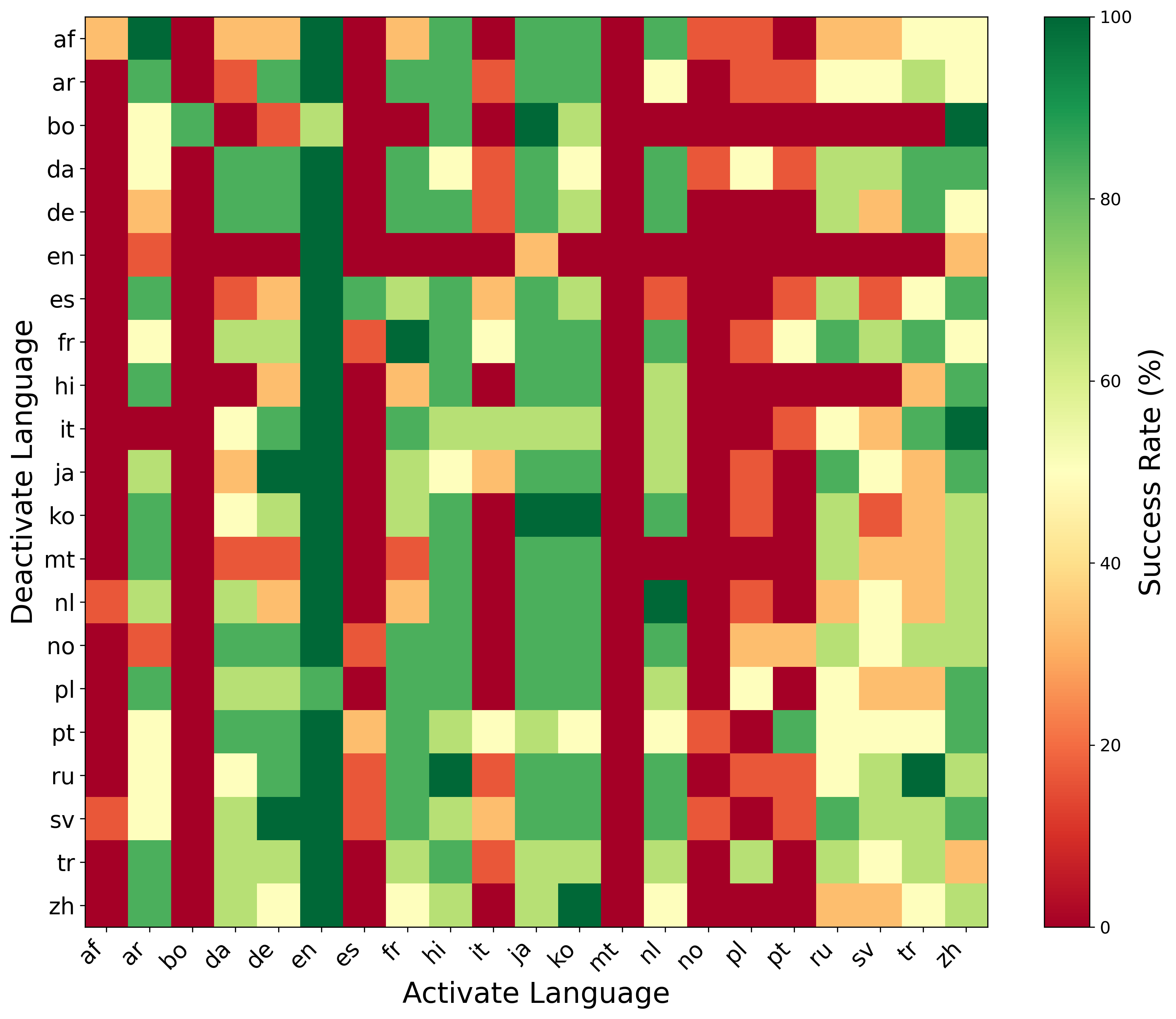}
    \caption{Deactivate-Activate - Top 1\%}
  \end{subfigure}
  \vspace{0.2cm}
  \begin{subfigure}[t]{\linewidth}
    \includegraphics[width=\linewidth]{assets/deactivate_activate_llama_5.png}
    \caption{Deactivate-Activate - Top 5\%}
  \end{subfigure}
\end{minipage}
\caption{Detailed language forcing results across 2 neuron ratios for two manipulation strategies - activate-only and deactivate-activate for \textbf{Llama-3.1-8B}. For deactivation, we set the neurons to 0.}
\label{fig:forcing_llama_2}
\end{figure*}
\clearpage

\begin{table*}[t]
\small
\centering
\resizebox{\textwidth}{!}{
\begin{tabular}{l l l ccccc}
\toprule
\textbf{Model} & \textbf{Intervention} & \textbf{Strategy} & \textbf{Top 1\%} & \textbf{Top 2\%} & \textbf{Top 3\%} & \textbf{Top 4\%} & \textbf{Top 5\%} \\
\midrule
\multirow{4}{*}{Mistral-Nemo} 
              & Additive     & Activate              & 17.49\% & 18.93\% & 19.99\% & 22.45\% & 22.75\% \\
              & Additive     &  Deactivate + Activate & \cellcolor{mediumgreen}25.54\% & \cellcolor{mediumgreen}27.51\% & \cellcolor{mediumgreen}29.59\% & \cellcolor{mediumgreen}32.39\% & \cellcolor{mediumgreen}33.22\% \\
              & Replacement  & Activate              & 12.43\%  & 12.69\%      & 13.86\%      & 15.19\%      & 14.47\%      \\
              & Replacement  & Deactivate + Activate & 20.03\% & 21.08\%      & 22.22\%      & 24.75\%      & 25.88\%      \\
              & DiffMean & Activate & 14.51\% & 16.18\% & 16.55\% & 17.15\% & 18.14\% \\
              & DiffMean & Deactivate + Activate & \cellcolor{lightgreen}21.96\% & \cellcolor{lightgreen}23.69\% & \cellcolor{lightgreen}25.62\% & \cellcolor{lightgreen}27.36\% & \cellcolor{lightgreen}28.11\% \\
\bottomrule
\end{tabular}
}
\caption{Overall success rates (\%) of language forcing for \textbf{Mistral-Nemo-12B} using three intervention types and two manipulation strategies across different top-\textit{k}\% neuron thresholds.}
\label{tab:language_forcing_nemo}
\end{table*}

\begin{table*}[ht]
\small
\centering
\resizebox{\textwidth}{!}{
\begin{tabular}{l l l ccccc}
\toprule
\textbf{Model} & \textbf{Intervention} & \textbf{Strategy} & \textbf{Top 1\%} & \textbf{Top 2\%} & \textbf{Top 3\%} & \textbf{Top 4\%} & \textbf{Top 5\%} \\
\midrule
\multirow{4}{*}{Aya-Expanse-8B} 
              & Additive     & Activate              & 66.40\% & 77.48\% & 81.29\% & 82.58\% & 84.73\% \\
              & Additive     &  Deactivate + Activate & \cellcolor{mediumgreen}75.59\% & \cellcolor{mediumgreen}83.30\% & \cellcolor{mediumgreen}85.22\% & \cellcolor{mediumgreen}84.47\% & \cellcolor{mediumgreen}86.81\% \\
              & Replacement  & Activate              & 49.62\%  & 61.19\%      & 65.31\%      & 65.61\%      & 69.39\%      \\
              & Replacement  & Deactivate + Activate & 65.72\% & 73.96\%      & 73.99\%      & 74.11\%      & 77.06\%      \\
              & DiffMean & Activate & 62.85\% & 70.52\% & 75.43\% & 77.82\% & 80.88\% \\
              & DiffMean & Deactivate + Activate & \cellcolor{lightgreen}73.74\% & \cellcolor{lightgreen}78.87\% & \cellcolor{lightgreen}81.67\% & \cellcolor{lightgreen}82.20\% & \cellcolor{lightgreen}84.24\% \\
\bottomrule
\end{tabular}
}
\caption{Overall success rates (\%) of language forcing for \textbf{Aya-Expanse-8B} using three intervention types and two manipulation strategies across different top-\textit{k}\% neuron thresholds.}
\label{tab:language_forcing_aya_8}
\end{table*}

\clearpage


\begin{figure*}[hbt]
\section{Language "Fallbacks"}
\label{app:fallbacks}
\centering

\begin{subfigure}{\linewidth}
    \centering
    \includegraphics[width=0.99\linewidth]{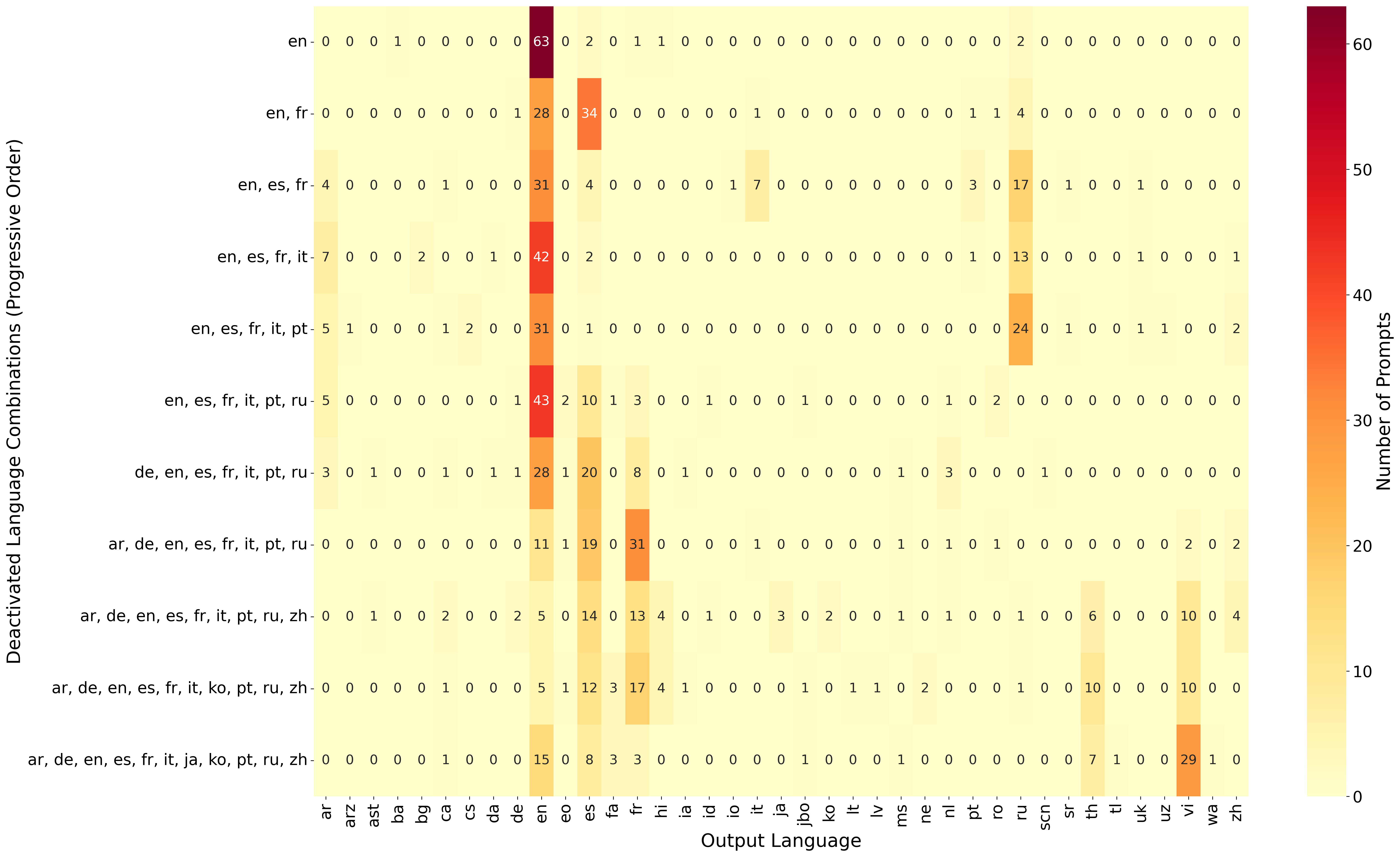}
    \caption{Top 4\% neurons}
\end{subfigure}

\vspace{0.5em}

\begin{subfigure}{\linewidth}
    \centering
    \includegraphics[width=0.99\linewidth]{assets/fallbacks_llama_5.png}
    \caption{Top 5\% neurons}
\end{subfigure}

\caption{Progressive deactivation of language-specific neurons for high-resource languages for \textbf{Llama-3.1-8B}. We set the language neurons to -1 for deactivation in this scenario; 0 does not produce the required effects.}
\label{fig:fallbacks_llama}
\end{figure*}

\begin{figure*}[hbt]
\begin{subfigure}{\linewidth}
    \centering
    \includegraphics[width=0.99\linewidth]{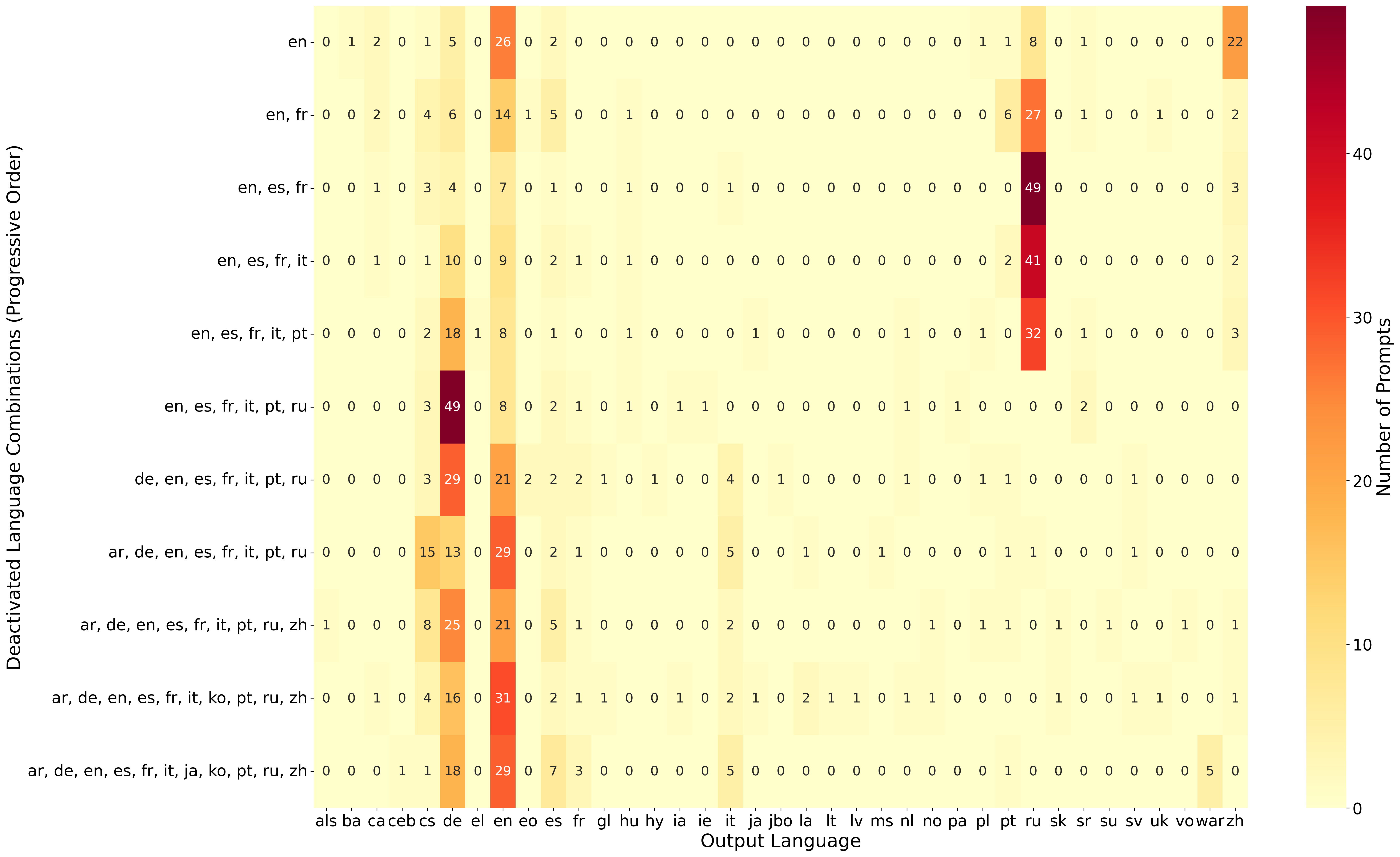}
    \caption{Top 4\% neurons}
\end{subfigure}

\vspace{0.5em}

\begin{subfigure}{\linewidth}
    \centering
    \includegraphics[width=0.99\linewidth]{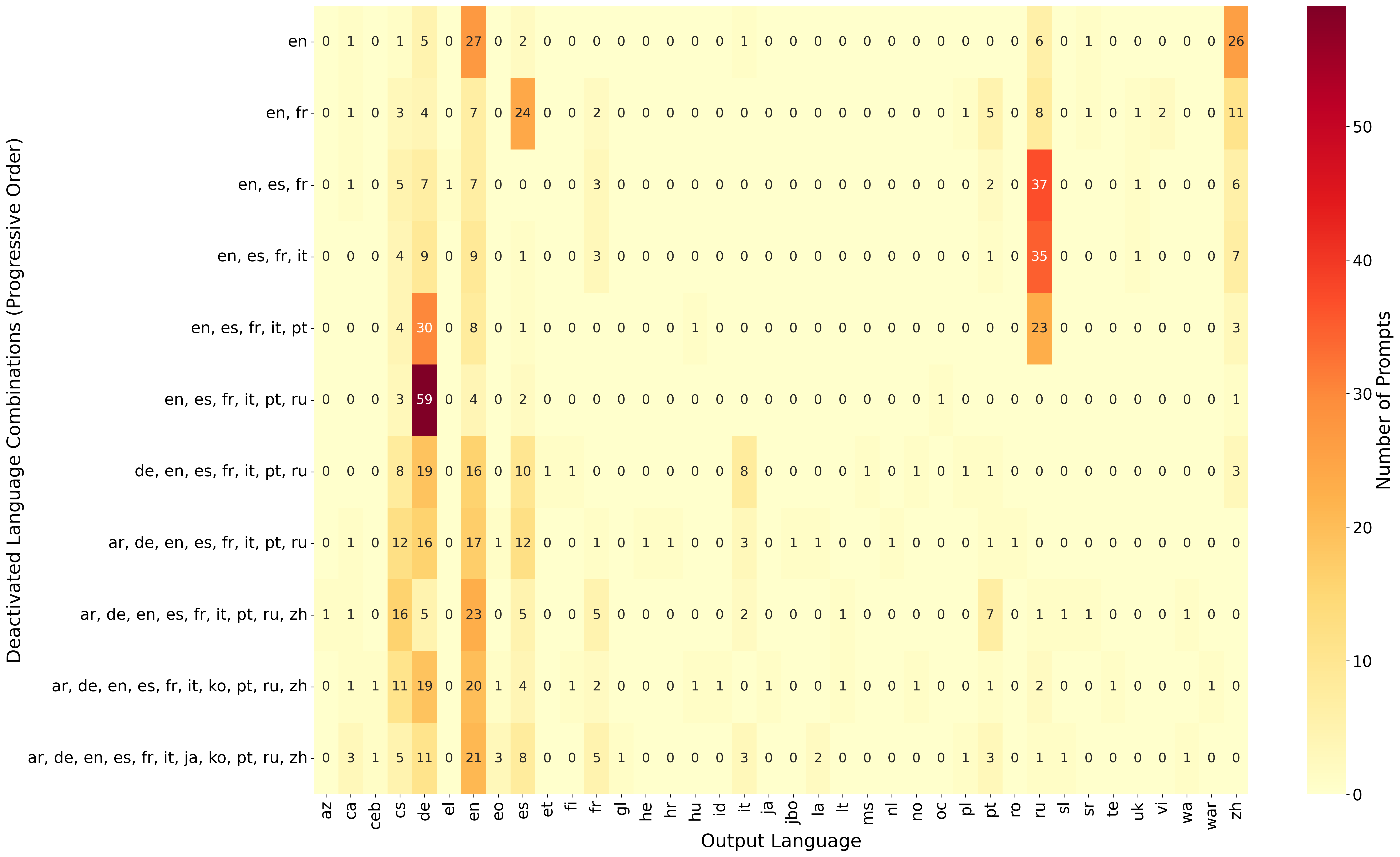}
    \caption{Top 5\% neurons}
\end{subfigure}

\caption{Progressive deactivation of language-specific neurons for high-resource languages for \textbf{Mistral-Nemo}. We set the language neurons to -2.5 for deactivation in this scenario; values from 0 to -2 do not produce the required effects.}
\label{fig:fallbacks_nemo}
\end{figure*}

\end{document}